\documentclass[11pt]{article}

\usepackage[inline]{enumitem}
\usepackage{floatrow}
\usepackage{tikz}
\usetikzlibrary{positioning}

\usepackage[inkscapelatex=false]{svg}

\usepackage[textsize=tiny]{todonotes}

\usepackage{tcolorbox}
\usepackage{multipleFigureHelpers}
\usepackage{comment}
\usepackage{algorithm}
\usepackage{algpseudocode}
\usepackage{placeins}

\makeatletter
\newcounter{phase}[algorithm]
\newlength{\phaserulewidth}
\newcommand{\setphaserulewidth}{\setlength{\phaserulewidth}}

\makeatother

\newcommand{\suppfigref}[1]{Supplementary Figure~\ref{#1}}
\newcommand{\supptabref}[1]{Supplementary Table~\ref{#1}}
\newcommand{\suppsecref}[1]{Supplementary Section~\ref{#1}}
\newcommand{\suppfigrangeref}[2]{Supplementary Figures~\ref{#1}--\ref{#2}}

\newtheorem{corollary}{Corollary}

\setphaserulewidth{.7pt}

\usepackage[perpage]{footmisc}
\usepackage[numbers,sort&compress]{natbib}
\usepackage{commands}
\usepackage{multirow}
\usepackage{tabularx}
\usepackage{authblk}

\usepackage{array}
\title{MEDAL: Manifold Embedding Distillation via Autoencoder Learning}

\usepackage{authblk}

\author[1]{Irene Chang}

\author[2]{Tarek M. Zikry}

\author[1,3,4,{\dag}]{Genevera I. Allen}

\affil[1]{Department of Statistics, Columbia University, New York, NY 10027}

\affil[2]{School of Data and Information Sciences, University of North Carolina at Chapel Hill, Chapel Hill, NC 27599}

\affil[3]{Irving Institute for Cancer Dynamics, Columbia University, New York, NY 10027}

\affil[4]{Center for Theoretical Neuroscience, Zuckerman Mind Brain Behavior Institute, Columbia University, New York, NY 10027}
\date{}
\begin{document}

\maketitle

\begingroup

\renewcommand{\thefootnote}{\fnsymbol{footnote}}

\footnotetext[2]{Corresponding author: \url{genevera.allen@columbia.edu}}

\endgroup

\begin{abstract}
Low-dimensional embeddings are widely used as visual summaries of high-dimensional data and to enable downstream scientific discoveries.  Yet, popular nonlinear dimension reduction methods, such as t-SNE and UMAP, are often selected based on visual appeal alone and without rigorous quantitative validation. A major reason is that manifold embeddings typically do not provide an out-of-sample map nor an inverse back to the original feature space; this makes held-out validation, the gold standard in supervised learning, all but impossible.  To address these challenges, we develop a novel framework, MEDAL (Manifold Embedding Distillation via Autoencoder Learning), which distills a fitted manifold embedding into a reusable encoder--decoder model. MEDAL trains a constrained autoencoder whose bottleneck exactly matches any teacher embedding while the decoder reconstructs the original input; this yields an explicit map for new samples, an approximate inverse, and a pointwise reconstruction-based measure of distortion in the manifold space. This converts static manifold embeddings into models that can be evaluated on held-out data, enabling quantitative validation including comparing different dimension reduction methods as well as hyperparameter tuning. Across multiple benchmark and scientific case studies, we show that MEDAL enables held-out validation to determine optimal manifold embeddings and hyperparameters, reveals biologically coherent regions that are difficult to preserve in two dimensional embeddings, and detects distribution shift when new samples are mapped into a fixed reference manifold. MEDAL provides a general validation wrapper to any existing dimension reduction technique that will improve the rigor and reliability of dimension reduction in scientific workflows.  
\end{abstract}


\section{Introduction}
Data arising in modern scientific applications are often very high-dimensional. This growth in data dimensionality has made dimension reduction (DR) a central tool for modern scientific analysis. Broadly, DR seeks to map observations from $\mathbb{R}^p$ into a much lower-dimensional space $\mathbb{R}^d$, where $d \ll p$, while preserving structure relevant to exploration, interpretation, and downstream analysis. A large methodological toolbox has been developed for this purpose. Linear methods such as Principal Component Analysis (PCA) \cite{hotelling1933analysis} are well-understood, but limited in their ability to represent nonlinear structure that are present in complex, modern datasets. Neighborhood- and graph-based nonlinear methods were introduced to overcome this limitation, beginning with classical approaches such as Isomap \cite{tenenbaum2000global} and Locally Linear Embedding \cite{roweis2000nonlinear}, and extending to modern methods such as t-SNE \cite{van2008visualizing}, UMAP \cite{mcinnes2018umap}, PHATE \cite{moon2019visualizing}, and PaCMAP \cite{wang2021understandingdimensionreductiontools}. These methods are no longer used only as exploratory visualization tools. 
Across biomedicine and other data-intensive sciences, low-dimensional embeddings are fundamental for discovery, used to guide decisions about cluster structure, class separation, trajectory inference, and disease-related heterogeneity, including in settings such as cancer characterization via single-cell transcriptomics, biomedical imaging, and electrophysiological measurements of neural activity \cite{Cashman_2025, fotopoulou2024review, debodt2025lowdimensionalembeddingshighdimensionaldata, kobak2019art, MOON201836, lunga2013manifold, cunningham2014dimensionality, orlov2025high, zhang2025comprehensive, blanco2022strategies, zikry2024cell, churchland2012neural}.
As a result, an embedding is not merely a picture; it is often treated as a scientifically meaningful representation of the data, and its reliability can directly affect critical downstream conclusions.

Despite this importance, the way embeddings are often used in practice remains surprisingly qualitative. Researchers frequently choose a DR method and its hyperparameters based on defaults or visual appeal, and then proceed as if the resulting geometry were a faithful representation of the underlying data. This workflow is fragile. Small changes in initialization, preprocessing, feature selection, or hyperparameter choice can materially alter the geometry of an embedding \cite{wang2021understandingdimensionreductiontools,kobak2021initialization,townes2019feature, gan2025machine}, and recent critiques have highlighted how easily visually compelling maps can be over-interpreted as scientific evidence \cite{jeon2025stopmisusingtsneumap,huang2022evaluation,chari2023specious}. This has also contributed to the greater crisis in rigor and reproducibility in modern science \cite{national2019reproducibility}, as these sensitive embeddings are used in consequential and critical scientific workflows. This fragility calls for more principled ideas from model validation and reliability assessment.

A growing literature has begun to address the problem of validation in dimension reduction \cite{allen2024irmreview}, particularly through work on hyperparameter tuning and embedding reliability. Existing approaches include post-hoc selection criteria tailored to particular applications or notions of embedding quality \cite{cao2017automaticselectiontsneperplexity,Lin_2024,gikera2025khyperparam,10811181, zikry2025limeade}, optimization-based formulations of hyperparameter search \cite{9476903,liao2023efficientrobustbayesianselection,nasim2025automatedmanifoldlearningreduced}, and tools for scaling visual or interactive exploration of embedding behavior \cite{appleby2021hypernpinteractivevisualexploration,sun2026dynamicviz}. Most closely related are methods that score local embedding reliability, including empirical-null or resampling-based approaches such as EMBEDR and scDEED \cite{johnson2022embedr,scdeeds,sun2026dynamicviz}, and geometric approaches based on map continuity or discontinuity such as neMDBD \cite{liu2025assessingimprovingreliabilityneighbor}. These methods address important aspects of the problem, but they remain task-specific or method-specific and do not provide the kind of principled held-out validation routine in supervised learning.

A central reason this gap persists is that principled validation is much easier for some DR methods than for others. For linear methods such as PCA, there exists both a forward map (projecting new data) and an inverse map (reconstruction from low-dimensional coordinates), providing a natural notion of reconstruction error. Therefore, a model trained on one subset of data can be tested by measuring information loss on held-out observations. This split-based approach mirrors the standard model selection procedure under supervised regime. By contrast, many widely used neighborhood-based nonlinear methods provide embeddings primarily for the training data and do not furnish a general, reliable way to embed new points or invert the map. As a result, nonlinear embeddings remain difficult to evaluate, compare, and reuse, despite already being deeply integrated in scientific workflows \cite{chang2025unsupervisedmachinelearningscientific}. 

Previous work has incorporated neural networks directly into manifold learning to address the absence of an explicit out-of-sample map. Parametric t-SNE and Parametric UMAP \cite{van2009learning, sainburg2021parametric}, for example, train a neural encoder from high-dimensional observations to low-dimensional coordinates, as well as autoencoder-based approaches that learn latent representations through reconstruction objectives \cite{tschannen2018recent}. However, despite the appeal of these methods, many applied analyses still use standard t-SNE and UMAP embeddings as fixed objects of interpretation \cite{jeon2025stopmisusingtsneumap}. What is missing is therefore not yet another embedding method to replace the existing ones, but rather, a \emph{general wrapper} that makes these nonparametric DR methods more reusable, comparable, and amenable to rigorous quantitative evaluation.

In this work, we propose \textbf{MEDAL}, a distillation framework that can be used to bridge the evaluation gap in DR by converting a fitted teacher embedding into a reusable model. MEDAL is inspired by knowledge distillation \cite{hinton2015distillingknowledgeneuralnetwork}, in which a \emph{student} model is trained to reproduce the behavior of a \emph{teacher} model more efficiently. Unlike classic distillation, our `teacher' is a pre‑computed embedding. Given a teacher embedding from UMAP, t‑SNE, Isomap, PCA or any other DR method, MEDAL trains a constrained autoencoder with two coupled objectives: (1) a distillation loss aligning the encoder with the teacher coordinates, and (2) a reconstruction loss that preserves input‑space information. (\autoref{fig:schematic}A). Once distilled, the embedding is no longer merely a static visualization; it becomes a model with an explicit forward map, an approximate inverse, and a pointwise reconstruction-based measure of information loss.

By equipping otherwise nonparametric embeddings with a shared learned mapping, MEDAL enables several capabilities that are typically unavailable within a single DR workflow (\autoref{fig:schematic}B). First, it makes held-out validation possible: new observations can be mapped into a fixed reference manifold and reconstructed back to the original space, hence embeddings can be assessed by how much information they preserve out of sample rather than by visual appeal alone. Second, this provides a principled basis for hyperparameter selection, since candidate teacher embeddings can be compared by held-out reconstruction error under a common protocol. More broadly, MEDAL creates a common evaluation scale for comparing different DR methods, a task that is otherwise difficult because PCA, UMAP, t-SNE, and related approaches provide different objects, diagnostics, and notions, or lack of notions, of goodness-of-fit. By scoring all teachers through the same learned reconstruction map, MEDAL allows method comparisons to be based on preserved input-space information rather than on method-specific heuristics or visual inspection. Finally, because reconstruction error is defined pointwise, MEDAL enables localized distortion analysis, identifying where the manifold is least faithfully represented. In this way, MEDAL does not replace existing DR methods, rather, it augments them with the tools needed for rigorous validation and interpretation.
\section{Methods}

\subsection{Overview: MEDAL as the validation protocol}
Linear methods like PCA allow for held-out validation because they provide both forward and inverse maps, making reconstruction error a natural measure of information loss. By contrast, widely-used nonlinear methods such as t-SNE, UMAP, and Isomap do not furnish general forward and inverse maps, making them difficult to evaluate on held-out data. MEDAL restores this validation structure by distilling an embedding generated from \emph{any} DR method of choice—the teacher—into an autoencoder. Autoencoders are encoder--decoder neural network models that learn low-dimensional representations by reconstructing the input from a compressed latent space \cite{hinton1994autoencoders}. In MEDAL, the compressed latent coordinates are not learned freely; instead, the encoder bottleneck is constrained to match a prescribed teacher embedding, while the decoder reconstructs the original high-dimensional input.

The MEDAL workflow is as follows: We first split the data into a training set $X_{\mathrm{tr}}$ and a held-out validation set $X_{\mathrm{val}}$. A teacher dimension reduction method is fitted only on $X_{\mathrm{tr}}$, producing training coordinates $Z_{\mathrm{tr}}$. We then train a student autoencoder on $X_{\mathrm{tr}}$ until the encoder matches the teacher embedding, $Z_{\mathrm{tr}}$, to a predefined, very small alignment threshold. The resulting encoder provides an out-of-sample map into the fitted teacher manifold, while the decoder turns reconstruction error into a quantitative measure of information preserved by that manifold. The held-out $X_{\mathrm{val}}$ is then passed through the learned encoder--decoder pair to obtain reconstruction errors for each validation observation. These errors provide a common basis for the four applications developed below, namely hyperparameter selection within a DR method, pointwise distortion analysis, out-of-sample embedding, and comparison across DR methods. 

The remainder of this section describes details on student training for rigorous interpretation, and then outlines the downstream validation tasks enabled by the resulting encoder--decoder pair.

\subsection{The MEDAL training objective}
Let
\[
X_{\mathrm{tr}}=\{x_i\}_{i=1}^{n_{\mathrm{tr}}}
\subset \mathbb{R}^p
\]
denote the training observations in the original feature space. The teacher
$T$ can be any dimension-reduction method fitted to $X_{\mathrm{tr}}$, producing low-dimensional coordinates
\[
Z_{\mathrm{tr}}=\{z_i\}_{i=1}^{n_{\mathrm{tr}}}
\subset \mathbb{R}^r, \qquad r<p.
\]
Here $z_i$ denotes the fitted teacher coordinate assigned to $x_i$.

MEDAL trains an encoder--decoder pair
\[
e_\theta:\mathbb{R}^p\to\mathbb{R}^r,
\qquad
d_\phi:\mathbb{R}^r\to\mathbb{R}^p,
\]
where the encoder bottleneck is constrained to reproduce the teacher
coordinates and the decoder reconstructs the original input. The reconstruction
loss on the training split is
\[
\mathcal{L}_{\mathrm{rec}}(X_{\mathrm{tr}};\theta,\phi)
=
\frac{1}{n_{\mathrm{tr}}}
\sum_{i=1}^{n_{\mathrm{tr}}}
\left\|
x_i-d_\phi(e_\theta(x_i))
\right\|^2,
\]
and the distillation loss is
\[
\mathcal{L}_{\mathrm{dist}}(X_{\mathrm{tr}},Z_{\mathrm{tr}};\theta)
=
\frac{1}{n_{\mathrm{tr}}}
\sum_{i=1}^{n_{\mathrm{tr}}}
\left\|
e_\theta(x_i)-z_i
\right\|^2.
\]

Conceptually, MEDAL seeks to minimize reconstruction while enforcing close agreement between the student bottleneck layer and the teacher embedding:
\begin{equation}
\min_{\theta,\phi}\; \mathcal{L}_\text{rec}(X_{\mathrm{tr}};\theta,\phi)
\quad \text{s.t.} \quad
\mathcal{L}_\text{dist}(X_{\mathrm{tr}},Z_{\mathrm{tr}};\theta)\le \varepsilon,
\label{eq:constrained}
\end{equation}
for a small alignment tolerance $\varepsilon \ge 0$. Operationally, we solve a Lagrangian relaxation of this problem, with a penalty $\lambda_d$ on the distillation loss
\begin{equation}
\min_{\theta,\phi}\;
\mathcal{L}(\theta,\phi)
=
\lambda_d\,\mathcal{L}_d(X,Z;\theta)
+
\mathcal{L}_r(X;\theta,\phi),
\qquad \lambda_d \ge 0.
\label{eq:scalarized}
\end{equation}

After training, let $(\hat\theta,\hat\phi)$ denote the fitted student
parameters. For a held-out observation $x\in X_{\mathrm{val}}$, MEDAL produces
the out-of-sample coordinate
\[
\hat z = e_{\hat\theta}(x),
\]
and the corresponding reconstruction
\[
\hat x = d_{\hat\phi}(\hat z)
=
d_{\hat\phi}(e_{\hat\theta}(x)).
\]
The held-out reconstruction error is then
\[
\ell_{\mathrm{val}}(x)
=
\left\|
x-d_{\hat\phi}(e_{\hat\theta}(x))
\right\|^2.
\]

\subsection{The design of MEDAL}
MEDAL departs from a conventional autoencoder in one critical respect. Its bottleneck is not freely learned but constrained to follow a prescribed teacher embedding (t-SNE, UMAP, or any other DR method), and reconstruction error is interpretable as a property of that embedding only when this constraint is enforced tightly. If the bottleneck deviates from the teacher coordinates, the resulting error conflates two distinct sources of loss---information discarded by the teacher and fitting error introduced by the student---with no way to separate them. We therefore require distillation to reach a predefined near-zero alignment threshold before any reconstruction-based comparison is made, and we treat this requirement as a prerequisite that drives the rest of the design.

That this prerequisite can be met at all rests on a known property of expressive neural networks. Given sufficient capacity, they can interpolate arbitrary targets on a finite training set \cite{yun2019smallrelunetworkspowerful}. In our setting, the teacher coordinates $z_i$ play the role of finite-sample targets for the encoder. In practice, we find that two to five hidden layers with a symmetric decoder suffices; very deep or very wide networks bring no benefit and only complicate optimization. The right depth and width are dataset-dependent, and we set them through architecture-capacity experiments, favoring the smallest configuration that achieves near-zero distillation reliably. Under a fixed parameter budget, high-dimensional inputs typically prefer narrower hidden layers, since the input-to-hidden and hidden-to-output maps already absorb much of the budget.

Nonlinear activations are what give the encoder enough expressiveness to reach near-zero distillation, but they are placed selectively rather than throughout. Hidden layers use ReLU \cite{nair2010relu} or SELU \cite{klambauer2017selfnormalizingneuralnetworks} activations, but the bottleneck mapping and the final reconstruction layer are linear. The linear bottleneck matters because teacher coordinates may be signed or scaled differently across DR methods, hence passing them through a nonlinearity would impose artificial range restrictions inconsistent with the teacher we seek to reproduce. The linear output layer matters because the reconstruction targets live in the full input space without bounded constraints.

Even with the right architecture, the distillation loss does not vanish on its own—it must be driven there by the penalty $\lambda_d$ in \autoref{eq:scalarized}. Too small a value leaves the bottleneck free to drift from the teacher, silently invalidating the scores. We sweep $\lambda_d$ on a logarithmic grid per dataset and adopt the value at which the alignment criterion is reliably met. The trade-off this incurs is mild. As \suppfigref{fig:lambda_tradeoff} shows, tightening $\lambda_d$ improves teacher matching by orders of magnitude while raising reconstruction error only marginally. Runs that fail the alignment criterion are discarded from downstream comparisons.

For comparisons across teacher methods or teacher hyperparameters, the student itself must also be held fixed. We use the same architecture and training protocol within each dataset so that differences in reconstruction error should reflect differences in the teacher manifold, not differences in student capacity. Some teacher embeddings are more difficult to distill than others, so we train with a generous maximum optimization budget but terminate early once the distillation loss reaches the predefined threshold. Embeddings that are easier to distill
therefore terminate training early, while harder ones are given the full budget
to reach the same threshold. Under this protocol, the final student bottleneck is
essentially indistinguishable from the teacher embedding, exemplified in
(\autoref{fig:schematic}C) where we place different teachers on a common basis for
reconstruction-based comparison.

A final question is whether reconstruction remains meaningful when the bottleneck is constrained to follow a prescribed manifold embedding. We address this
directly by using PCA as the teacher on MNIST and comparing reconstruction
curves across bottleneck ranks for four models: PCA itself, a vanilla
autoencoder trained only for reconstruction, linear MEDAL, and nonlinear
MEDAL. Here, linear MEDAL denotes a distilled model with linear activations
throughout the encoder and decoder, whereas nonlinear MEDAL uses nonlinear
hidden layers with linear maps into the bottleneck and out to the output. Both
MEDAL variants are distilled from the same PCA teacher. In \autoref{fig:mnist}D, linear MEDAL exactly matches the PCA reconstruction
curve across bottleneck ranks, as expected
\cite{BourlardKamp1988,BaldiHornik1989,plaut2018principalsubspacesprincipalcomponents}.
This confirms that the distillation constraint itself does not introduce
additional reconstruction artifacts. With nonlinear students, MEDAL distilled
from the same PCA teacher remains competitive with an unconstrained
autoencoder and can outperform PCA at moderate to large bottleneck ranks.
Together, these results show that enforcing teacher alignment need not degrade
reconstruction quality, supporting reconstruction error as a meaningful
quantitative signal after successful distillation. Additional architecture
experiments, optimization details, and stopping criteria are provided in
\suppsecref{sec:supp_methods} and \suppsecref{sec:recon-valid}.

\subsection{Downstream validation tasks enabled by MEDAL}

After a teacher embedding has been distilled into a 
encoder--decoder pair, MEDAL can be used as a validation layer for several
dimension-reduction workflows.

\paragraph{Task 1: Hyperparameter selection within a dimension reduction method.}
In practice, hyperparameters for methods such as t-SNE and UMAP are often chosen by visually inspecting the resulting two-dimensional embeddings.
MEDAL turns hyperparameter selection for a fixed dimension reduction method into
a held-out validation problem, the gold-standard type of validation in supervised learning. For any DR method, each
candidate hyperparameter setting is used to fit a teacher embedding on
$X_{\mathrm{tr}}$. The resulting embedding is then distilled into a student
autoencoder and evaluated on $X_{\mathrm{val}}$ by held-out reconstruction error.
Since student training may vary across random initializations, we repeat
distillation multiple times for each teacher embedding and retain only runs that
reach the predefined near-zero alignment threshold. Validation reconstruction
error is then summarized across successful runs, yielding a validation curve over
the candidate hyperparameter values. We select hyperparameters from this curve,
for example using the minimum or one-standard-error rule \cite{breiman1984classification}. As all candidates use the same method family and are distilled to the same alignment threshold under a shared student architecture, differences in validation reconstruction error can be attributed to the hyperparameter choice alone, providing a principled criterion for selection.

\paragraph{Task 2: Pointwise distortion analysis.}
Low-dimensional manifold embeddings are known to introduce distortion, because compressing high-dimensional structure into two or three dimensions can alter neighborhoods, distances, densities, and global organization, thereby affecting apparent clusters and downstream scientific conclusions \cite{meila2023manifold}.
Since reconstruction error is defined for each observation, MEDAL supports
pointwise distortion analysis. Observations with large reconstruction error
identify regions where the fitted low-dimensional geometry least faithfully
preserves the original high-dimensional structure. These pointwise errors can be
summarized by class, cell type, batch, domain, or other metadata to identify
systematic regions of embedding infidelity.

\paragraph{Task 3: Detection of distribution shift.}
MEDAL can also be used to detect when new observations are poorly represented by
a reference manifold. A teacher embedding is fitted and distilled on a reference
training distribution, and the learned encoder--decoder pair is then applied to
new observations, batches, or domains. If these observations have systematically
larger reconstruction errors than held-out reference observations, this indicates
that they are not well captured by the fitted low-dimensional geometry. Thus,
MEDAL provides a reconstruction-based diagnostic for distribution shift relative
to a reference embedding.

\paragraph{Task 4: Comparison across dimension reduction methods.}
Prior to this work, comparisons across dimension reduction methods lacked a shared validation principle, leaving method choice largely dependent on qualitative visual judgment. MEDAL provides a common reconstruction-based criterion for comparing fitted
embeddings across different dimension reduction methods. Within a dataset, each candidate method (e.g., t-SNE, UMAP, Isomap, PCA, and etc.) is fitted to $X_{\mathrm{tr}}$ using a fixed or optimized (from Task 1) hyperparameter configuration. Following the same distillation and validation protocol as Task 1, each embedding is evaluated by its validation reconstruction error. The method that achieves the lowest validation reconstruction error is preferred, as this indicates its geometry best preserves information on new observations. Because all methods are evaluated under the same student architecture, training procedure, and validation split, differences in reconstruction error reflect differences in the methods' underlying geometry rather than asymmetries in the evaluation protocol.
\section{Results} \label{sec:results}
We evaluate MEDAL as a validation framework through four case studies that progress from a controlled benchmark to biological applications. On the MNIST benchmark \cite{lecun1998gradient}, we demonstrate Tasks 1, 2, and 4 in a setting with known class structure. We then apply the same workflow to two single-cell RNA-seq datasets, on whole-animal Hydra from \cite{SeibertHydra} and on mouse neocortex from \cite{tasic2018shared}, showing that these tasks extend to single-cell genomics datasets where reconstruction error highlights biologically interpretable distortion. We further compare MEDAL with existing embedding-diagnostic methods for Tasks 1 and 4. Finally, using a single-cell RNA-seq macaque retina dataset with multiple experimental subjects from \cite{PENG20191222}, we demonstrate Task 3: reference-based detection and localization of distribution shift.

\subsection{Case study: MNIST benchmark}
We begin with the MNIST handwritten dataset \cite{lecun1998gradient} as a controlled proof of principle, using its known digit labels and well-understood visual structure to assess whether MEDAL produces quantitative signals that agree with intuitive embedding quality. In this benchmark setting, we use MEDAL to demonstrate held-out hyperparameter selection, comparison across dimension-reduction methods, and pointwise distortion analysis.

\paragraph{Task 1: Tuning \texttt{n\_neighbors} on UMAP} On MNIST visualizations, suboptimal hyperparameters do not necessarily produce embeddings that are visibly poor. Many training-set embeddings appear plausible by visual inspection, even when they differ in how well they preserve structure for held-out observations. We therefore ask whether MEDAL can distinguish embeddings that look similarly acceptable in sample but differ in out-of-sample information preservation.

Consistent with this validation curve (\autoref{fig:mnist}A), the MEDAL-selected embedding
($\texttt{n\_neighbors}=35$) balanced class separation with global coherence
(\autoref{fig:mnist}A). In contrast, a large-neighborhood embedding
($\texttt{n\_neighbors}=186$) compressed several digit classes into nearby
regions of the manifold, producing a visually plausible but less discriminative
embedding. A small-neighborhood embedding ($\texttt{n\_neighbors}=6$) produced
sharper apparent separation on the training set, but fragmented the manifold
and placed some held-out observations far from their corresponding class
regions (\autoref{fig:mnist}B). This behavior is consistent with overfitting to training structure, where normally the
training reconstruction loss is low, but validation and test reconstruction
loss increase sharply. By contrast, the selected embedding avoided both extremes. When held-out digits are embedded onto the manifold of the training data, they remained close to the appropriate class structure without excessive compression across classes, or excessive fragmentation within classes. Thus, even in a simple dataset such as MNIST, the most visually appealing training embedding is not necessarily the one that best preserves structure out of sample. 

\paragraph{Task 2: Distortion investigation} Next in \autoref{fig:mnist}A, the scattered or unstable points are precisely those estimated with high reconstruction error, indicating where the low-dimensional manifold is most poorly represented. Importantly, these high-error points are not arbitrary. As shown in \autoref{fig:mnist}E, nearby points on the same manifold can differ markedly in reconstruction quality. Low-error digits are typically reconstructed faithfully, whereas high-error digits tend to be more ambiguous, atypical, or poorly written, and their reconstructions are visibly blurrier and less accurate. Reconstruction error therefore carries semantic information about sample difficulty. It identifies which observations are naturally accommodated by the learned geometry and which are being forced into it with greater distortion.

\paragraph{Task 4: Comparing DR methods} On MNIST, we compare t-SNE, UMAP, spectral embedding, PHATE, and PCA, using each method at its selected hyperparameter where applicable (\autoref{fig:mnist}D; \suppfigref{fig:mnist-tuning}). UMAP at its selected hyperparameter achieved strong held-out reconstruction performance and produced a coherent manifold under this criterion. In contrast, methods such as PCA and spectral embedding retained less information in this setting. 

\subsection{Case study: Hydra single-cell RNA-seq dataset} 

We next evaluated MEDAL on a whole-animal \emph{Hydra} single-cell RNA-seq dataset, comprising approximately 25,000 cells spanning the major Hydra lineages \cite{SeibertHydra}. Hydra is a stringent testbed because it is both biologically well-characterized and developmentally dynamic; cells are continuously renewed, and the interstitial lineage in particular contains rich transitional structure \cite{scdeeds}. This makes Hydra well-suited for determining whether a quantitatively selected two-dimensional embedding truly recovers known biology rather than merely producing an appealing visualization.

\paragraph{Task 1: Perplexity tuning on t-SNE} 
The MEDAL-selected perplexity (\(499\)) recovered the major Hydra lineages---endoderm, ectoderm, and interstitial---and projected validation and test cells into the same lineage-specific regions of the training manifold (\autoref{fig:hydra}A,C). In comparison, a smaller perplexity (\(5\)) produced a sharper, more fragmented embedding, with finer local separation on the training set but a more diffuse and less stable organization for held-out cells (\autoref{fig:hydra}B). At the opposite extreme, a much larger perplexity (\(4999\)) over-smoothed the manifold, compressing biologically distinct populations into a more globally constrained layout and reducing separation across lineages. The selected perplexity avoided both extremes: it preserved broad lineage structure without excessive fragmentation at low perplexity or excessive compression at high perplexity. Thus, the embedding that appears most granular or most globally smooth on the training set is not necessarily the one that best preserves biological structure out of sample.

\paragraph{Task 2: Distortion analysis} In Hydra, high-error regions are repeatedly concentrated in the same biologically distinct cell populations, including interstitial nematocytes, basal disk epithelial cells, and male germline cells (\autoref{fig:hydra}E). We interpret this pattern cautiously, but its consistency is notable. These are not arbitrary outliers; rather, they correspond to cell populations that are either developmentally active or highly specialized, suggesting that two‑dimensional compression struggles to represent their complex transcriptomic variability \cite{SeibertHydra,davis1975histological,holstein2023hydra}.

Importantly, the same cell types remained highlighted across embeddings with different perplexities (\suppfigref{fig:hydra-biocoherence}), suggesting that MEDAL is detecting a stable biological signature of distortion rather than an artifact of a single visualization.

\subsubsection{Comparison with existing embedding diagnostics}

\begin{figure}
    \centering
    \includegraphics[width=\linewidth]{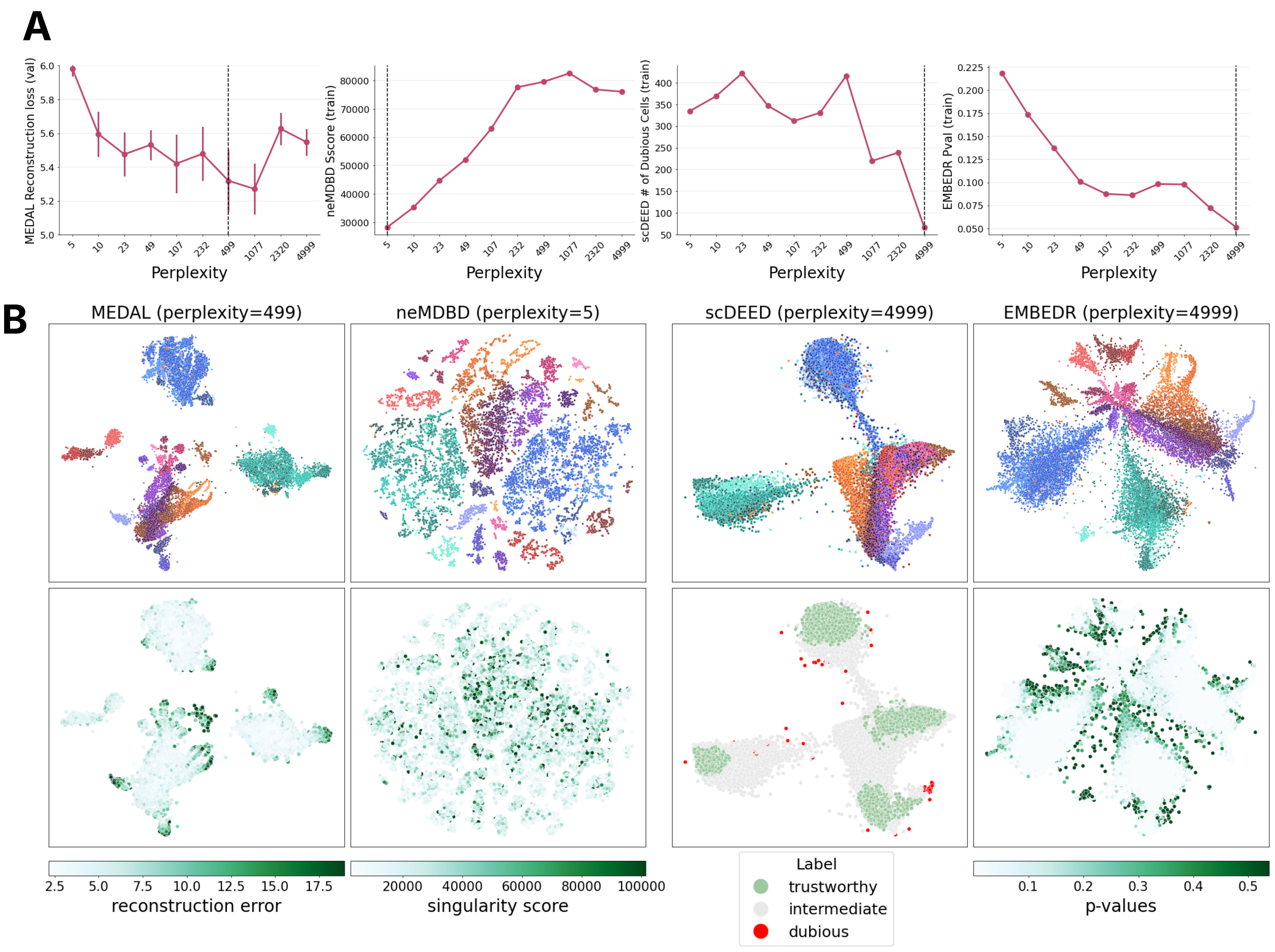}
    \caption{\textbf{Comparison of MEDAL with existing embedding diagnostics on
    a t-SNE embedding on Hydra (\cite{SeibertHydra}) single-cell RNA-seq dataset.}
    \textbf{A,} Hyperparameter-selection curves across t-SNE perplexities for MEDAL,
    neMDBD (\cite{liu2025assessingimprovingreliabilityneighbor}), scDEED (\cite{scdeeds}), and EMBEDR (\cite{johnson2022embedr}). Dashed vertical lines indicate the selected
    perplexity for each method. MEDAL selects perplexity 499 by held-out
    reconstruction loss, whereas neMDBD selects perplexity 5 and scDEED and EMBEDR
    select perplexity 4999. 
    \textbf{B,} Hydra single-cell RNA-seq embeddings at each method's selected perplexity. Top row:
    embeddings colored by reported cell type. Bottom row: pointwise diagnostic
    signals produced by each method: MEDAL reconstruction error, neMDBD singularity
    score, scDEED trustworthy/intermediate/dubious labels, and EMBEDR p-values.
    Because these methods are evaluated as deployed, this is a whole-system
    comparison: each method uses its own embedding implementation and optimization
    pipeline, so embeddings may differ even when the same perplexity is selected.
    MEDAL selects a more balanced embedding, retaining coherent lineage structure
    without favoring the extreme local or global regimes selected by competing
    diagnostics. Moreover, MEDAL's pointwise reconstruction error has a direct
    interpretation as local information loss in the original feature space, making
    the resulting distortion map easier to interpret than method-specific
    singularity scores, reliability labels, or p-values.}
    \label{fig:comparison}
\end{figure}

We next ask whether the distortion signal identified by MEDAL on Hydra was also recovered by existing embedding validation methods. We compared MEDAL with EMBEDR \cite{johnson2022embedr}, scDEED \cite{scdeeds}, and neMDBD \cite{liu2025assessingimprovingreliabilityneighbor}, three methods that assign pointwise measures of embedding reliability or distortion and can be used to guide hyperparameter selection. These competing diagnostics score an embedding generated from a single fit on the training data. MEDAL instead provides an out-of-sample criterion for hyperparameter selection, asking whether the selected embedding remains stable for held-out observations. On Hydra, this criterion selected an intermediate embedding that balanced local cell-state neighborhoods with global lineage organization while preserving consistent held-out placement. Competing methods often selected hyperparameters at the edge of the search grid (\autoref{fig:comparison}A), where very small perplexities can fragment the manifold and very large perplexities can compress distinct biological populations into a smoother but less discriminative layout. Although MEDAL's assessment is mediated by its learned encoder--decoder map, that mapping is precisely what enables the held-out test that the competing in-sample diagnostics lack.

MEDAL also produced a more interpretable pointwise distortion signal. High reconstruction error localized to biologically identifiable populations (\autoref{fig:hydra}E). On the other hand, competing pointwise signals were more diffuse and less directly interpretable (\autoref{fig:comparison}B). Thus, MEDAL reconstruction error served not only as a tuning criterion, but also as a biological diagnostic for cell populations whose transcriptomic structure was difficult to preserve in two dimensions.

To connect these qualitative patterns with external structure-preservation criteria, we evaluated each selected embedding using the Local Continuity Meta-Criterion (LCMC) for local neighborhood preservation and random triplet accuracy for broader distance-order preservation \cite{an2025consensusdimensionreductionmultiview}. These metrics are not intended to define a single ground truth for embedding quality, but they show that MEDAL selects embeddings that preserve recoverable high-dimensional signal while remaining competitive with standard local and global structure-preservation summaries (\supptabref{tab:supp_full_results}). At the same time, MEDAL provides a clear computational advantage. On the Hydra data set with t-SNE, MEDAL completed a full hyperparameter sweep substantially faster than
EMBEDR and scDEED, and neMDBD (\autoref{tab:hydra_tsne_time}). Supplementary runtime comparisons show that this advantage persists across datasets and teacher families \suppfigrangeref{fig:mnist-compare-tsne}{fig:tasic-compare-umap}. This per-seed efficiency makes repeated distillations and averaged validation curves feasible at the scale of large biological datasets, whereas repeatedly rerunning null-, permutation-, or geometry-based diagnostics across seeds and hyperparameter grids is considerably more costly.

\begin{table}[ht]
    \centering
    \caption{Computation time on the Hydra dataset using t-SNE as the teacher
    embedding. Times are reported in minutes as mean $\pm$ SE across 3 seeds.
    scDEED ($*$) was run after PCA preprocessing, following its recommended workflow.}
    \label{tab:hydra_tsne_time}
    \begin{tabular}{lc}
        \toprule
        \textbf{Method} & \textbf{Computation Time (min)} \\
        \midrule
        MEDAL & $43.08 \pm 4.81$ \\
        neMDBD \citep{liu2025assessingimprovingreliabilityneighbor} & $63.39 \pm 0.39$ \\
        EMBEDR \citep{johnson2022embedr} & $109.51 \pm 16.79$ \\
        scDEED$^*$ \citep{scdeeds} & $351.02 \pm 1.54$ \\
        \bottomrule
    \end{tabular}
\end{table}

\subsection{Case study: mouse neocortex single-cell RNA-seq dataset}

We next applied MEDAL to mouse neocortical single-cell RNA-seq data first reported
by \citet{tasic2018shared}. This dataset contains 23,822 cells collected from
two distant cortical areas, primary visual cortex (VISp) and anterior lateral
motor cortex (ALM). The original study resolved 133 transcriptomic cell types
and showed that most GABAergic types are shared across areas, whereas many
glutamatergic types are area-specific. The rich structures and extensive annotations, containing both major neuronal classes and a
smaller set of non-neuronal populations, make the \citet{tasic2018shared}
dataset a strong biological test case for manifold learning.

\paragraph{Task 1: Perplexity tuning on t-SNE.}
MEDAL selected a perplexity of \(53\), which preserved major cell-class organization and placed validation and test cells into the corresponding regions of the training manifold (\autoref{fig:tasic}A). The non-selected perplexities illustrate the same two failure modes observed in the previous case studies: overfitting to local substructure at very small perplexity (\(5\)), and over-compression at a large perplexity (\(1241\)) (\autoref{fig:tasic}B). This low-perplexity behavior is relevant to the original Tasic atlas setting, where the analysis emphasized fine transcriptomic subtype structure across many annotated cortical cell types \citep{tasic2018shared}. Our perplexity-\(5\) embedding represents an intentionally more local version of this regime, illustrating how visually sharp subtype separation on the training set can come at the cost of poorer organization for held-out cells. 

\paragraph{Task 2: Distortion analysis.}
In the selected t-SNE embedding, as in the Hydra dataset, high-error regions were
not randomly distributed across the manifold. Instead, they concentrated in
specific broad cell classes, especially non-neuronal and endothelial
populations, whereas the major neuronal classes---GABAergic and glutamatergic
neurons---were reconstructed more faithfully (\autoref{fig:tasic}D). This pattern is biologically
plausible given the dominant geometry of the atlas is largely organized by neuronal
heterogeneity, with many resolved GABAergic and glutamatergic subclasses, while
non-neuronal populations form smaller and transcriptionally distinct groups. 
In a global two-dimensional embedding of this taxonomy, smaller or more
specialized populations may be harder to represent without loss if their
variation is not aligned with the dominant structure of the embedding. This distortion signal was also consistent beyond the selected t-SNE embedding.
In \suppfigref{fig:tasic-biocoherence}, the same broad pattern appeared for a small-perplexity t-SNE teacher and for a UMAP teacher—higher reconstruction error again localized disproportionately to non-neuronal and
endothelial regions.

\paragraph{Task 4: Comparison across dimension-reduction methods}
We observe consistent results as in the MNIST benchmark previously. t-SNE at its selected
perplexity achieved the lowest held-out reconstruction loss among the methods
considered, followed by UMAP, while spectral embedding and PCA retained
substantially less information (\autoref{fig:tasic}C).

\subsection{Case study: evaluate macaque retina single-cell RNA-seq data on Task 3}

Finally, we used the macaque retina dataset \cite{PENG20191222} to illustrate Task 3: detecting distribution
shift relative to a fixed reference manifold. Once a teacher manifold has been distilled, new observations can be mapped into
a fixed reference space and evaluated by reconstruction error. This allows MEDAL
to ask whether incoming observations are supported by the geometry of the reference
distribution, rather than simply forcing them into an existing low-dimensional
visualization.

We applied this idea to a single-cell RNA-seq dataset of macaque retinal cells
collected across multiple subjects, denoted macaque 1
(M1), macaque 2 (M2), and macaque 3 (M3). This dataset provides a useful test
case because retinal cell identity is strongly structured, but inter-subject and batch variation can substantially affect joint single-cell embeddings. Multi-sample single-cell analyses often require explicit batch correction or data integration to avoid confounding biological cell-type structure with technical or donor-specific effects
\citep{haghverdi2018batch,stuart2019comprehensive,korsunsky2019fast,
lopez2018deep,tran2020benchmark}. Here, we do not attempt to pre-assess or correct these
effects; instead, we use subject-level variation to test whether MEDAL reconstruction error can detect when new cells are poorly supported by a fitted reference manifold.

In the full macaque retina dataset, a joint t-SNE embedding appears coherent
when colored by cell type, but coloring the same embedding by subject reveals
strong subject-level structure across the manifold (\autoref{fig:macaque}A).
This motivates a reference-mapping question: if a manifold embedding is learned from one
subject, can we detect  new cells belonging to another subject that no longer conform to the reference
geometry?

To test this, we trained MEDAL on M1 training cells and used the learned encoder
to project two held-out datasets onto the M1 reference manifold, an
in-distribution batch of M1 test cells and a mixed batch containing M1 test
cells together with M2 cells. MEDAL did not use subject labels when embedding or
scoring these cells. For the in-distribution M1 test batch, cells were placed
neatly into the existing M1 geometry, and reconstruction error remained low and
spatially diffuse across the manifold. In contrast, in the mixed batch, the M1
test cells remained aligned with the reference structure, whereas M2 cells
occupied a displaced region of the embedding and showed a spatially coherent
increase in reconstruction error (\autoref{fig:macaque}B). Thus, MEDAL detects
out-of-distribution structure by displacing OOD cells relative to the reference manifold and assigning them higher reconstruction errors,
while stably embedding in-distribution cells into the expected geometry.

We then asked which cell types contributed most strongly to this shift. Comparing
cell-type-specific reconstruction errors between M1 test cells and M2 cells on
the M1 reference manifold showed that the increase was not uniform across the
atlas (\autoref{fig:macaque}C). Although M2 cells generally had higher
reconstruction error than the M1 baseline, the largest shifts occurred in
microglia, endothelial cells, and pericytes. These non-neuronal populations are
biologically plausible sources of subject-level mismatch because they often show
stronger inter-individual, immune, vascular, or environmental variation than
major neuronal cell classes. MEDAL therefore not only detects the presence of a
distribution shift, but also localizes the mismatch to specific regions of the
manifold and specific cell populations.

Additional analyses in the Supplementary Materials show that this signal is not
specific to the M1-versus-M2 comparison. When M3 cells are projected onto the
M1 reference manifold, reconstruction error again increases relative to the M1
baseline and remains concentrated in specific cell populations
(\suppfigref{fig:macaque-m3-vs-m1}). Reversing the reference subject by training
MEDAL on M2 cells yields a similar pattern for M1 cells projected onto the M2
reference manifold (\suppfigref{fig:macaque2}). 
MEDAL therefore augments out-of-sample embedding with a quantitative reliability signal for detecting and localizing distribution shift.

\section{Discussion}

Low-dimensional embeddings are now used as scientific summaries of complex high-dimensional data, yet the dominant workflow for nonlinear dimension reduction remains difficult to validate quantitatively. In this work, we introduced MEDAL as a model-agnostic framework for turning a fitted embedding into a reusable statistical object. Given a teacher embedding from any DR technique, MEDAL trains a constrained autoencoder whose encoder matches the teacher embedding and whose decoder reconstructs the original input. This distillation step converts an otherwise static manifold embedding into a model with an out-of-sample map, an approximate inverse, and a pointwise reconstruction-based measure of information loss.

Across benchmark and scientific datasets, this framework enabled several forms of validation that are typically unavailable in standard unsupervised workflows. MEDAL makes it possible to evaluate candidate embeddings on held-out observations, select hyperparameters using reconstruction error rather than visual inspection alone, compare different dimension reduction methods on a shared input-space scale, and localize regions of the embedding where low-dimensional compression is least faithful. In the biological case studies, the resulting pointwise distortion scores highlighted coherent cell populations and distribution-shifted samples, suggesting that MEDAL can be used not only to choose an embedding, but also to audit the resulting scientific conclusions.

Practically, MEDAL is designed to fit into the workflow that many scientists already use \cite{chang2025unsupervisedmachinelearningscientific}, while making the interpretation of that workflow more rigorous. A typical analysis fits a familiar method such as t-SNE or UMAP, where design decisions are determined by visual inspection of the resulting embedding, and uses that embedding as the basis for downstream claims. MEDAL inserts a validation step before the embedding is used for downstream analyses such as clustering, cell-type annotation, trajectory inference, differential expression, batch-effect assessment, or rare-population detection. With this increased confidence in embeddings, and hence, downstream scientific discovery, MEDAL is a significant contribution towards helping resolve the crisis in scientific reproducibility and replicability \cite{national2019reproducibility}.

MEDAL's role here, however, should be understood as validation of a \emph{chosen teacher embedding}; it does not give an absolute guarantee that an embedding is scientifically correct. In fact, MEDAL inherits the assumptions and failures of the teacher. If the teacher removes biologically meaningful variation or emphasizes spurious structure, MEDAL will faithfully distill that geometry. Likewise, reconstruction error measures information loss relative to the given teacher manifold and student architecture. For this reason, MEDAL is best viewed as one quantitative validation criterion within a broader scientific workflow, to be considered alongside domain knowledge and complementary metrics of local or global structure preservation.

Several directions for future work follow naturally from this framework. First, MEDAL raises theoretical questions about when reconstruction error provides a reliable proxy for information loss, and how this criterion relates to teacher geometry, student capacity, neighborhood preservation, stability, and generalization in unsupervised learning \citep{belkin2003laplacian,venna2006local,lee2009quality,bengio2013representation,bousquet2002stability}. 
Second, although MEDAL currently uses an MLP encoder--decoder as the student model, the same distillation principle could be paired with attention-based or transformer architectures that better capture structured dependencies among genes, features, or modalities \citep{vaswani2017attention,theodoris2023transfer,yang2022scbert,cui2024scgpt}.
Third, MEDAL could be extended from single-modality embeddings to multimodal, multi-sample, and spatial datasets, where validation must account for multiple molecular measurements, experimental conditions, tissue sections, or spatial neighborhoods rather than a single feature matrix \citep{gayoso2021joint,ashuach2023multivi,boyeau2025deep,dong2022deciphering,biancalani2021deep,kleshchevnikov2022cell2location}. 
Finally, connecting MEDAL to reference-atlas mapping and downstream analyses would place reconstruction-based validation directly within the scientific workflows that rely on low-dimensional representations for clustering, trajectory inference, annotation transfer, perturbation analysis, and disease-state discovery \citep{hao2021integrated,kang2021efficient,lotfollahi2022mapping,michielsen2023single,lotfollahi2019scgen,lotfollahi2023predicting}.

MEDAL reframes dimension reduction from a purely \emph{visual} endpoint into a quantitatively testable \emph{modeling} step, and in doing so suggests a broader way to practice unsupervised learning. By distilling fitted embeddings into reusable encoder--decoder models, MEDAL brings the rigor and reproducibility of model selection and validation into a setting where such tools are often limited, treating latent representations not only as illustrations of structure but also as \emph{hypotheses about it}. For statisticians and domain scientists alike, MEDAL points toward a broader research direction in which ideas from statistical inference, including generalization, stability, and structured model comparison, can be brought to bear on representation learning as an inferential activity. This perspective narrows the long-standing asymmetry between supervised and unsupervised methodology by making learned representations---even without labels---testable claims about structure rather than qualitative visual summaries.

\section*{Acknowledgments}
The authors acknowledge funding from NSF DMS-2516872 and also thank Matthew Shen for developing an early prototype of the idea. 

\section*{Data and Code Availability}

All datasets analyzed in this study are publicly available. The mouse neocortex
single-cell RNA-seq data \cite{tasic2018shared} are available through the \href{https://brain-map.org/our-research/cell-types-taxonomies/cell-types-database-rna-seq-data}{Allen Institute Cell Types Database RNA-seq data portal} and at Gene Expression Omnibus (GEO) accession \href{https://www.ncbi.nlm.nih.gov/geo/query/acc.cgi?acc=GSE115746}{GSE115746}.
The Hydra single-cell RNA-seq data \cite{SeibertHydra} are available under GEO accession 
\href{https://www.ncbi.nlm.nih.gov/geo/query/acc.cgi?acc=GSE121617}{GSE121617}.
The macaque retina single-cell RNA-seq data \cite{PENG20191222} are available from GEO under accession \href{https://www.ncbi.nlm.nih.gov/geo/query/acc.cgi?acc=GSE118546}{GSE118546}.
For the APOGEE experiments, we used the processed globular cluster data and
train--test splits released from \citet{chang2025unsupervisedmachinelearningscientific}, available
from \cite{tiffany_tang_2026_19194851}.
Additional documentation for the APOGEE preprocessing workflow is available at
\url{https://dataslingers.github.io/unsupervised-workflow-astro/}.

Code implementing MEDAL and scripts for reproducing the main analyses are
available at \url{https://github.com/DataSlingers/MEDAL} via \cite{irene_chang_2026_20347573}. Pre-processing scripts will be provided from the publicly available source data.

\begin{figure}
    \centering
    \includegraphics[width=\columnwidth, height=0.96\textheight, keepaspectratio]{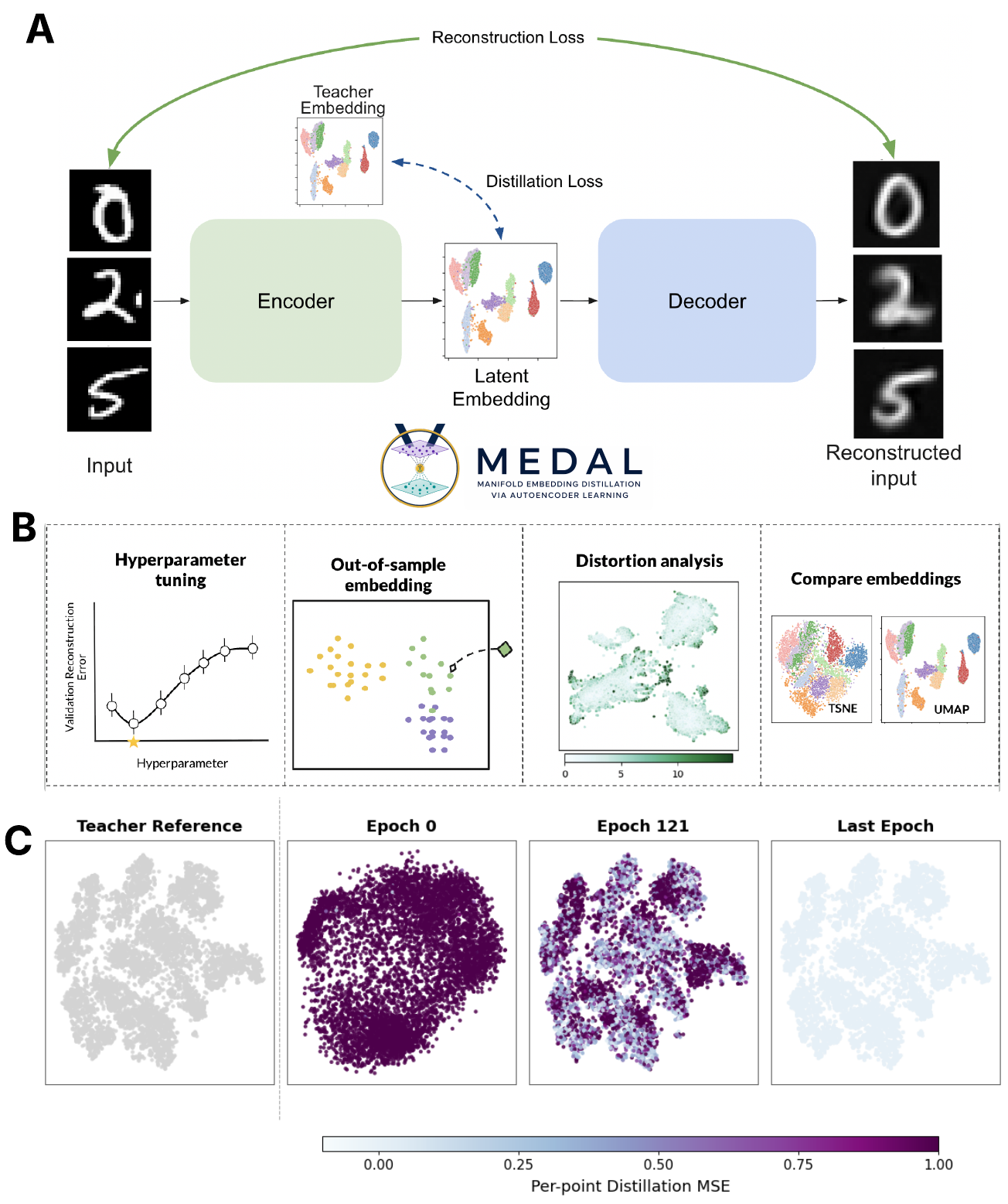}
    \caption{}
    \label{fig:schematic}
\end{figure}

\begin{figure*}
    \ContinuedFloat
     \caption{\textbf{MEDAL inputs a fitted manifold embedding and distills this into a reusable model that permits quantitative validation of the embedding.}
\textbf{A,} MEDAL distills a teacher embedding into a constrained autoencoder by jointly optimizing a distillation loss that aligns the bottleneck with the teacher manifold and a reconstruction loss that preserves input-space information.
\textbf{B,} Once distilled, the learned encoder--decoder supports held-out evaluation, hyperparameter selection, out-of-sample embedding, pointwise distortion analysis, and comparison across multiple manifold embeddings under a shared protocol.
\textbf{C,} Distillation progression on MNIST data toward a t-SNE teacher embedding, showing convergence from random initialization to near-exact teacher matching.}
\end{figure*}

\begin{figure*}
    \centering
    \includegraphics[width=\columnwidth]{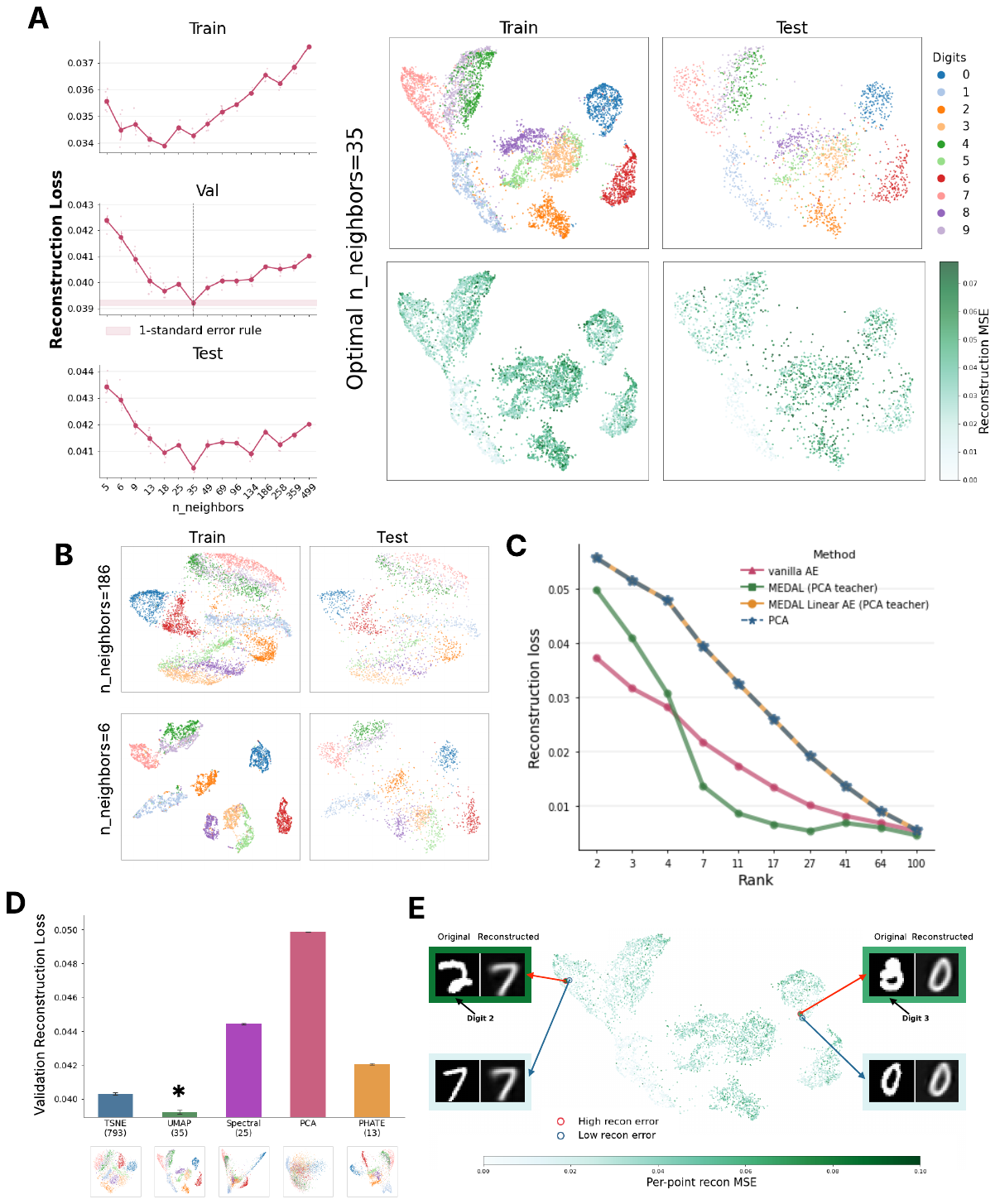}
    \caption{}
    \label{fig:mnist}
\end{figure*}

\begin{figure*}
    \ContinuedFloat
    \caption{\textbf{MEDAL enables quantitative validation, comparison, and
distortion analysis on MNIST data.}
\textbf{A,} UMAP hyperparameter selection using held-out reconstruction error.
For each value of \texttt{n\_neighbors}, a fitted UMAP teacher embedding was
distilled into a MEDAL student, and reconstruction loss was evaluated on train,
validation, and test splits. The selected model
(\texttt{n\_neighbors}=35; dashed line) balances class separation with global
coherence. The corresponding train and test embeddings are shown colored by
digit label and by pointwise reconstruction error.
\textbf{B,} Representative non-selected UMAP embeddings. A large-neighborhood
embedding (\texttt{n\_neighbors}=186) compresses digit classes into nearby
regions, whereas a small-neighborhood embedding
(\texttt{n\_neighbors}=6) fragments the manifold and places some held-out
observations far from their corresponding class regions.
\textbf{C,} Reconstruction loss as a function of bottleneck rank for PCA, vanilla autoencoders, and MEDAL distilled from a PCA teacher. Linear MEDAL recovers the PCA reconstruction curve, as predicted by classical results linking linear autoencoders to principal subspaces \citep{BourlardKamp1988,BaldiHornik1989,plaut2018principalsubspacesprincipalcomponents}, showing that MEDAL does not introduce artifacts through the distillation constraint. Nonlinear MEDAL remains competitive with an unconstrained autoencoder at larger ranks, further supporting reconstruction loss as a valid information preservation criterion under teacher-matching constraints.
\textbf{D,} Held-out reconstruction loss provides a shared criterion for
comparing dimension-reduction methods. Each nonlinear teacher is evaluated at
its selected hyperparameter, with example embeddings shown below each bar.
\textbf{E,} Pointwise reconstruction error localizes distortion within the
selected UMAP manifold. Low-error observations are reconstructed faithfully,
whereas high-error observations often correspond to ambiguous or atypical digit
images whose reconstructions are visibly degraded.}
\label{fig:mnist}
\end{figure*}

\begin{figure}
    \centering
    \includegraphics[width=\columnwidth, height=0.95\textheight, keepaspectratio]{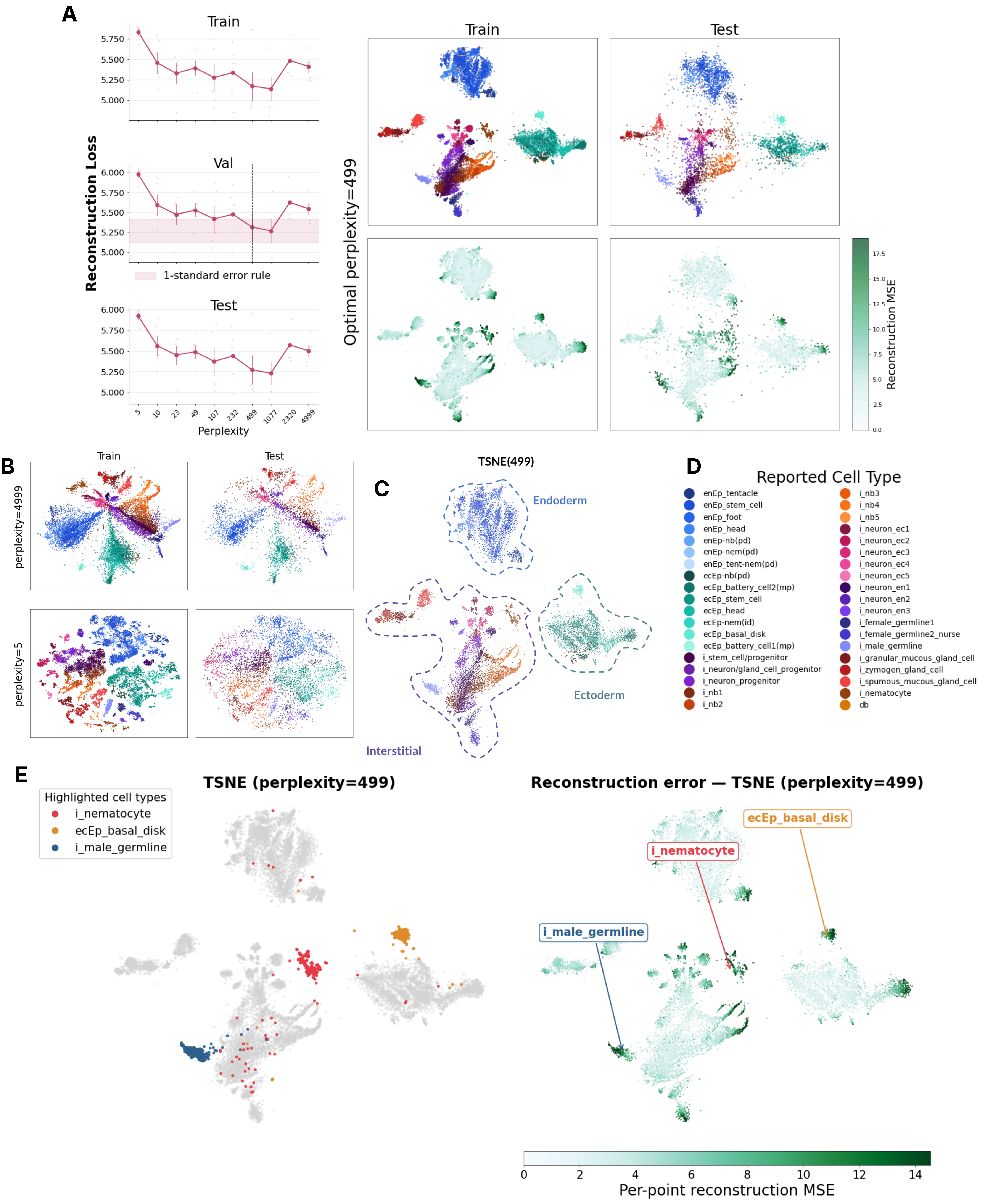}
    \caption{}
    \label{fig:hydra}
\end{figure}

\begin{figure*}
    \ContinuedFloat
    \caption{\textbf{MEDAL selects a biologically coherent embedding of the whole-animal Hydra single-cell RNA-seq dataset \cite{SeibertHydra} and localizes cell-type-specific distortion.}
    \textbf{A,} MEDAL hyperparameter tuning for t-SNE teachers on the Hydra single-cell atlas. Reconstruction loss was evaluated across t-SNE perplexities on train, validation, and test splits. The selected perplexity, chosen by the validation curve using the one-standard-error rule, was \(499\). The corresponding training and test embeddings preserve the major Hydra lineages, and per-point reconstruction error highlights localized regions of high distortion.
    \textbf{B,} Comparison with unselected t-SNE perplexities. A very large perplexity (\(4999\)) produces a more globally compressed embedding in which distinct populations are drawn closer together, whereas a very small perplexity (\(5\)) fragments the manifold and generalizes poorly to held-out cells.
    \textbf{C,} The MEDAL-selected embedding recovers the three major Hydra lineages: endoderm, ectoderm, and interstitial cells.
    \textbf{D,} Reported cell-type annotations used to color the embeddings.
    \textbf{E,} Pointwise reconstruction error reveals biologically structured manifold embedding distortion. High-error regions are enriched in specific cell populations, including interstitial nematocytes, ectodermal basal disk epithelial cells, and male germline cells, suggesting that these dynamic or specialized populations are less faithfully represented by the two-dimensional manifold.}
\end{figure*}

\begin{figure*}
    \centering
    \includegraphics[width=\linewidth, height=0.95\textheight, keepaspectratio]{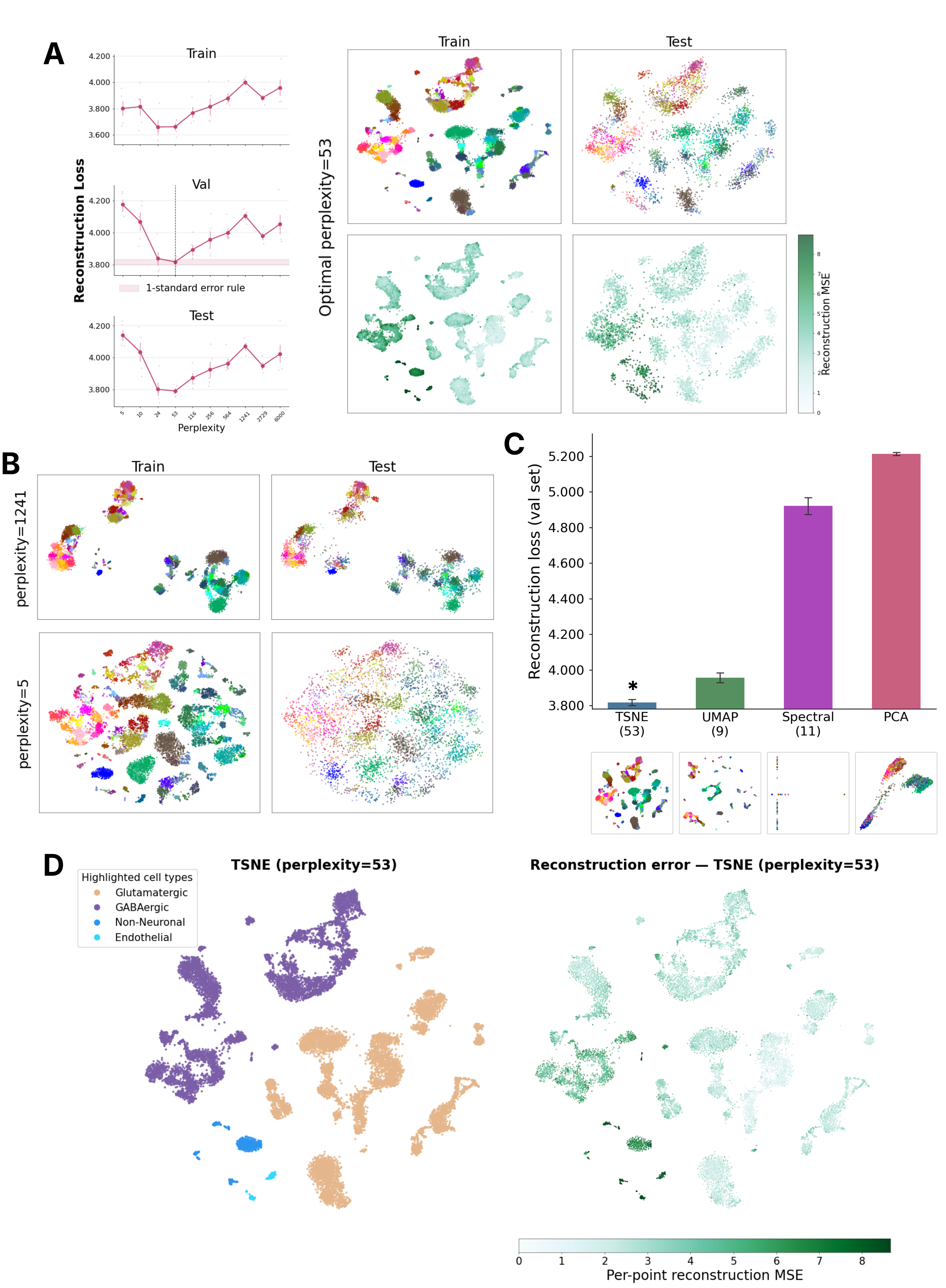}
    \caption[]{}   
    \label{fig:tasic}
\end{figure*}

\begin{figure*}
    \ContinuedFloat
    \caption{\textbf{MEDAL selects a manifold embedding for mouse neocortex single-cell RNA-seq (\cite{tasic2018shared}) and reveals cell types poorly represented on the manifold.}
    \textbf{A,} MEDAL hyperparameter tuning for t-SNE teachers on the neocortex atlas. Reconstruction loss was evaluated across t-SNE
    perplexities on train, validation, and test splits. The selected perplexity was
    \(53\). The corresponding training and test embeddings preserve major
    transcriptomic structure, and the per-point reconstruction-error maps highlight
    localized regions of higher distortion.
    \textbf{B,} Comparison with unselected t-SNE perplexities. A large perplexity
    (\(1241\)) produces a more globally smoothed embedding, whereas a small
    perplexity (\(5\)) produces sharper local subclusters on the training set but a
    more diffuse organization for held-out cells. These examples illustrate that
    visually detailed training embeddings do not necessarily preserve structure out
    of sample.
    \textbf{C,} Comparison across dimension-reduction teachers under the MEDAL
    validation protocol. Each method is shown at its selected hyperparameter, with
    validation reconstruction loss summarized across repeated distillations. t-SNE
    at perplexity \(53\) achieved the lowest held-out reconstruction loss among the
    methods considered, followed by UMAP, whereas spectral embedding and PCA
    retained less information in this setting.
    \textbf{D,} Pointwise distortion analysis for the selected t-SNE embedding.
    The left panel highlights broad annotated cell classes, including glutamatergic
    neurons, GABAergic neurons, non-neuronal cells, and endothelial cells. The
    right panel shows MEDAL per-point reconstruction mean squared error on the same
    embedding. High reconstruction error is concentrated disproportionately in
    non-neuronal and endothelial populations, suggesting that these smaller and
    transcriptionally distinct populations are less faithfully represented by the
    two-dimensional manifold.}
\end{figure*}

\begin{figure*}
    \centering
    \includegraphics[width=\linewidth, height=0.9\textheight, keepaspectratio]{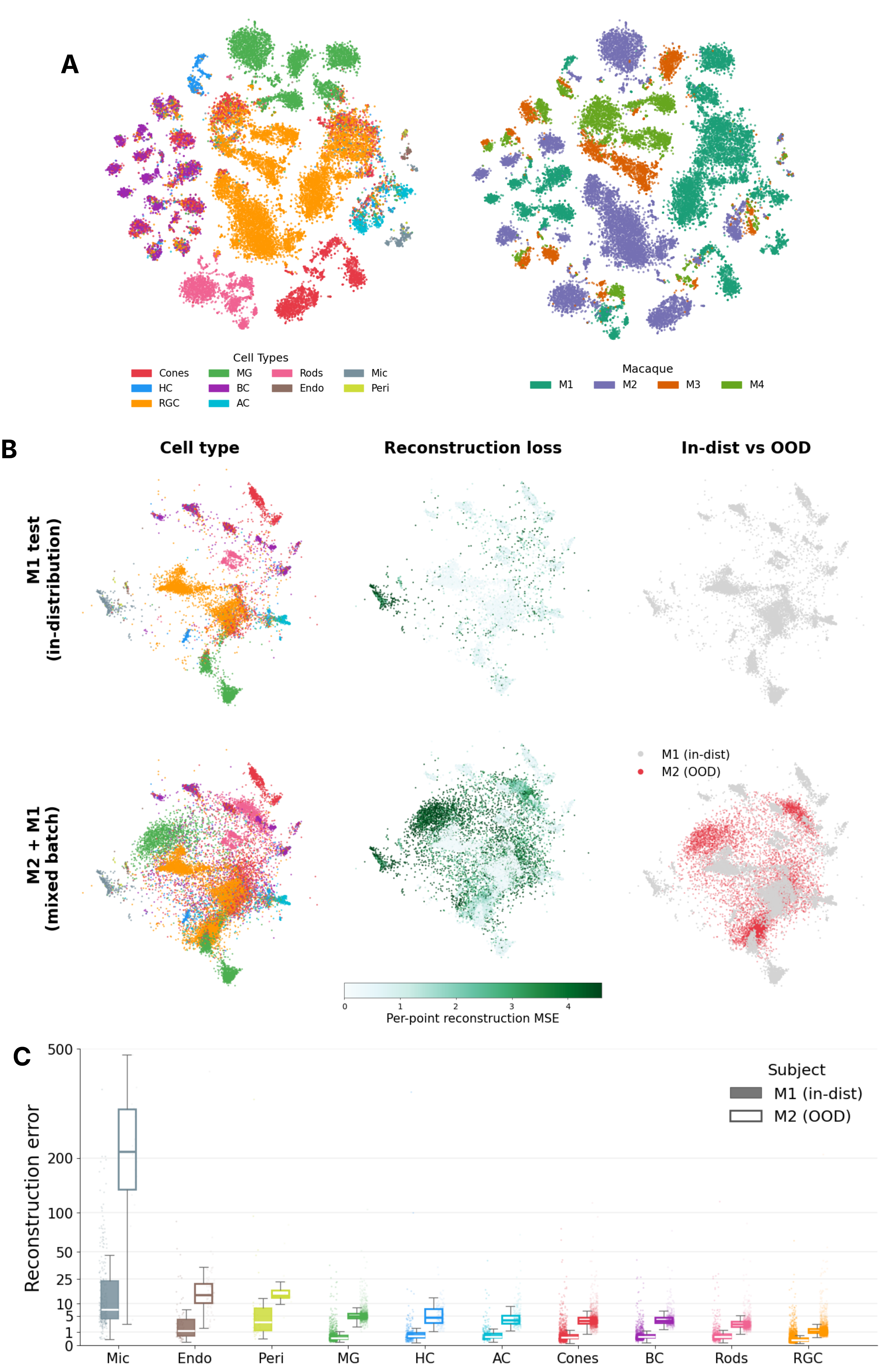}
    \caption[]{}   
    \label{fig:macaque}
\end{figure*}

\begin{figure*}
    \ContinuedFloat
    \caption{\textbf{MEDAL detects subject-level distribution shift in macaque retina single-cell RNA-seq (\cite{PENG20191222}) by embedding new cells into a fixed reference manifold.}
    \textbf{A,} Joint t-SNE embedding of macaque retinal cells from multiple subjects. 
    The embedding appears organized by cell type when colored by annotated retinal
    cell class, but coloring the same embedding by subject reveals substantial
    subject-level structure, indicating that biological cell identity and
    inter-subject variation are entangled in the joint visualization.
    \textbf{B,} MEDAL reference mapping using M1 as the training distribution. MEDAL
    was trained on M1 training cells and used to embed either held-out M1 test cells
    or a mixed batch containing M1 test cells and M2 cells into the same M1 reference
    manifold. For the in-distribution M1 test batch, cells are placed coherently
    within the reference geometry and show low reconstruction error. In the mixed
    batch, M1 test cells remain aligned with the reference structure, whereas M2
    cells occupy a displaced region of the embedding and exhibit elevated,
    spatially coherent reconstruction error. MEDAL therefore identifies
    out-of-distribution cells without using subject labels during embedding or
    scoring.
    \textbf{C,} Cell-type-specific reconstruction error for M1 test cells and M2
    cells projected onto the M1 reference manifold. Reconstruction error is higher
    for M2 cells across multiple cell types, with the strongest shifts in microglia,
    endothelial cells, and pericytes. This indicates that the subject-level mismatch
    detected by MEDAL is not only global, but is concentrated in specific retinal
    cell populations.}
\end{figure*}
\clearpage
\newpage

\bibliographystyle{plainnat}
\bibliography{main_bib}

\clearpage
\newpage

\appendix 
\setcounter{section}{0}
\renewcommand{\thesection}{S\arabic{section}}

\setcounter{subsection}{0}
\renewcommand{\thesubsection}{\thesection.\arabic{subsection}}

\setcounter{table}{0}
\renewcommand{\thetable}{S\arabic{table}}
\renewcommand{\tableautorefname}{Supplementary Table}

\setcounter{figure}{0}
\renewcommand{\thefigure}{S\arabic{figure}}
\renewcommand{\figureautorefname}{Supplementary Figure}

\section{Datasets}
\paragraph{MNIST} The MNIST dataset consists of grayscale images of handwritten digits (0–9). For this study, we use a subset of 10,000 images, each with spatial resolution $28 \times 28$. Images are flattened into 784-dimensional feature vectors, and pixel intensities are normalized to lie in $[0,1]$. MNIST serves as a low-dimensional, low-noise benchmark for evaluating representation learning methods under controlled conditions.

\paragraph{Hydra} The Hydra single-cell RNA-seq dataset represents the first comprehensive cell atlas of the adult Hydra polyp \cite{SeibertHydra,scdeeds}. The dataset contains transcriptomes from $n = 25{,}052$ cells profiled using Drop-seq, with expression measurements for 33,391 genes. Data are processed, normalized, and scaled using the Seurat pipeline, and curated cluster annotations are provided. To reduce computational burden and facilitate overparameterized representation learning and distillation, we project the data onto the top 500 principal components (PCs), which capture the dominant biological variation while substantially reducing dimensionality.

\paragraph{Neocortex} This dataset comprises single-cell transcriptomic profiles of the adult mouse neocortex, as described by \cite{tasic2018shared}. It includes approximately 23,822 cells sampled from the primary visual cortex (VISp) and anterior lateral motor cortex (ALM). Expression data are processed and scaled using Seurat, with cell-type annotations provided. The dataset captures a rich diversity of neuronal and non-neuronal populations, including glutamatergic neurons, GABAergic neurons, and multiple glial cell types. To manage computational complexity and enable overparameterized modeling, we retain the top 1,000 principal components (PCs), preserving major transcriptional structure while reducing dimensionality.neuronal types. 

\paragraph{Macaque} TThe Macaque dataset is a single-cell transcriptomic atlas of the primate retina, profiling both foveal and peripheral regions of Macaca mulatta, as described by \cite{PENG20191222}. The dataset comprises approximately 48,000 cells collected across a total of three individual macaques, providing a molecular characterization of all major retinal cell classes, including photoreceptors (rods and cones), bipolar cells (BC), horizontal cells (HC), amacrine cells (AC), retinal ganglion cells (RGC), and Müller glia (MG). The presence of multiple biological donors introduces natural inter-individual variability, enabling evaluation of our framework’s ability to capture and generalize across domain shifts arising from subject-specific transcriptional differences.

\paragraph{Darmanis} 
The Darmanis dataset is a single-cell RNA-seq dataset profiling the adult human brain, as described by \cite{DARMANIS20171399}. It comprises transcriptomic measurements from thousands of individual cells sampled across multiple brain regions, including the cerebral cortex, hippocampus, and other areas. The dataset captures a diverse range of major neural and non-neural cell types, including neurons, astrocytes, oligodendrocytes, microglia, and endothelial cells, with expression profiles reflecting both cell-type identity and regional heterogeneity. As a high-dimensional, noise-rich biological dataset with complex cellular composition, the Darmanis dataset provides a challenging benchmark for representation learning and manifold-based methods, particularly for evaluating robustness to biological variability and subtle cell-type–specific transcriptional programs.

\paragraph{PANCAN} \cite{gene_expression_cancer_rna_seq_401} The PANCAN dataset used in this study is a bulk RNA-sequencing dataset derived from The Cancer Genome Atlas (TCGA) pan-cancer project, which profiled trans-criptome-wide gene expression across multiple tumor types as part of a large effort to characterize molecular aberrations across cancers. The subset accessed via the UCI Machine Learning Repository consists of 801 tumor samples spanning multiple cancer types—including breast invasive carcinoma (BRCA), kidney renal cell carcinoma (KIRC), colon adenocarcinoma (COAD), lung adenocarcinoma (LUAD), and prostate adenocarcinoma (PRAD)**—with no missing values. Each sample is represented by 20,531 gene expression features measured on the Illumina HiSeq platform, corresponding to real-valued normalized RNA-Seq counts for individual genes. This results in a wide data matrix (many more features than samples) typical of transcriptomic datasets, with high dimensionality and biological heterogeneity that pose substantial challenges for representation learning, overfitting control, and generalization in downstream modeling. Because of these characteristics, PANCAN serves as a demanding benchmark for evaluating dimensionality reduction and autoencoder architectures in the presence of complex, noise-rich biological signals.

\paragraph{APOGEE} This astronomical dataset consists of approximately 8,000 stellar spectra, each represented by flux measurements across 19 wavelength bands \cite{apogeedata}. The data and preprocessing pipeline follow the methodology described in \cite{chang2025unsupervisedmachinelearningscientific}. This dataset provides a structured, low-dimensional scientific benchmark with continuous-valued features, enabling evaluation of representation learning methods in a physical sciences context.

\paragraph{General preprocessing pipeline} Each dataset in useis partitioned into a training set (80\%) and a test set (20\%) to evaluate the model's ability to generalize to unseen data point.

\section{Methods}
\label{sec:supp_methods}

This supplement provides additional methodological details for MEDAL. The main
text introduces MEDAL as a constrained autoencoder framework that distills a
fitted teacher embedding into a reusable parametric model. Here we expand on
four points that are central to the method: (i) the practical operating regime
in which near-zero empirical distillation is achieved, (ii) why such alignment
is feasible in principle, (iii) why reconstruction remains meaningful after
enforcing distillation, and (iv) the held-out evaluation protocol used for
hyperparameter selection and method comparison.

\subsection{Practical operating regime for near-zero distillation}
\label{sec:supp_optimization}
A central practical question for MEDAL is how to train the student so that
empirical distillation loss is driven to near zero while reconstruction remains
useful. In our experiments, this operating regime is determined primarily by the
distillation weight $\lambda_d$, the student architecture, the encoder
activation design, and an adaptive optimization schedule.

\paragraph{$\lambda_d$ analysis.}
\label{app:lambda-d}

\begin{figure}
    \centering
    \includegraphics[width=0.9\linewidth]{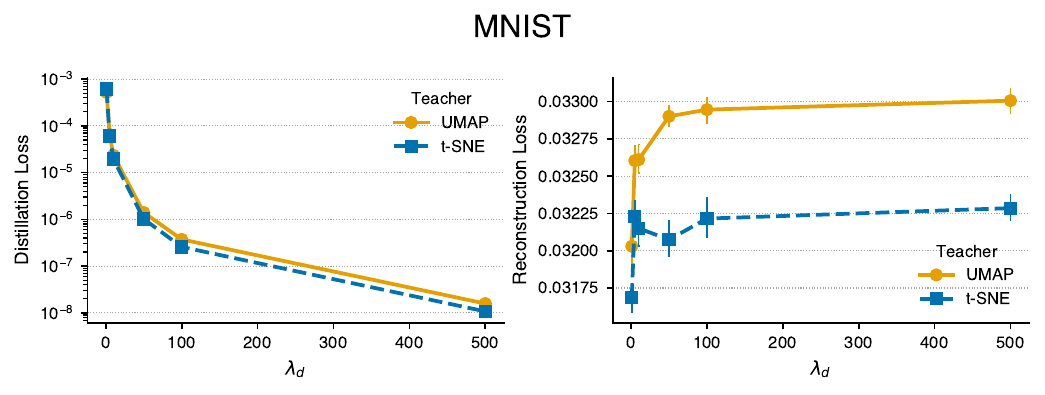}\\[4pt]
    \includegraphics[width=0.9\linewidth]{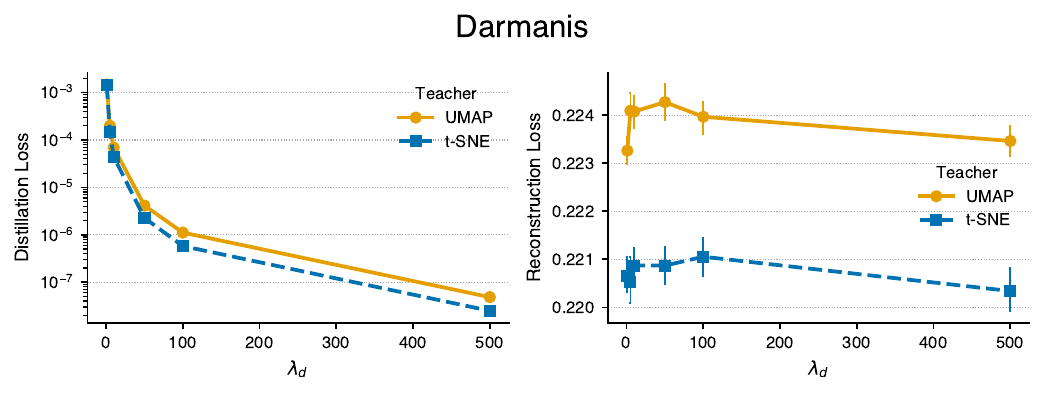}\\[4pt]
    \includegraphics[width=0.9\linewidth]{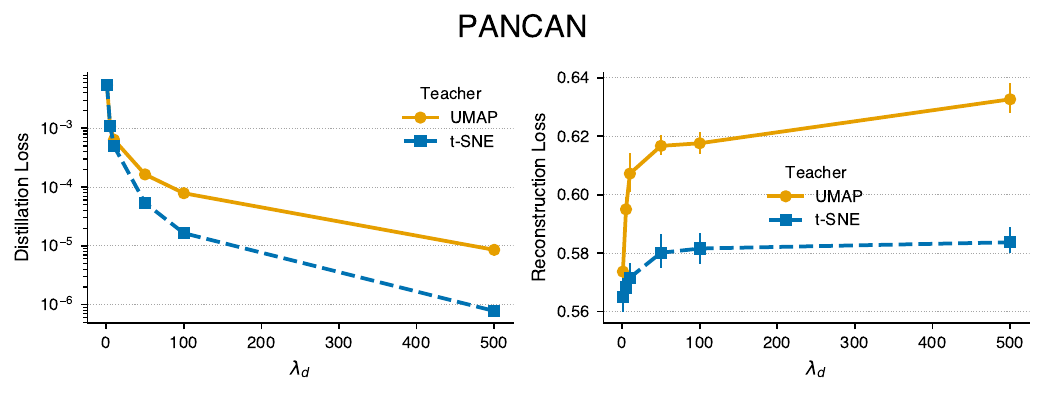}
    \caption{\textbf{Effect of the distillation weight $\lambda_d$ on teacher matching and
reconstruction.}
Final distillation loss and reconstruction loss are shown across increasing
values of $\lambda_d$ for MNIST, Darmanis, and PANCAN. Larger values of
$\lambda_d$ place greater weight on matching the teacher embedding in
\autoref{eq:scalarized}. Across datasets, increasing $\lambda_d$ drives the
distillation loss down by orders of magnitude, often into the near-zero regime,
while reconstruction loss increases only modestly. This tradeoff motivates the
operating principle used throughout MEDAL: choose $\lambda_d$ large enough to
achieve faithful teacher recovery, then compare teacher settings using
reconstruction loss only among successfully distilled runs.}
    \label{fig:lambda_tradeoff}
\end{figure}

The distillation weight in \autoref{eq:scalarized} is the primary control
governing the near-zero-distillation regime. Empirically, increasing
$\lambda_d$ reduces the final distillation loss by orders of magnitude, while
reconstruction error increases comparatively modestly
(\autoref{fig:lambda_tradeoff}). Operationally, this makes the
choice straightforward: we prioritize faithful teacher matching and accept a
small reconstruction penalty in exchange.

\paragraph{Architecture capacity: depth and width.}
\label{app:depth}

In practice, MEDAL does not require extremely deep or extremely wide networks.
Across datasets with very different sample sizes and input dimensions, moderate
multilayer perceptron encoder--decoder architectures were sufficient to achieve
near-zero empirical distillation. To give practical guidance on architecture
selection, we study how depth and width affect both distillation and
reconstruction.

\begin{figure}[H]
    \centering
    \includegraphics[width=\linewidth]{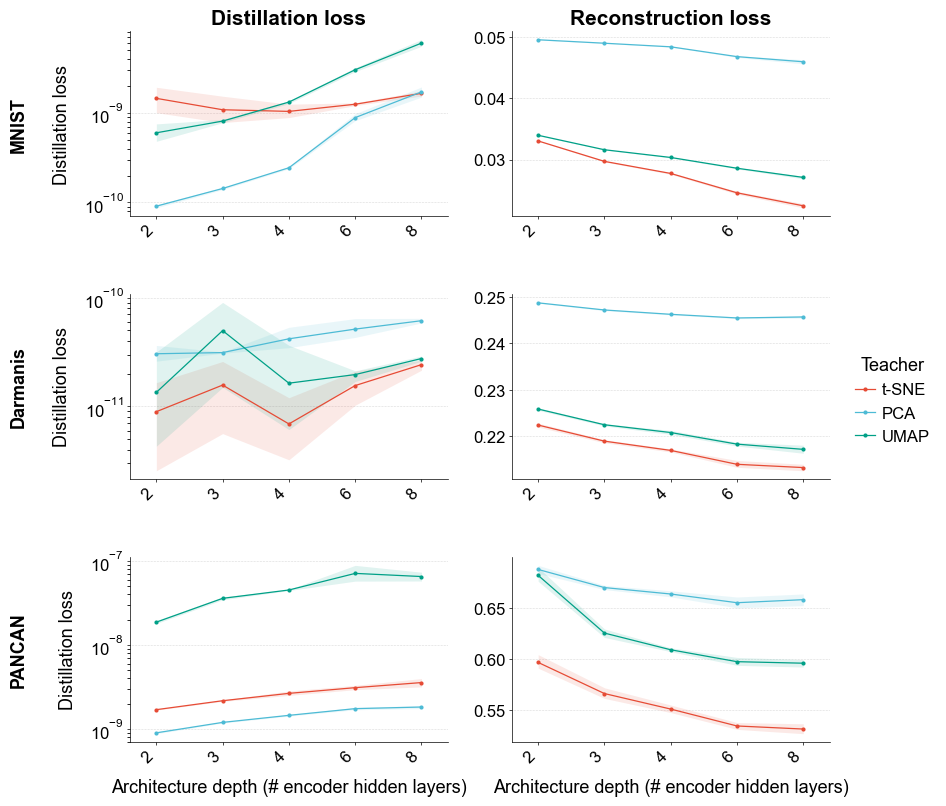}
    \caption{%
        \textbf{Effect of architecture depth on distillation and reconstruction
        loss under a fixed parameter budget.}
        Each line shows the mean $\pm$ 95\% CI across 20 random seeds.
        \emph{(Left)} Distillation loss on a log scale.
        \emph{(Right)} Reconstruction loss.
        Rows correspond to three datasets; colors denote teacher type.
        Width at each depth is adjusted to preserve a fixed total parameter
        count (see \autoref{tab:depth-width}).
    }
    \label{fig:depth}
\end{figure}

For a symmetric autoencoder with $L$ hidden layers per side, constant width $w$,
input dimension $D$, and latent dimension $z$, the total parameter count
(ignoring biases) is
\begin{equation}
    P(L,w) = 2(L-1)w^2 + 2(D+z)w.
    \label{eq:total-param}
\end{equation}
For the depth analysis, we fix a reference parameter budget
$P^{*}=P(L_{\rm ref},w_{\rm ref})$ with $L_{\rm ref}=3$ and
$w_{\rm ref}=256$, and solve for the width that preserves this budget at each
target depth:
\[
    w(L) = \frac{-(D+z)+\sqrt{(D+z)^2+2(L-1)P^{*}}}{2(L-1)}.
\]
All other training choices are held fixed across depths. The resulting
width--depth pairs are listed in \autoref{tab:depth-width}, and the empirical
results are shown in \autoref{fig:depth}.

\begin{table}[H]
\centering
\caption{Parameter-matched width--depth pairs. Reference architecture:
$L_{\mathrm{ref}}=3$, $w_{\mathrm{ref}}=256$, $z=2$.}
\label{tab:depth-width}
\begin{tabular}{lcccccc}
\toprule
Dataset & $D$ & $w(L{=}2)$ & $w(L{=}3)$ & $w(L{=}4)$ & $w(L{=}6)$ & $w(L{=}8)$ \\
\midrule
MNIST              &   784 & 305 & 256 & 227 & 191 & 169 \\
PANCAN             & 20531 & 260 & 256 & 253 & 247 & 242 \\
Darmanis           &   500 & 317 & 256 & 222 & 183 & 160 \\
\bottomrule
\end{tabular}
\end{table}

The central finding is that near-zero distillation does \emph{not} require deep
networks. Across datasets and teacher types, distillation loss is already at or
near machine precision with relatively shallow architectures, and increasing
depth beyond this point provides little systematic benefit. Under a fixed
parameter budget, deeper models also reduce per-layer width, which can make
optimization less efficient and increase training time without improving final
alignment. Reconstruction shows a somewhat more nuanced dependence on depth, but
the main practical conclusion is that moderate depth is sufficient for MEDAL's
distillation objective.

\paragraph{Per-layer width analysis.}
\label{app:size}

\begin{figure}[H]
    \centering
    \includegraphics[width=\linewidth]{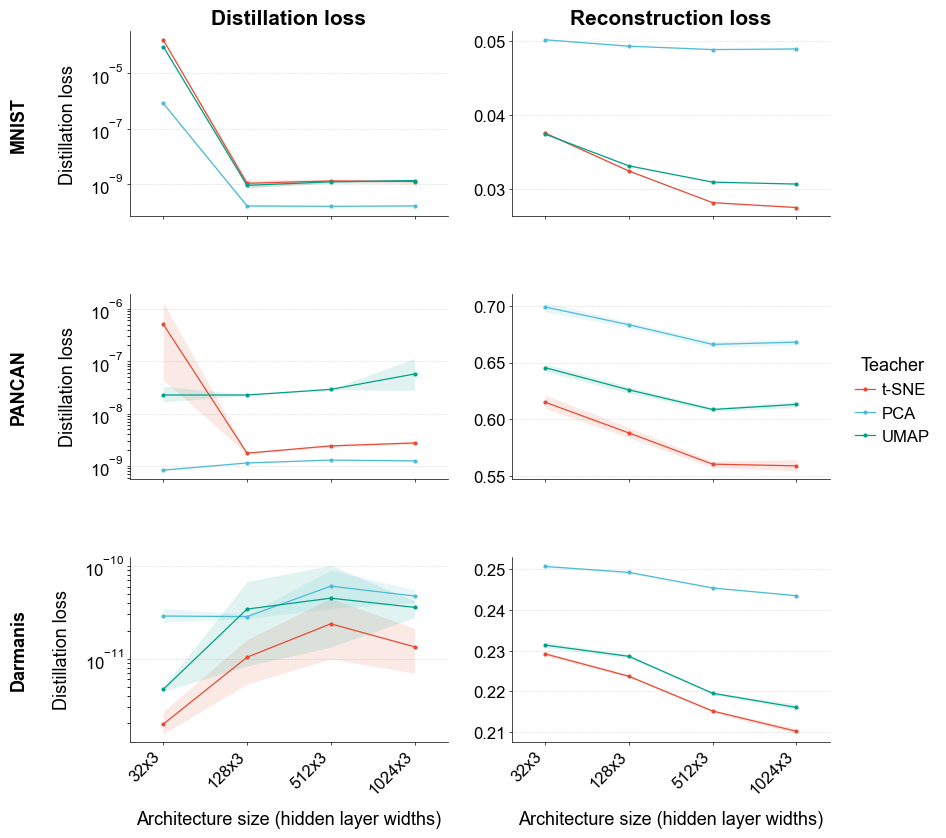}
    \caption{%
        \textbf{Effect of per-layer width on distillation and reconstruction
        loss under a fixed depth.}
        Each line shows the mean $\pm$ 95\% CI across 20 random seeds.
        \emph{(Left)} Distillation loss on a log scale.
        \emph{(Right)} Reconstruction loss.
        Rows correspond to three datasets; colors denote teacher type.
    }
    \label{fig:size}
\end{figure}

Complementing the depth study, we analyze the effect of model size by fixing the
number of hidden layers $L$ and varying the per-layer width $w$, shared across
layers. This directly changes the total parameter count in
\autoref{eq:total-param} while holding all other factors fixed: data splits and
preprocessing, teacher embedding, loss weights, initialization, optimizer,
batch size, and optimization budget. Results are shown in \autoref{fig:size}.

Once the student is sufficiently expressive, near-zero empirical distillation
becomes feasible. Increasing width improves approximation capacity and can
reduce distillation loss, but only up to a point; beyond that, gains saturate
while computational cost continues to increase. In practice, this suggests using
architectures that are moderately overparameterized rather than extremely wide.

\paragraph{Activation analysis.}
\label{app:activ-appendix}

\begin{figure}[H]
    \centering
    \includegraphics[width=\linewidth]{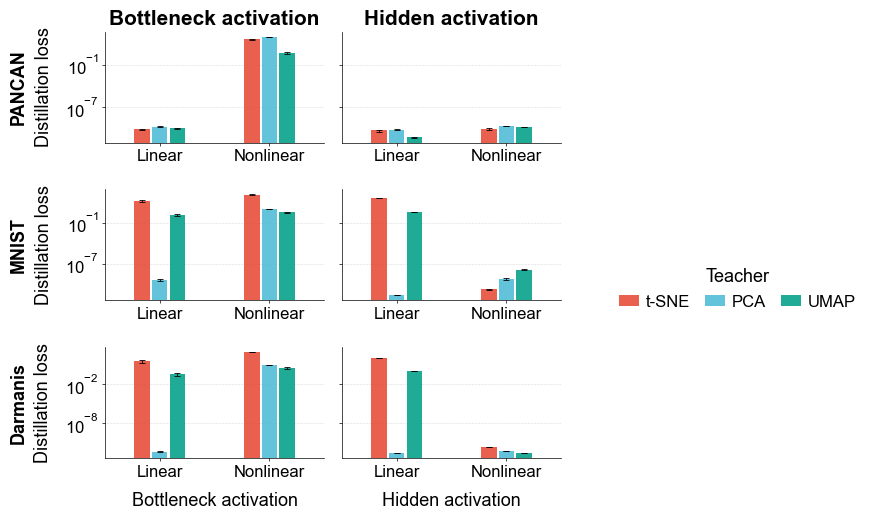}
    \caption{%
        \textbf{Effect of bottleneck and hidden-layer activation on distillation
        loss.}
        Each bar shows the median distillation loss across 20 random seeds.
        \emph{(Left)} Loss when the bottleneck activation is linear (identity)
        versus nonlinear (\textsc{ReLU} or \textsc{SELU}), marginalizing over
        all hidden-layer activations.
        \emph{(Right)} Loss when the hidden-layer activation is linear versus
        nonlinear, conditioned on a linear bottleneck.
        Results are shown for three datasets (PANCAN, MNIST, Darmanis) and four
        teachers (t-SNE, PCA, UMAP, Spectral).
    }
    \label{fig:activation-distill}
\end{figure}

This ablation examines how the encoder activation design affects distillation.
Write the encoder output as
\[
    z_\theta(x) = \phi_{\rm bn}\!\big(A\,h_\theta(x)+b\big),
\]
where $h_\theta$ is the penultimate representation, $A,b$ are the final affine
parameters, and $\phi_{\rm bn}$ is the bottleneck activation. We train
autoencoders over the grid
\[
    \phi_{\rm bn}\in\{\text{None},\text{ReLU},\text{SELU}\},
    \qquad
    \phi\in\{\text{None},\text{ReLU},\text{SELU}\},
\]
holding all other training choices fixed, and report the median distillation
loss over 20 random seeds. Results are shown in
\autoref{fig:activation-distill}, where the results for ReLU and SELU are
aggregated as ``Nonlinear.''

\textbf{A linear bottleneck is required for near-zero distillation.}
Exact matching requires $z_\theta(x_i)=z_i$ for all training points. When
$\phi_{\rm bn}$ is the identity, the final affine head is unconstrained and can
in principle reproduce any target coordinates. When $\phi_{\rm bn}$ is
nonlinear, the output is restricted to the image of that function:
\textsc{ReLU} constrains each coordinate to $[0,\infty)$, and \textsc{SELU} to
$(-\alpha\lambda,\infty)$ \cite{klambauer2017selfnormalizingneuralnetworks}.
Teacher embeddings can contain values outside these ranges, producing an
irreducible error floor regardless of how expressive $h_\theta$ is. The left
panel of \autoref{fig:activation-distill} confirms this: a nonlinear bottleneck
raises distillation loss by several orders of magnitude across datasets and
teachers.

\textbf{Hidden-layer nonlinearity is needed for nonlinear teachers.}
Conditioned on a linear bottleneck, a fully linear encoder can only represent
affine functions of the input and therefore cannot faithfully distill nonlinear
teachers such as t-SNE or UMAP. The right panel of
\autoref{fig:activation-distill} shows that switching the hidden layers from
linear to nonlinear sharply reduces distillation loss for nonlinear teachers,
whereas PCA, being a linear teacher, is already well matched by a linear
encoder.

Together, these results give a simple design rule: near-zero distillation
requires \textbf{(i)} a linear bottleneck activation and \textbf{(ii)} at least
one hidden-layer nonlinearity when the teacher itself is nonlinear.

\paragraph{Adaptive optimization, stopping criteria, and teacher normalization.}
\label{app:early-stopping}

Different teacher embeddings exhibit markedly different optimization dynamics:
some can be distilled to near-zero alignment rapidly, whereas others require
substantially longer training. To handle this heterogeneity without
teacher-specific tuning, we train all models with a common initial learning
rate, a sufficiently large maximum budget, and PyTorch's
\texttt{ReduceLROnPlateau} scheduler. The scheduler monitors distillation loss
and reduces the learning rate when progress plateaus, allowing easier
distillation tasks to converge quickly while harder ones continue training with
progressively smaller step sizes.

To monitor progress toward successful distillation, we define a
target band corresponding to our near-zero distillation regime for the distillation loss and track when training enters this band. In practice, we
treat runs with $\mathcal{L}_d \le 9\times 10^{-6}$ as successfully distilled.
Empirically, this threshold provides a good balance across datasets, teachers,
and teacher configurations: it yields embeddings that are visually
indistinguishable from their teacher references while avoiding unnecessary
compute once reconstruction has already stabilized.

Early stopping is based on both successful distillation and loss stabilization. Once the distillation loss enters the predefined near-zero regime and sufficient training history
has accumulated, we estimate the local slope of both losses over a fixed
stability window:
\[
\frac{\big|\mathcal{L}_d^{(t)}-\mathcal{L}_d^{(t-w)}\big|}{w}
< \epsilon_{\rm distill},
\qquad
\frac{\big|\mathcal{L}_r^{(t)}-\mathcal{L}_r^{(t-w)}\big|}{w}
< \epsilon_{\rm recon},
\]
where $w$ is the stability window. If both conditions hold, the run is counted
as stable for that checkpoint; otherwise the stability counter is reset. Training
is terminated only after stability has been maintained for a predefined number
of consecutive checkpoints (the patience parameter), and only when the run lies
in the final target band. In our implementation, we typically use
$\epsilon_{\rm distill}=10^{-7}$, $\epsilon_{\rm recon}=10^{-3}$, and patience $=50$. If these criteria are not met,
training continues until the maximum budget is reached. This adaptive schedule
provides a single protocol that remains comparable across teachers while
avoiding both over-training of easy cases and premature truncation of harder
ones.

A further issue is that teacher reference embeddings can differ substantially in
scale: the coordinate ranges of their $x$- and $y$-axes may vary across methods,
hyperparameters, and datasets. Without adjustment, this would place the
distillation objective and the induced reconstruction error on different
numerical scales across teacher instances, making comparisons less fair. To
address this, MEDAL normalizes each teacher embedding before distillation by
first centering it at its empirical mean and then scaling by its root-mean-square
radius,
\[
\tilde Z = \frac{Z-\bar Z}{s},
\qquad
s = \sqrt{\frac{1}{n}\sum_{i=1}^n \|z_i-\bar Z\|^2},
\]
where $\bar Z = n^{-1}\sum_{i=1}^n z_i$. This normalization puts teacher
embeddings on a common scale, so that both distillation difficulty and
reconstruction error are comparable across datasets, methods, and hyperparameter
settings.

\subsubsection*{Recommendations for architecture and optimization setup}

The preceding ablations suggest a practical training protocol rather than a
single universally optimal architecture. In our experiments, effective MEDAL
training is achieved by using enough capacity to reach the near-zero
distillation regime, while avoiding architectures that are unnecessarily large
relative to the dataset. We therefore recommend the following default setup.

\begin{itemize}
    \item \textbf{Use a symmetric multilayer perceptron autoencoder with a
    linear bottleneck.}
    Hidden layers may use standard nonlinear activations such as \textsc{ReLU}
    or \textsc{SELU}, but the encoder bottleneck should be left linear
    because the teacher coordinates are real-valued and should not be restricted
    by the range of a nonlinear activation. We also use a linear output layer
    for reconstruction.

    \item \textbf{Start with a moderate architecture and increase capacity only
    if distillation fails.}
    For many datasets, two to four hidden layers per side with widths in the
    range of 256--512 provide a useful starting point. For example, smaller or
    lower-dimensional datasets may be well served by architectures such as
    $[256,256]$, whereas higher-dimensional or more heterogeneous datasets may
    require $[512,512]$ or $[512,512,512,512]$. The goal is not to maximize
    depth or width, but to use sufficient capacity so that the final
    distillation loss reaches the predefined near-zero regime.

    \item \textbf{Tune $\lambda_d$ on a logarithmic scale and prioritize
    successful teacher matching.}
    In practice, we sweep $\lambda_d$ over a small logarithmic grid, such as
    $\{10,10^2,10^3,10^4\}$, or sample from the same range during random search.
    Values that are too small may give favorable reconstruction loss but fail
    to recover the teacher geometry, values that are excessively large for a given dataset can lead to inefficient optimization and, in finite training, potentially higher reconstruction error than necessary. We therefore
    choose $\lambda_d$ large enough to reliably achieve near-zero distillation
    and then compare reconstruction only among successfully distilled runs.

    \item \textbf{Use adaptive optimization rather than a fixed training
    horizon.}
    We typically sweep the initial learning rate over
    $\{10^{-3},10^{-4},10^{-5}\}$, with $10^{-4}$ or $10^{-5}$ often providing
    stable behavior in larger experiments. A learning-rate scheduler such as
    \texttt{ReduceLROnPlateau} is useful because different teachers can have
    different optimization difficulty. Training is stopped only after the model
    enters the near-zero distillation band and both distillation and
    reconstruction losses have stabilized.

    \item \textbf{Keep the selected protocol fixed across teacher settings.}
    Once an architecture and training protocol are chosen for a dataset, the
    same setup should be reused across teacher hyperparameters and teacher
    methods. This prevents the held-out reconstruction comparison from being
    confounded by teacher-specific architecture tuning.
\end{itemize}

Overall, the recommended practice is to treat architecture and optimization as
a feasibility step for achieving faithful teacher distillation. After this
criterion is met, the primary comparison of interest is the reconstruction
behavior induced by different teacher embeddings, rather than small differences
in training configuration.
\subsection{Why near-zero distillation is feasible in principle}
\label{sec:supp_theory}

The practical results above show that near-zero empirical distillation is
achievable with moderate architectures and careful optimization. This is also
consistent with standard expressivity results for neural networks.

\paragraph{Classical universal approximation.}
Standard universal approximation theorems establish that multilayer perceptrons
with non-polynomial activation functions are dense in spaces of continuous
functions on compact domains
\cite{HORNIK1989359, Cybenko1989ApproximationBS, HORNIK1991251, LESHNO1993861}.
Later refinements show universality under bounded-width constructions
\cite{lu2017expressivepowerneuralnetworks, hanin2018approximatingcontinuousfunctionsrelu}
and characterize approximation rates for deep ReLU networks
\cite{YAROTSKY2017103, guhring2020expressivitydeepneuralnetworks, Barron1993UniversalAB}.
These results justify the view that sufficiently expressive encoder networks can
approximate a broad class of teacher maps.

\paragraph{Finite-sample interpolation.}
MEDAL, however, does not require approximation of a population-level map; it
requires fitting teacher coordinates on a fixed finite training set. A
complementary line of work shows that ReLU networks can interpolate arbitrary
labels on a finite sample. In particular, \citet{zhang2017understandingdeeplearningrequires}
showed that, for any set of $n$ distinct inputs and scalar labels, a two-layer
ReLU network can fit the labels exactly. For vector-valued targets,
\citet{yun2019smallrelunetworkspowerful} give an explicit construction showing
that a three-layer ReLU network with hidden widths $d_1$ and $d_2$ can memorize
arbitrary $r$-dimensional outputs whenever $d_1d_2 \ge 4nr$. Since the teacher
coordinates in MEDAL are precisely vector-valued labels attached to the
observed inputs, these results imply that exact fitting on the training set is
feasible in principle.

\begin{corollary}
Let $S=\{(x_i,z_i)\}_{i=1}^n$ with $x_i\in\mathbb{R}^k$ and
$z_i\in\mathbb{R}^r$. Under the conditions of Theorem 3.1 of
\citet{yun2019smallrelunetworkspowerful}, there exists a three-layer ReLU
network $g_\theta:\mathbb{R}^k\to\mathbb{R}^r$ with hidden widths
$d_1,d_2$ satisfying $d_1d_2\ge 4nr$ such that
\[
g_\theta(x_i)=z_i, \qquad i=1,\ldots,n.
\]
\end{corollary}

\paragraph{Interpretation for MEDAL.}
These results are existence statements: they certify that weights achieving
exact interpolation exist, but they do not guarantee that gradient-based
training with reconstruction pressure will always find them at modest widths.
For that reason, our main claim is empirical rather than purely theoretical. The
theory explains why near-zero alignment should be possible in principle, while
the ablations above show that it is achievable in practice with moderate
architectures and reasonable computation.

\subsection{Reconstruction remains meaningful under distillation}
\label{sec:recon-valid}
A central question for MEDAL is whether reconstruction error remains meaningful
once the bottleneck is constrained to match a prescribed teacher embedding. If
the distillation constraint were too restrictive, reconstruction error might
simply reflect optimization failure rather than information loss induced by the
teacher manifold. To study this, we examine how reconstruction performance
changes with bottleneck rank when the teacher is PCA. For each rank, we compare
four models: an unconstrained vanilla autoencoder, PCA, a linear MEDAL model
distilled from PCA, and a nonlinear MEDAL model distilled from the same PCA
teacher. Reconstruction error is evaluated on both training and held-out test
data. This experiment is designed to answer a basic question: if we require the
bottleneck to match a chosen manifold, how much reconstruction do we lose?

PCA provides a useful sanity check for this comparison. Classical results show
that the global minima of the linear autoencoder objective recover the principal
$r$-dimensional subspace of the data, up to a rotation within that subspace
\cite{BourlardKamp1988,BaldiHornik1989,plaut2018principalsubspacesprincipalcomponents}.
Accordingly, when the teacher is PCA and both encoder and decoder are linear, a
distilled MEDAL model should reproduce the classical PCA reconstruction profile.
At the same time, the vanilla autoencoder provides an important baseline:
because it optimizes reconstruction alone, without any geometric constraint, it
has an inherent advantage when reconstruction is the sole objective. Any
competitive MEDAL curve should therefore be interpreted as evidence that
reconstruction error remains informative even under a distillation constraint.
All architectural choices, schedulers, early stopping criteria, and training
protocols are kept identical across methods.

\begin{figure}
    \centering
    \includegraphics[width=0.9\linewidth]{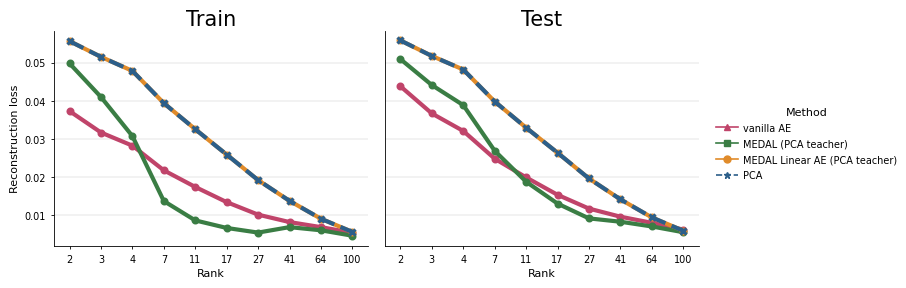}
    \caption{\textbf{Reconstruction error versus bottleneck rank on MNIST.}
    Train (left) and test (right) reconstruction loss are shown as a function of
    bottleneck rank for Vanilla AE, MEDAL + PCA, MEDAL Linear AE + PCA, and PCA.
    Linear MEDAL + PCA overlaps exactly with PCA across ranks, providing a sanity
    check in the linear setting. The nonlinear MEDAL + PCA model remains highly
    competitive with the unconstrained Vanilla AE and can outperform it at
    moderate and large ranks, showing that enforcing a geometrically meaningful
    bottleneck need not come at a substantial reconstruction cost.}
    \label{fig:mnist-rank-sweep}
\end{figure}

\autoref{fig:mnist-rank-sweep} shows the resulting reconstruction curves on
MNIST. The exact overlap between linear MEDAL + PCA and PCA across all ranks
provides a clean sanity check: when both the teacher and the student are
linear, MEDAL reproduces the classical PCA reconstruction behavior and does not
introduce artificial distortion. This establishes that the distillation
framework itself is compatible with a setting where the correct reconstruction
profile is already known.

The nonlinear MEDAL + PCA model yields a second and more practically important
observation. At very small bottleneck ranks, the vanilla autoencoder performs
better, which is expected: when the latent dimension is extremely limited, an
unconstrained model can devote the entire bottleneck to reconstruction. As the
rank increases, however, nonlinear MEDAL + PCA quickly becomes competitive with
the vanilla autoencoder and at moderate to large ranks can even outperform it on
both training and test loss. One plausible interpretation is that reconstruction-only
training leaves greater freedom in how the bottleneck is organized, whereas
distillation to a strong teacher such as PCA anchors the latent representation
to a meaningful low-dimensional geometry that can stabilize optimization and
improve decoder use of those coordinates.

Taken together, these results support the validity of reconstruction error as a
criterion after distillation. Distillation does not simply throw away
reconstruction in exchange for geometry. Rather, it imposes structure on the
bottleneck while retaining---and in this setting sometimes improving---
reconstruction quality. Once teacher matching is essentially exact, the
remaining reconstruction error can therefore be interpreted as a meaningful
measure of how much input-space information is lost by compression through the
teacher manifold. This is precisely the quantity later used in MEDAL for
held-out model selection, method comparison, and pointwise distortion analysis.

\subsection{Held-out evaluation protocol}
\label{sec:supp_validation}

Once a suitable student architecture is identified for a given dataset, we reuse
that architecture across the full grid of teacher hyperparameters to ensure fair
comparison across teacher settings. For each setting, we repeat student training
multiple times to account for optimization stochasticity and summarize
performance across runs.

MEDAL enables a shared held-out evaluation protocol for nonlinear embeddings. For
each teacher hyperparameter setting, we first fit the teacher embedding on a
training split and then distill it into a student autoencoder. Because held-out
evaluation is only meaningful when the student has faithfully recovered the
teacher geometry, we first filter to runs that achieve our predefined near-zero
distillation criterion. The learned encoder and decoder from these successfully
distilled runs are then applied to held-out validation observations: the encoder
maps each point into the fixed teacher manifold learned from the training split,
and the decoder maps that point back to the original feature space.
Reconstruction loss on the held-out data, reported as mean $\pm$ one standard
error over the successfully distilled reruns, is then used as the evaluation
criterion.

This protocol differs from conventional DR practice, where hyperparameters are
often chosen visually from the training embedding alone. Because MEDAL equips
the teacher with an explicit out-of-sample map and approximate inverse, teacher
settings can instead be compared by how well they preserve information for
unseen data. The same protocol also allows different DR methods to be evaluated
on a more common basis, since each teacher is distilled into the same student
class and assessed by the same held-out reconstruction criterion.

\paragraph{Compute and parallelization.}
To keep wall-clock time reasonable while sweeping teacher hyperparameters and
seeds, we launch trials in parallel with \emph{Ray Tune}. Each trial runs the
same training loop and terminates as soon as it (i) enters the near-zero
distillation band and (ii) satisfies the stabilization check for both distill
and reconstruction losses, so compute is not spent past the point of diminishing
returns (see \autoref{app:early-stopping}). Trials are independent
(teacher $\times$ hyperparameter $\times$ seed) and scheduled concurrently with
per-trial resource caps; exact allocations are adjusted per dataset and affect
only wall-clock time, not the reported metrics. Ray Tune handles logging,
checkpointing, and automatic retries, allowing many candidates to be swept
quickly while keeping the student architecture and training budget fixed.

\section{Additional Case Study Results}
Throughout the experiments, we tune the standard neighborhood-size parameters of each teacher method. For UMAP, \texttt{n\_neighbors} controls the size of the local neighborhood used to construct the manifold graph: smaller values emphasize finer local structure, whereas larger values place more weight on global organization. For t-SNE, perplexity plays an analogous role as an effective neighborhood size in the high-dimensional affinity construction, with smaller perplexities favoring more local separation and larger perplexities producing smoother, more global embeddings. For PHATE and spectral embedding, we similarly tune the neighborhood parameter used to construct the underlying graph. Unless otherwise stated, all hyperparameter comparisons are made after distilling each fitted teacher embedding into the same MEDAL student architecture within a dataset.

\subsection{MNIST}
\begin{figure}[H]
    \centering
    \includegraphics[width=\linewidth]{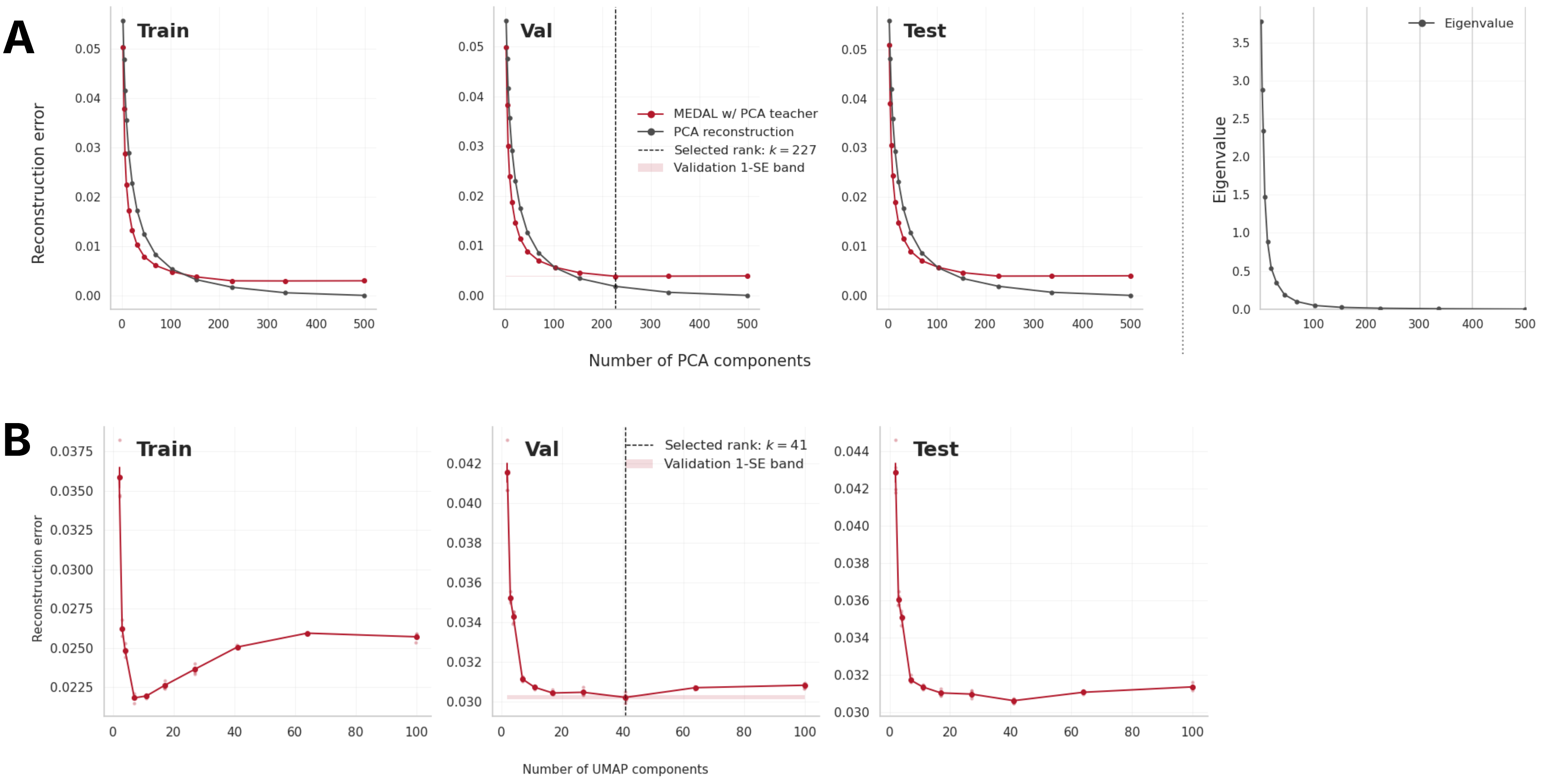}
    \caption{\textbf{Tuning the retained dimension of PCA and UMAP teachers.}
    \textbf{(A)} PCA provides a proof-of-concept setting in which the exact
    rank-$k$ reconstruction error is available. Although MEDAL is not needed to
    tune the rank of a PCA teacher, the MEDAL reconstruction score closely tracks
    the exact PCA reconstruction error across train, validation, and test splits.
    The eigenvalue scree plot is shown for reference. In contrast to the classical
    scree-plot elbow criterion, validation reconstruction can favor a larger
    retained rank because it directly optimizes held-out input-space recovery:
    many low-variance components contribute little individually but collectively
    reduce reconstruction error.
    \textbf{(B)} For nonlinear teachers such as UMAP, there is no analogous
    eigenvalue scree plot. MEDAL extends the same validation-reconstruction
    principle to select the embedding dimension of a fitted nonlinear DR teacher.
    This provides a quantitative criterion for choosing the dimension of a UMAP
    representation before downstream analyses such as clustering, cell-state
    annotation, or other embedding-based tasks \cite{becht2019dimensionality,
    hasanaj2022interactive}. Vertical dashed lines mark the selected dimension;
    shaded bands indicate the validation one-standard-error band.}
    \label{fig:mnist-tune-rank}
\end{figure}

The same validation principle can also be used to choose the embedding dimension with MEDAL. For PCA, the retained rank is commonly selected from a scree plot and the elbow heuristic \cite{cattell1966scree}. For nonlinear teachers such as UMAP, analogous dimension-selection criteria are ill-defined; the embedding dimension is instead a user-specified hyperparameter, despite such embedding often used as an intermediate representation for downstream clustering or scientific inference \cite{dorrity2020umap,rosito2023application}. In \suppfigref{fig:mnist-tune-rank}, we show that MEDAL provides a shared criterion for selecting the student bottleneck dimension across PCA and UMAP teachers by minimizing held-out reconstruction error on MNIST.

\begin{figure}[H]
    \centering
    \includegraphics[width=\linewidth,height=0.9\textheight, keepaspectratio]{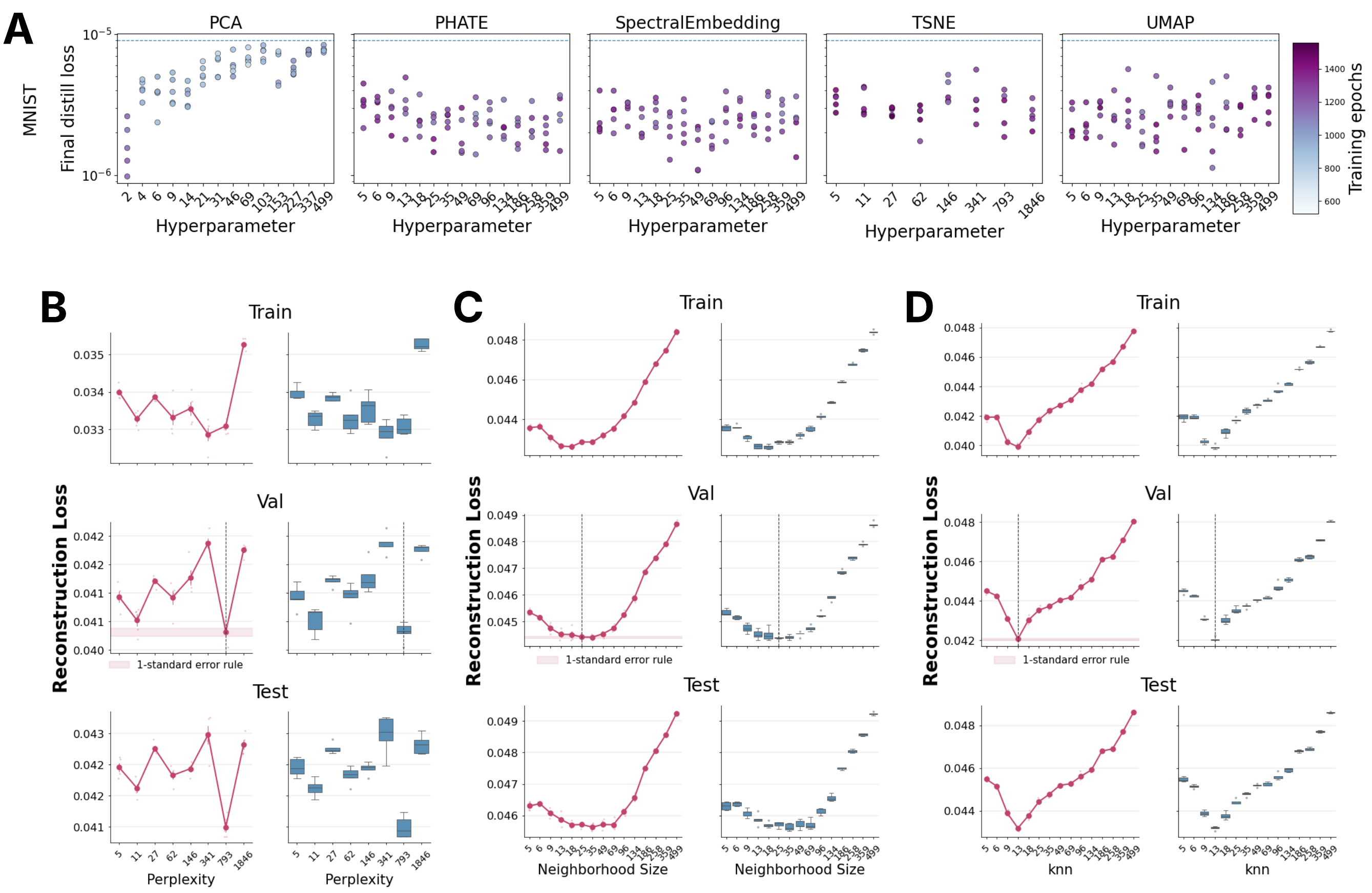}
    \caption{\textbf{MNIST Hyperparameter Tuning}. \textbf{A}, Distillation results for all MNIST runs. \textbf{B}, t-SNE teacher, tuning \texttt{perplexity}. \textbf{C}, SpectralEmbedding teacher, tuning \texttt{n\_neighbors}. \textbf{D}, PHATE teacher, tuning \texttt{knn}.}
    \label{fig:mnist-tuning}
\end{figure}

We extended the MNIST hyperparameter-tuning analysis to additional teacher methods, including t-SNE, PHATE and spectral embedding (Supplementary \autoref{fig:mnist-tuning}). For each teacher, we distilled embeddings across a method-specific hyperparameter grid using the same student architecture and training protocol, and summarized reconstruction performance across five runs using both the mean and the median. 
We set up an encoder network of 4 512-neuron layers, successfully distilling teacher embeddings across all hyperparameters.

\subsection{Hydra}

\begin{figure}[!p]
    \centering
    \includegraphics[width=\linewidth,height=0.9\textheight, keepaspectratio]{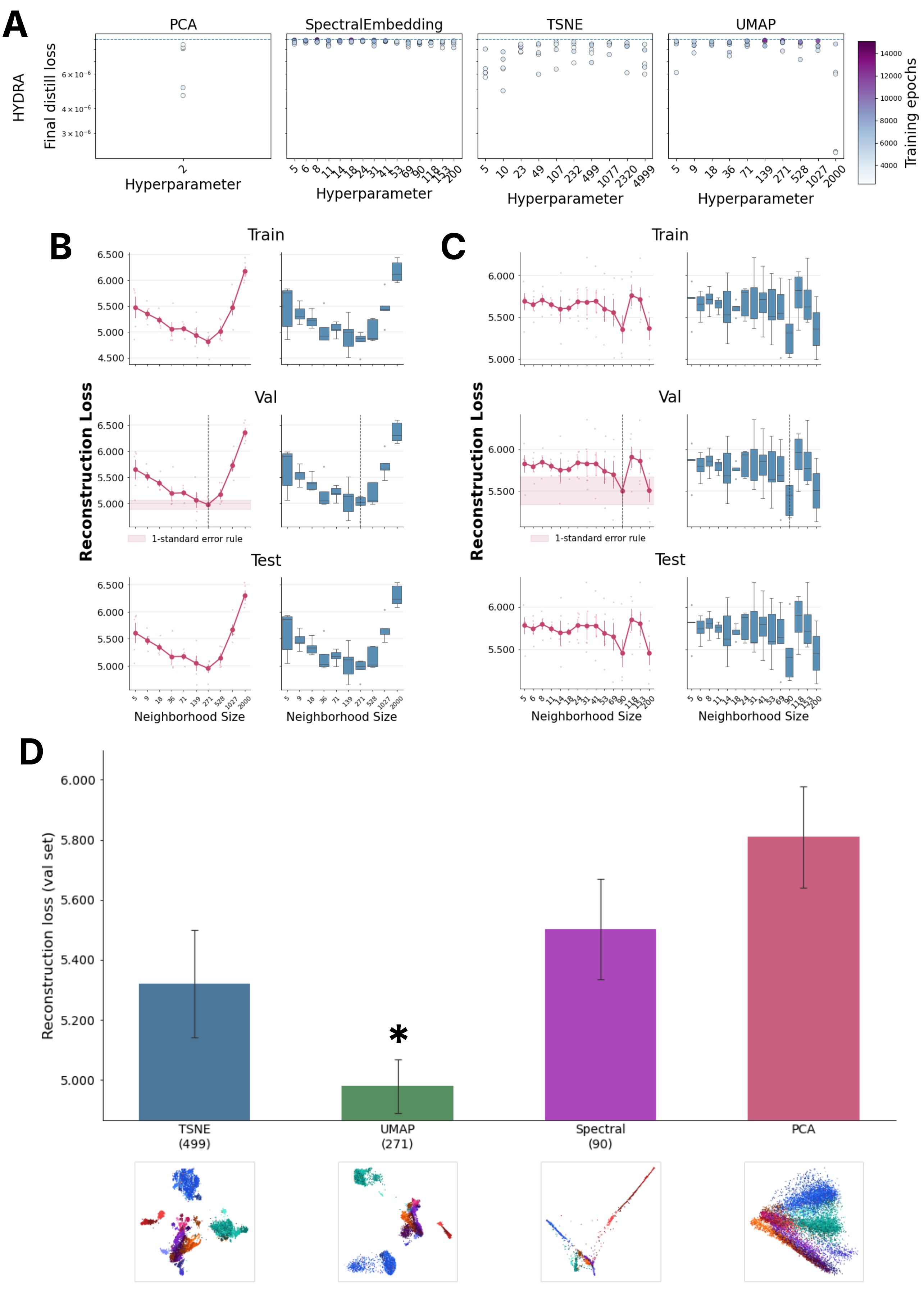}
    \caption{\textbf{Hydra Hyperparameter Tuning}. \textbf{A}, Distillation results for all Hydra runs. \textbf{B}, UMAP teacher, tuning \texttt{n\_neighbors}. \textbf{C}, SpectralEmbedding teacher, tuning \texttt{n\_neighbors}. \textbf{D}, Comparing different DR methods on Hydra using heldout reconstruction error.}
    \label{fig:hydra-tuning}
\end{figure}

\begin{figure}[!p]
    \centering
    \includegraphics[width=\linewidth]{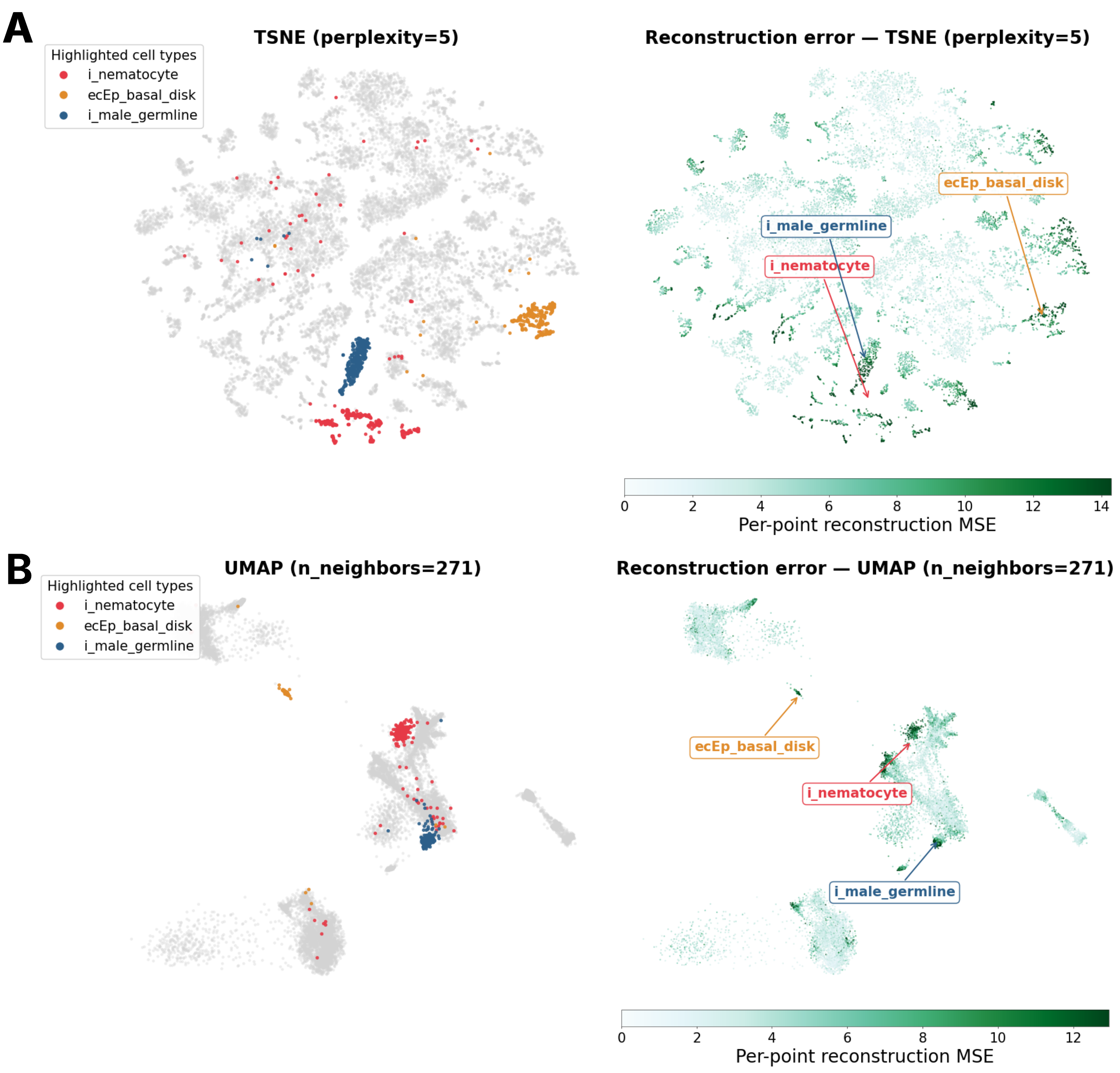}
    \caption{\textbf{MEDAL identifies consistent biologically structured distortion across Hydra teacher embeddings.}
    \textbf{A,} MEDAL distortion analysis for a t-SNE teacher with small perplexity
    (\(\text{perplexity}=5\)). Left: Hydra embedding with three cell populations
    highlighted: interstitial nematocytes (\texttt{i\_nematocyte}), ectodermal basal
    disk epithelial cells (\texttt{ecEp\_basal\_disk}), and male germline cells
    (\texttt{i\_male\_germline}). Right: the same embedding colored by per-point
    reconstruction mean squared error. 
    \textbf{B,} MEDAL distortion analysis for a UMAP teacher
    (\(\texttt{n\_neighbors}=271\)), shown with the same highlighted cell
    populations and corresponding reconstruction-error map. Across both teacher
    settings, high reconstruction error repeatedly localizes near the same
    biologically distinct cell populations highlighted in the main Hydra analysis.
    This consistency suggests that the distortion signal is not an artifact of a
    single t-SNE hyperparameter choice, but reflects cell populations whose
    high-dimensional transcriptomic structure is difficult to preserve in a
    two-dimensional embedding. Color scales are shown separately for each teacher
    embedding.}
    \label{fig:hydra-biocoherence}
\end{figure}

\subsection{Neocortex}
\begin{figure}[!p]
    \centering
    \includegraphics[width=\linewidth,height=0.9\textheight, keepaspectratio]{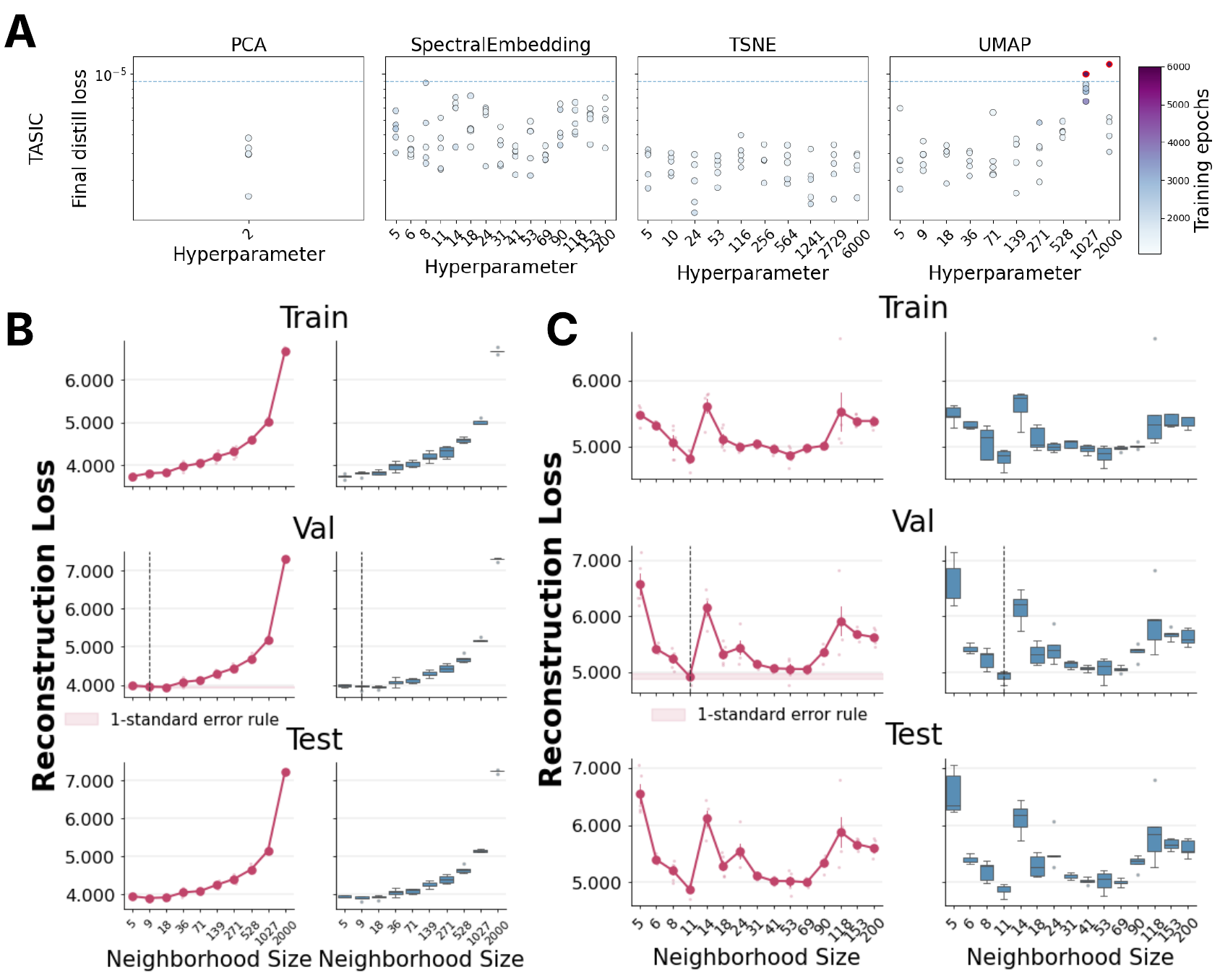}
    \caption{\textbf{Neocortex Hyperparameter Tuning}. \textbf{A}, Distillation results for all Neocortex runs. \textbf{B}, UMAP teacher, tuning \texttt{n\_neighbors}. \textbf{C}, SpectralEmbedding teacher, tuning \texttt{n\_neighbors}.}
    \label{fig:tasic-tuning}
\end{figure}

\begin{figure}[!p]
    \centering
    \includegraphics[width=\linewidth]{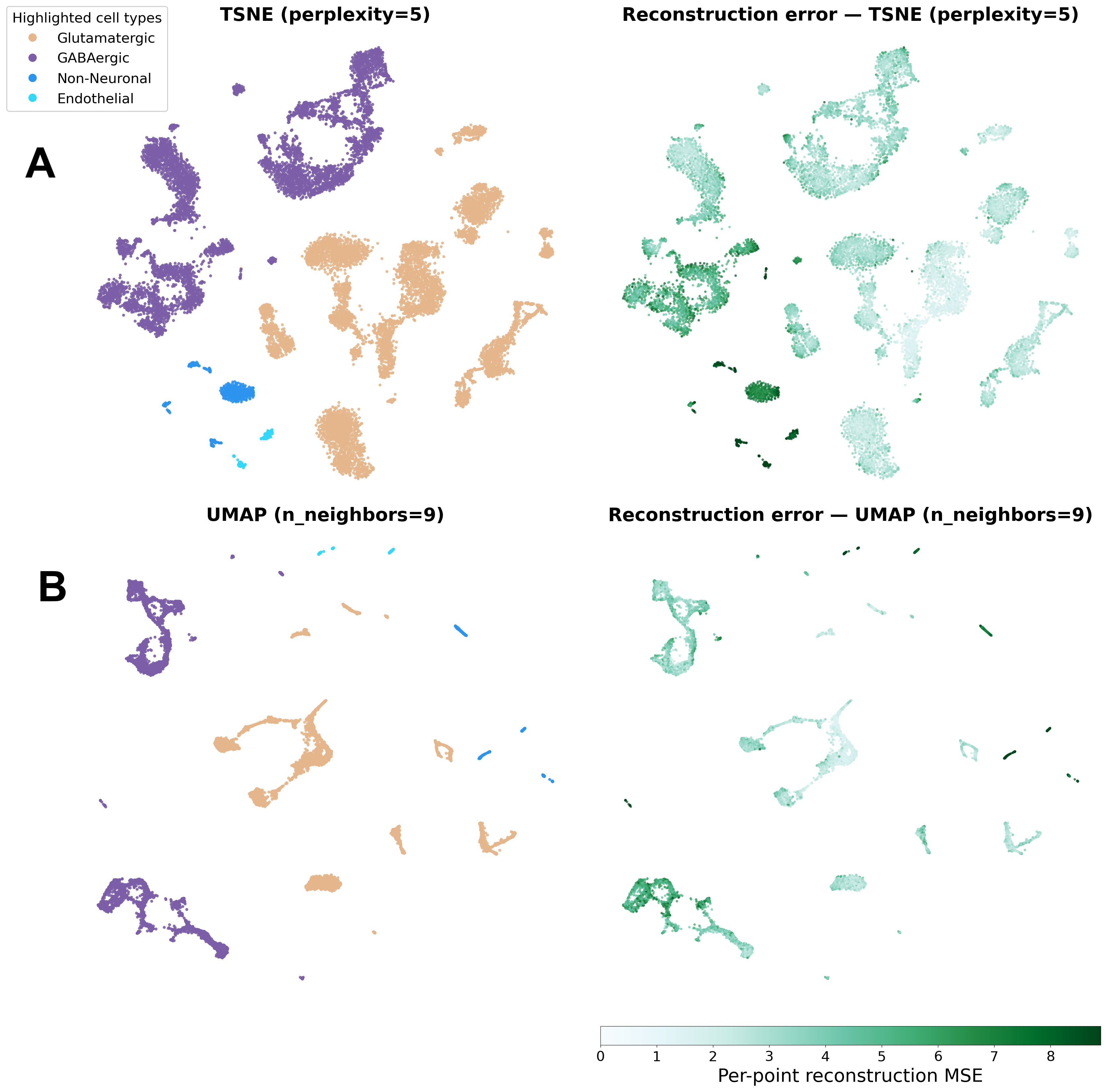}
    \caption{\textbf{MEDAL identifies consistent biologically structured distortion across Neocortex teacher embeddings.}
    \textbf{A,} Distortion analysis for a t-SNE teacher with small perplexity
(\(\text{perplexity}=5\)). Left: the Neocortex embedding with broad annotated cell
classes highlighted, including glutamatergic neurons, GABAergic neurons,
non-neuronal cells, and endothelial cells. Right: the same embedding colored by
MEDAL per-point reconstruction mean squared error.
\textbf{B,} Distortion analysis for a UMAP teacher
(\(\texttt{n\_neighbors}=9\)), shown with the same highlighted cell classes and
corresponding reconstruction-error map. Across both teacher settings, higher
reconstruction error remains concentrated disproportionately in non-neuronal and
endothelial regions relative to the major neuronal populations. This consistency
supports the main-text observation that MEDAL identifies a stable, biologically
coherent distortion signal in the Neocortex dataset, rather than an artifact of a single teacher
or hyperparameter choice.}
\label{fig:tasic-biocoherence}
\end{figure}

\FloatBarrier

\subsection{APOGEE stellar spectra: additional physical-sciences benchmark}
\label{sec:supp-apogee}

To assess whether MEDAL's validation workflow extends beyond image and
single-cell genomics datasets, we also applied MEDAL to APOGEE stellar spectra.
This dataset consists of approximately 8,000 stellar spectra, each represented
by flux measurements across 19 wavelength bands. The data and preprocessing
pipeline follow \citet{chang2025unsupervisedmachinelearningscientific}. Unlike
MNIST or the single-cell atlases, APOGEE does not have externally validated
class labels for the present analysis; cluster assignments from the prior
unsupervised analysis are therefore used only for visualization, not as
ground-truth labels.

Despite this limitation, APOGEE provides a useful domain-diversity check for
MEDAL. We repeated the same validation workflow used in the main case studies:
distilling t-SNE and UMAP teachers across hyperparameter grids, selecting
hyperparameters by held-out reconstruction loss, comparing selected teacher
methods under a shared reconstruction criterion, and visualizing pointwise
reconstruction error. MEDAL selected a t-SNE teacher with perplexity \(6\) and a
UMAP teacher with \(\texttt{n\_neighbors}=359\). Under the shared validation
criterion, t-SNE achieved the lowest held-out reconstruction loss among the
teachers considered, followed by UMAP and PCA
(\suppfigref{fig:apogee}). These results show that MEDAL can be applied as a
general validation layer in a physical-sciences dataset, although the absence of
ground-truth labels limits the strength of biological or physical
interpretation relative to the main case studies.

\begin{figure*}[!p]
    \centering
    \includegraphics[width=\linewidth]{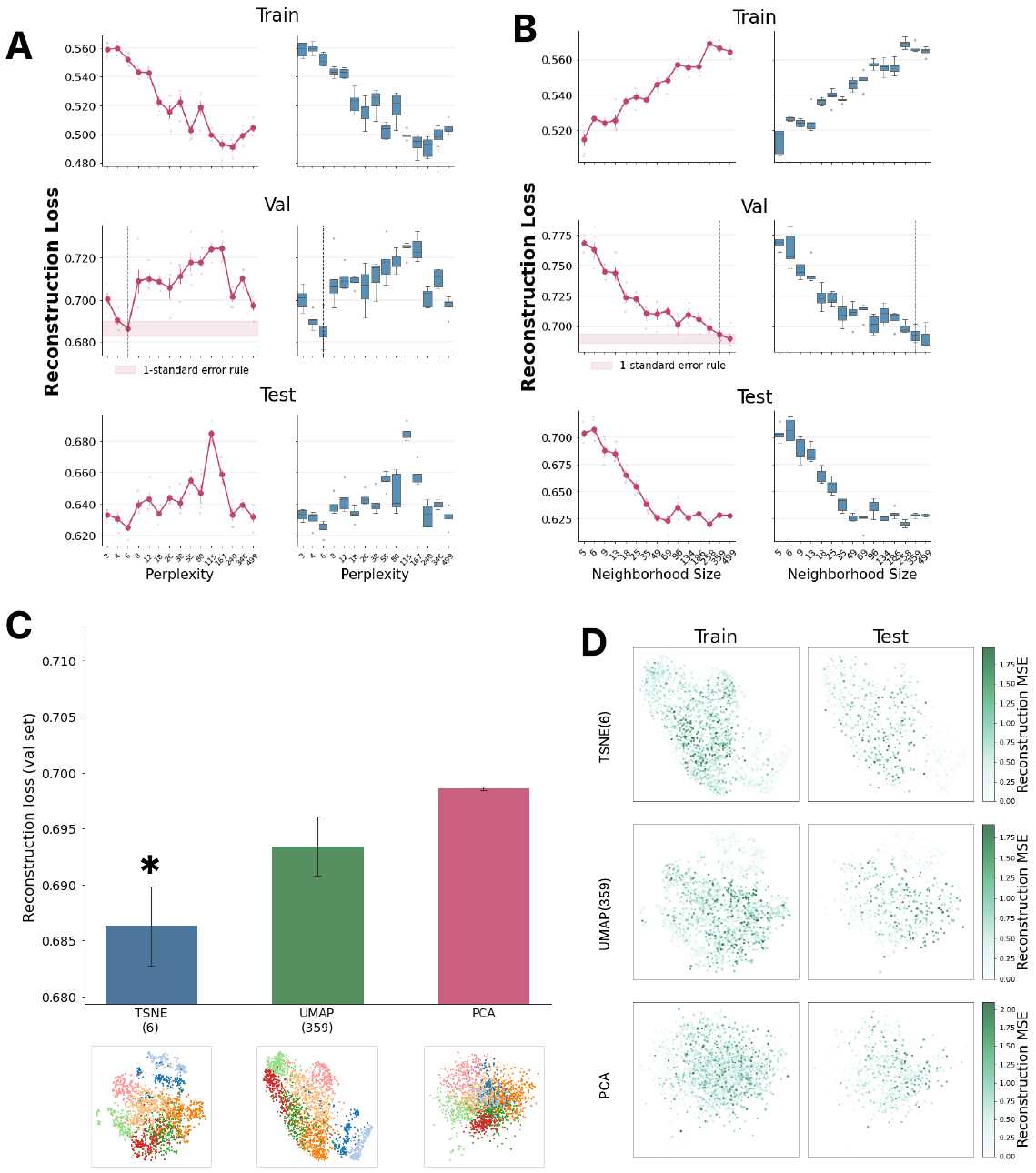}
    \caption{}
    \label{fig:apogee}
\end{figure*}

\begin{figure*}
    \ContinuedFloat
    \caption{\textbf{MEDAL validation on APOGEE stellar spectra.}
    \textbf{A,} MEDAL hyperparameter tuning for t-SNE teachers across perplexity
    values. Reconstruction loss is shown on train, validation, and test splits, with
    the selected perplexity indicated by the dashed vertical line.
    \textbf{B,} MEDAL hyperparameter tuning for UMAP teachers across neighborhood
    sizes. Reconstruction loss is again evaluated on train, validation, and test
    splits, with the selected \(\texttt{n\_neighbors}\) indicated by the dashed
    vertical line.
    \textbf{C,} Comparison across dimension-reduction teachers under the MEDAL
    validation protocol. t-SNE at perplexity \(6\) achieved the lowest validation
    reconstruction loss among the methods considered, followed by UMAP at
    \(\texttt{n\_neighbors}=359\) and PCA.
    \textbf{D,} Pointwise reconstruction-error maps for the selected t-SNE, UMAP,
    and PCA embeddings on train and test splits. Cluster colors are used only for
    visualization and are taken from the prior unsupervised analysis of this dataset,
    rather than from externally validated ground-truth labels.}
\end{figure*}

\newpage
\subsection{Macaque retina: additional mixed-subject batches and comparison with \texttt{umap.transform()}}
\label{sec:supp-macaque}

We performed additional macaque retina analyses to test whether the
out-of-distribution signal reported in the main text was specific to a single
subject comparison or to a single choice of reference manifold. We considered
three extensions. First, in addition to the M1-versus-M2 comparison shown in
\autoref{fig:macaque}, we projected M3 cells onto the M1 reference manifold.
Second, we reversed the reference subject by training MEDAL on M2 cells and
projecting M1 cells onto the M2 reference manifold. Third, we compared MEDAL
with \texttt{umap.transform()}, a commonly used practical tool for embedding new
data into an existing manifold. Throughout, M1, M2, and M3 denote cells from
macaque subjects 1, 2, and 3, respectively.

\paragraph{Additional mixed-subject batches on the M1 reference manifold.}

\begin{figure}[!p]
    \centering
    \includegraphics[width=\linewidth]{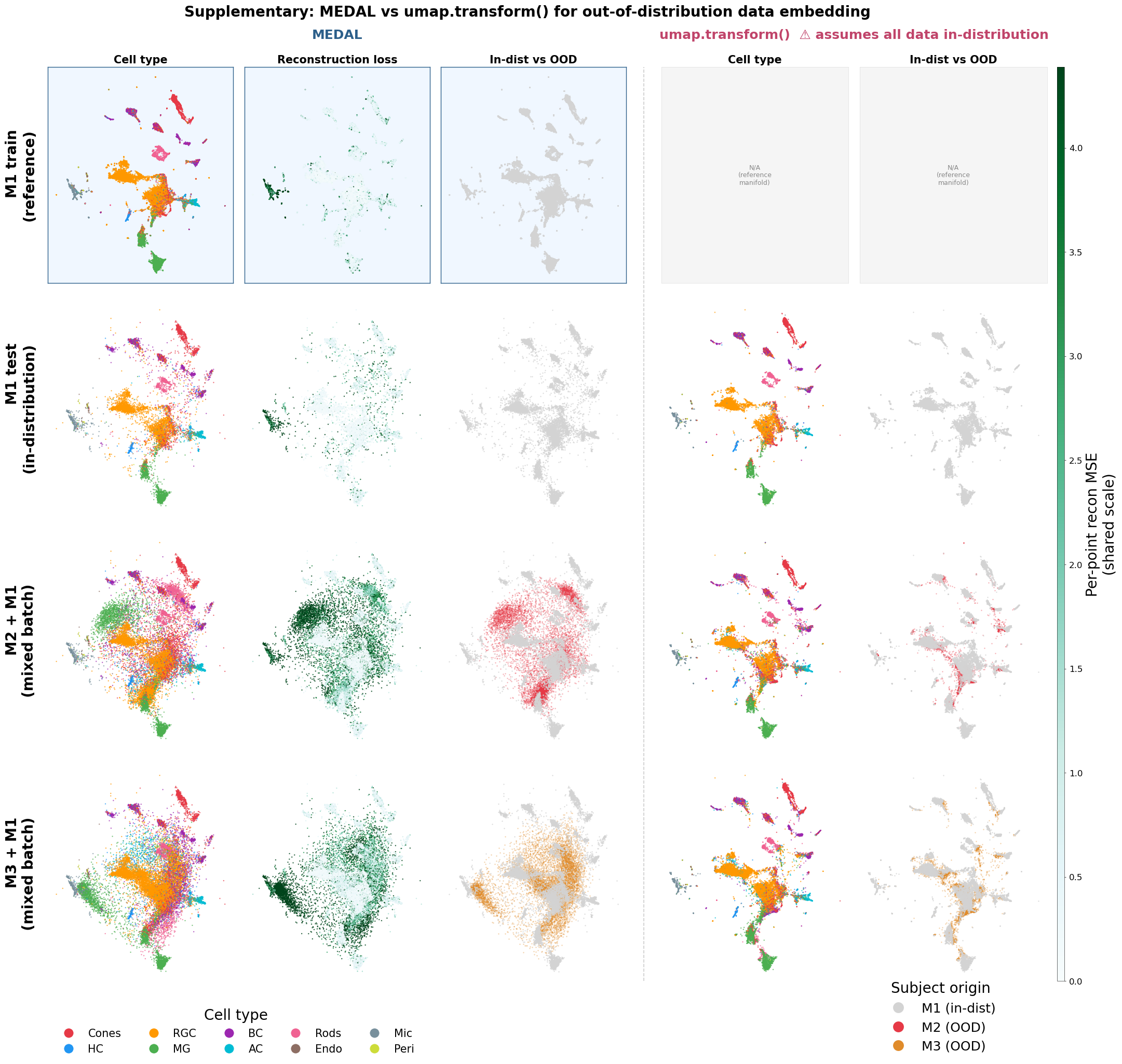}
    \caption{\textbf{MEDAL detects subject-level mismatch when embedding new macaque retina cells into a fixed reference manifold.}
MEDAL was trained on M1 training cells and used to embed M1 test cells,
M1 test + M2 mixed cells, and M1 test + M3 mixed cells onto the same M1 reference manifold.
For in-distribution M1 test cells, reconstruction error remains low and the
embedded cells align with the reference geometry. In mixed-subject batches,
out-of-distribution M2 and M3 cells occupy shifted regions of the embedding and
show spatially coherent increases in reconstruction error. The right panels show
the corresponding \texttt{umap.transform()} embeddings. Although
\texttt{umap.transform()} places new cells into the existing coordinate system,
it does not provide an accompanying reconstruction-based diagnostic of whether
those cells are supported by the reference distribution.}
\label{fig:macaque-embed}
\end{figure}

We first trained MEDAL on M1 training cells and projected three held-out datasets
onto the resulting M1 reference manifold: an in-distribution batch of M1 test
cells, a mixed batch containing M1 test and M2 cells, and a mixed batch
containing M1 test and M3 cells. For the in-distribution M1 test batch, embedded
cells remained aligned with the reference geometry and reconstruction error was
low. In contrast, both mixed-subject batches showed spatially coherent increases
in reconstruction error concentrated primarily among the out-of-distribution
cells (\autoref{fig:macaque-embed}). Thus, the shift-detection behavior observed
in the main text is not specific to the M1-versus-M2 comparison, but extends to
a second held-out subject.

\begin{figure}[!p]
    \centering
    \includegraphics[width=\linewidth]{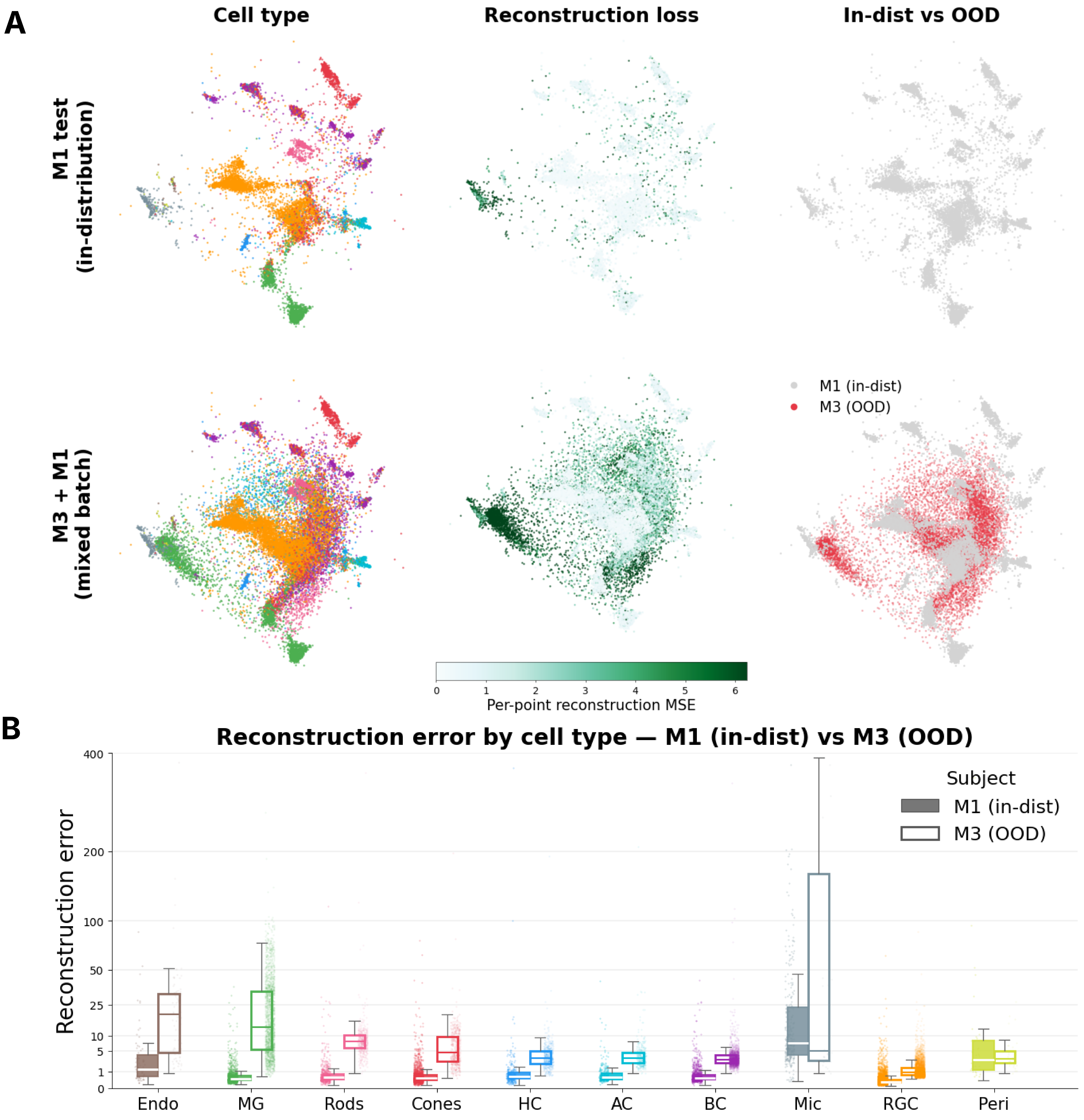}
    \caption{\textbf{Cell-type-resolved reconstruction error for M1 versus M3 in macaque retina.}
MEDAL was trained on M1 training cells and used to project held-out M1 test
cells and M3 cells onto the M1 reference manifold. Reconstruction error
distributions are shown by retinal cell type for in-distribution M1 cells and
out-of-distribution M3 cells. Errors are shown on a log scale. Relative to the
M1 baseline, M3 cells show elevated reconstruction error across most cell
types, with the largest shifts observed in M\"uller glia and endothelial
populations, followed by rods. This indicates that the subject-level mismatch
detected by MEDAL is concentrated in specific biological populations rather than
being uniformly distributed across the atlas.}
\label{fig:macaque-m3-vs-m1}
\end{figure}

We further quantified the M1-versus-M3 comparison by examining reconstruction
error within each retinal cell type (\autoref{fig:macaque-m3-vs-m1}). Relative
to the M1 in-distribution baseline, M3 cells showed elevated reconstruction
error across most cell types, with the largest shifts observed in M\"uller glia
and endothelial populations, followed by rods. This confirms that the
subject-level mismatch detected by MEDAL is not only a global batch effect, but
is concentrated in specific biological populations.

\begin{figure}[!p]
    \centering
    \includegraphics[width=\linewidth]{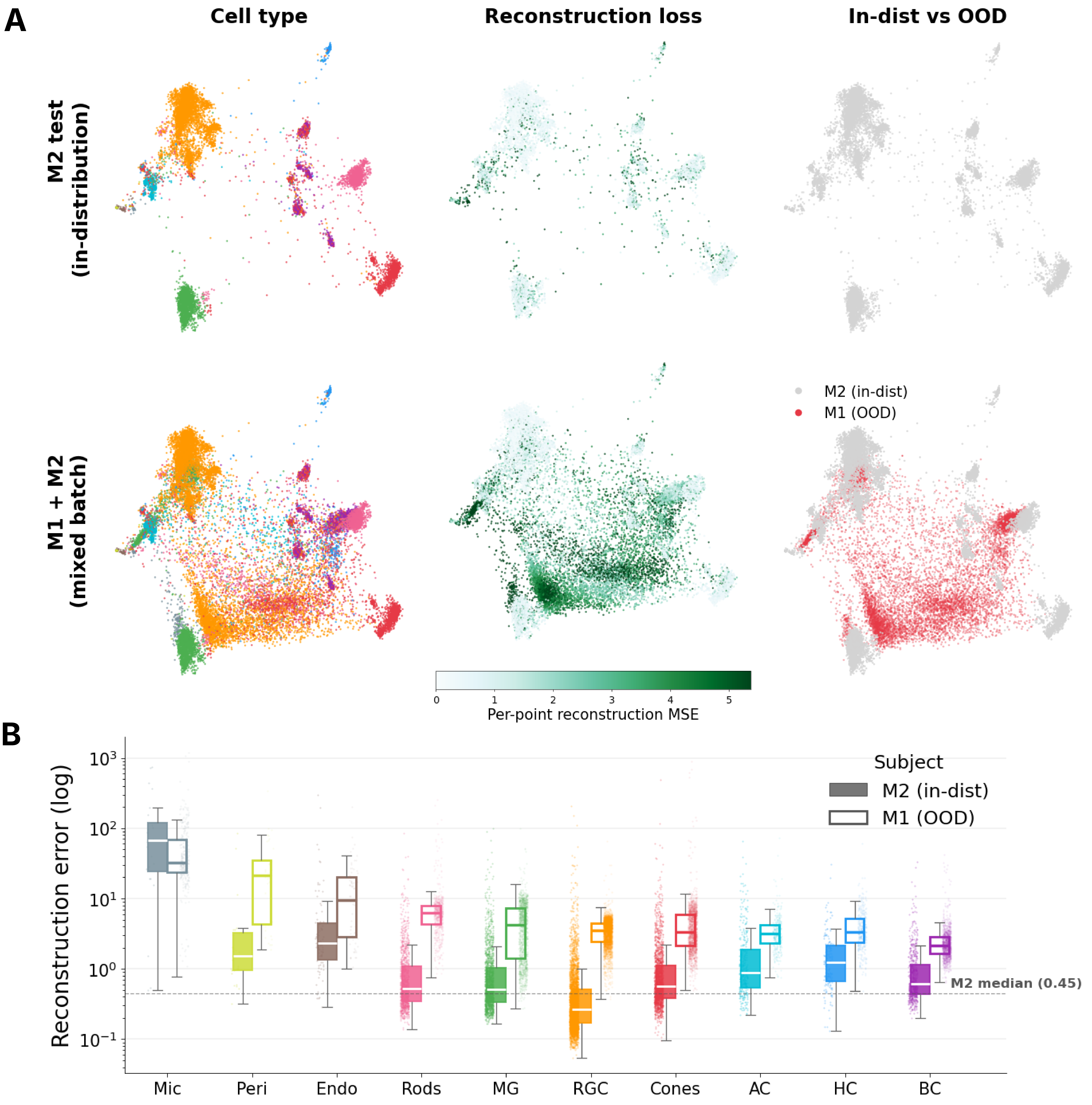}
    \caption{\textbf{Reverse-reference experiment with M2 as the reference manifold.}
MEDAL was trained on M2 training cells and used to embed held-out M2 test cells
and a mixed batch containing M1 and M2 cells onto the M2 reference manifold.
In-distribution M2 cells remain aligned with the reference geometry and show
lower reconstruction error, whereas out-of-distribution M1 cells exhibit
elevated reconstruction error. The strongest shifts again occur in
non-neuronal populations, indicating that the subject-level mismatch detected by
MEDAL is not specific to using M1 as the reference subject. Reconstruction error is shown on a log scale.}
\label{fig:macaque2}
\end{figure}

\paragraph{Reverse-reference experiment.}
We next reversed the reference subject by training MEDAL on M2 cells and
embedding held-out M2 test cells together with a mixed batch containing M1 and
M2 cells onto the M2 reference manifold (\autoref{fig:macaque2}). The same broad
pattern reappeared: out-of-distribution M1 cells showed elevated reconstruction
error relative to the M2 in-distribution baseline, with prominent shifts in
non-neuronal populations. This indicates that the subject-level mismatch
detected by MEDAL is not tied to a single reference subject, but reflects a more
stable pattern of inter-subject distributional shift.

\paragraph{Comparison with \texttt{umap.transform()}.}

Finally, we compared MEDAL with \texttt{umap.transform()} applied to the same M1
reference embedding (\autoref{fig:macaque-embed}). Both approaches can place new
cells into an existing low-dimensional coordinate system. However,
\texttt{umap.transform()} provides only coordinates: it does not report whether
the new cells are well supported by the reference distribution. As a result,
out-of-distribution M2 and M3 cells can appear visually integrated into the M1
manifold, even though the same cells exhibit clear subject-level mismatch under
MEDAL's reconstruction-error diagnostic.

This contrast highlights the practical value of MEDAL's encoder--decoder
construction. The encoder maps new cells into the reference manifold, while the
decoder quantifies whether those embedded cells can be faithfully reconstructed
from the reference geometry. Reconstruction error therefore acts as an
unsupervised reliability signal for out-of-sample embedding: it detects
subject-level mismatch, localizes the mismatch spatially on the manifold, and
identifies the cell populations most affected by the shift.

\subsection{Comparison to other methods} \label{sec:supp-comparison}
We compared MEDAL with three existing embedding-diagnostic methods: EMBEDR,
scDEED, and neMDBD. These methods differ in both their diagnostic target and
their computational workflow. EMBEDR assigns pointwise p-values by comparing an
embedding statistic against a permutation-based null distribution
\cite{johnson2022embedr}. scDEED identifies trustworthy, intermediate, and
dubious embedding structure using null embeddings and selects hyperparameters by
minimizing dubious points \cite{scdeeds}. neMDBD instead uses a local singularity score to measure geometric distortion \cite{liu2025assessingimprovingreliabilityneighbor}.
In contrast, MEDAL distills each candidate embedding into an encoder--decoder
model and selects hyperparameters by held-out reconstruction error. 

Because these methods are designed as complete systems, we evaluated each method
as it would be used in practice. Thus, the comparison is a whole-system
comparison: each method uses its own embedding implementation, optimization
pipeline, and recommended diagnostic criterion. As a result, two methods may
produce different embeddings even when they select the same nominal
hyperparameter. neMDBD was omitted from UMAP comparisons because the available
package does not support UMAP. Following \citet{scdeeds}, we applied
PCA preprocessing before running scDEED. We retained 6 principal components for
MNIST, 5 principal components for APOGEE and 4 principal components for Hydra and Neocortex. Consequently, scDEED
results are computed from PCA-reduced inputs rather than from the original
feature space, and should be interpreted as evaluating scDEED under its
recommended preprocessing workflow.
\FloatBarrier

\paragraph{Hyperparameter tuning and distortion comparisons}

For each dataset and teacher method, we show the hyperparameter-selection curve
for each diagnostic and the embedding selected by that method. The top row of
each embedding panel is colored by the available class, cell-type, or cluster
annotation, while the bottom row shows the method-specific pointwise diagnostic:
MEDAL reconstruction error, neMDBD singularity score, scDEED
trustworthy/intermediate/dubious labels, or EMBEDR p-values. These pointwise
scores are not on a common numerical scale; the comparison is qualitative and
aims to show whether the selected hyperparameter and diagnostic signal produce coherent
regions of embedding distortion.

Across these comparisons, the existing diagnostics sometimes select
hyperparameters at the boundaries of the candidate range, such as very small
perplexities or very large neighborhood sizes. We do not interpret these
boundary selections as inherently incorrect, since different diagnostics
emphasize different aspects of embedding quality. However, they often correspond
to visually distinct embedding regimes---for example, highly local embeddings
with fragmented structure or highly global embeddings with smoother,
more compressed geometry. As shown in the main case studies, such regimes can appear
plausible on the fitted embedding while differing substantially in how well they
support held-out observations. This is the central distinction of MEDAL: after
distilling each candidate embedding into an encoder--decoder model, MEDAL
evaluates reconstruction error on held-out data, providing an explicit
out-of-sample criterion for hyperparameter selection.

The resulting pointwise signals also differ in
interpretability. scDEED produces discrete trustworthy/intermediate/dubious
labels, which are useful for flagging suspicious embedding structure but can be
coarse for localizing graded distortion. neMDBD reports a singularity score,
which captures local geometric irregularity but can be spatially noisy and less
directly tied to information loss in the original feature space. EMBEDR provides
pointwise p-values for neighborhood preservation, but these are less directly
interpretable as a measure of how much high-dimensional signal is lost by the
embedding. By contrast, MEDAL's reconstruction error has a direct operational
interpretation because it measures how well an observation can be recovered after being
compressed through the fitted teacher manifold. In the examples below, this
often yields spatially coherent distortion maps and selects embeddings that
balance local separation with global organization, a pattern also reflected in
the complementary LCMC and triplet-accuracy metrics reported next.

\begin{figure}[!p]
    \centering
    \includegraphics[width=\linewidth]{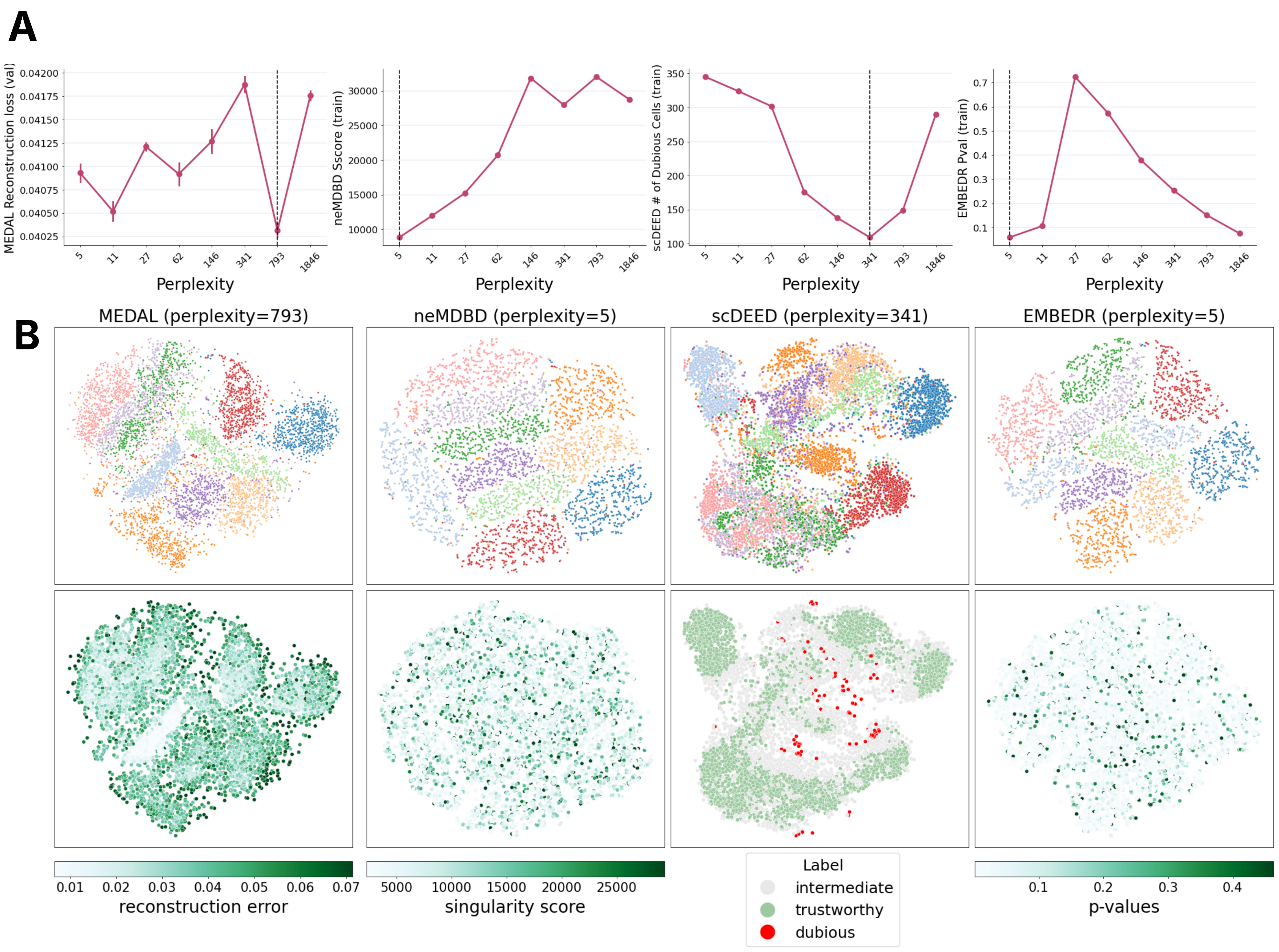}
    \caption{\textbf{Comparison of MEDAL with existing diagnostics on MNIST t-SNE
embeddings.}
\textbf{A,} Hyperparameter-selection curves across t-SNE perplexities for MEDAL,
neMDBD, scDEED, and EMBEDR. Dashed vertical lines indicate the selected
perplexity under each method's criterion. 
\textbf{B,} Embeddings selected by each method. Top row: embeddings colored by
digit label. Bottom row: method-specific pointwise diagnostics: MEDAL
reconstruction error, neMDBD singularity score, scDEED
trustworthy/intermediate/dubious labels, and EMBEDR p-values. MEDAL selects a
larger perplexity embedding, whereas neMDBD and EMBEDR select the smallest
perplexity in the grid.}
    \label{fig:mnist-compare-tsne}
\end{figure}

\begin{figure}[!p]
    \centering
    \includegraphics[width=\linewidth]{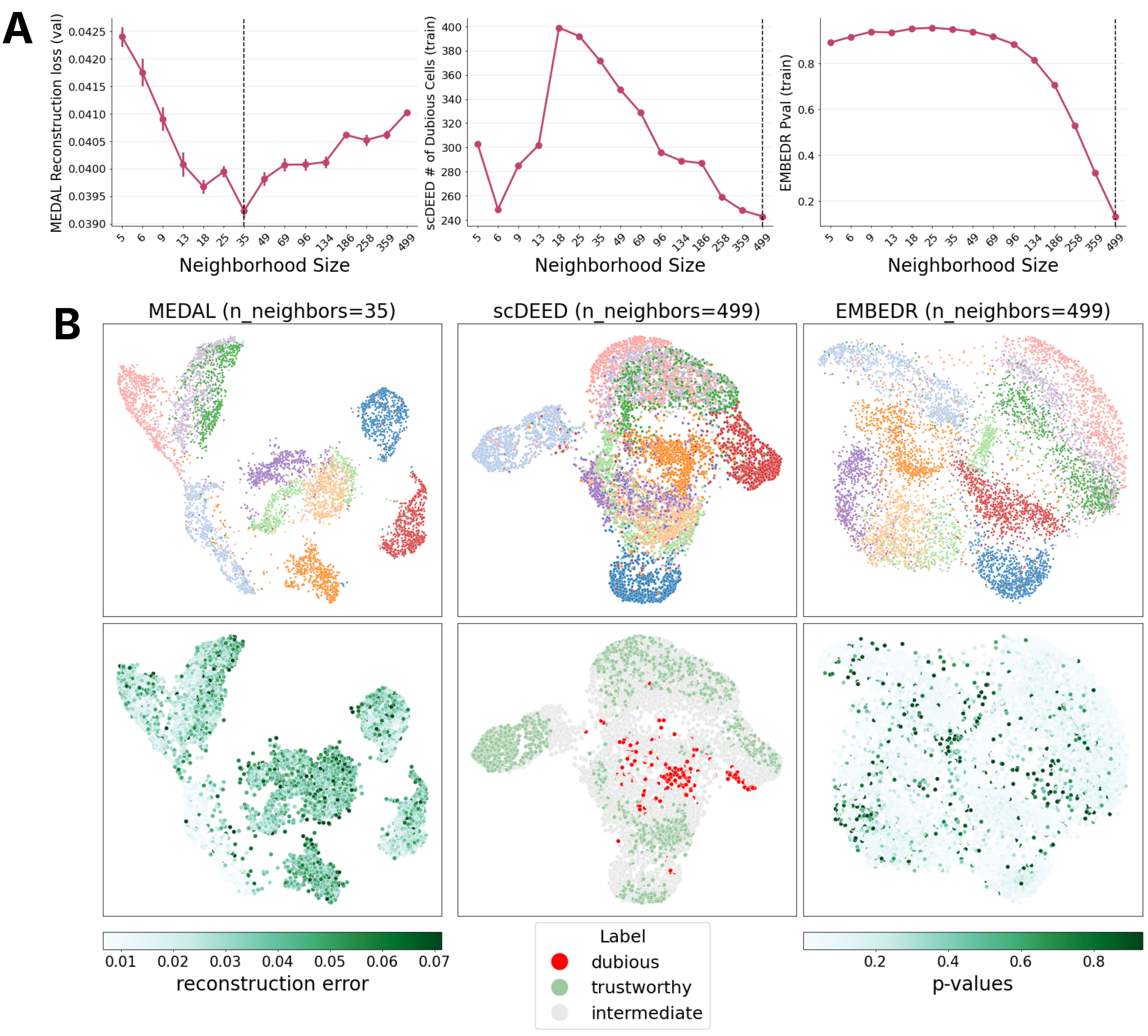}
   \caption{\textbf{Comparison of MEDAL with existing diagnostics on MNIST UMAP
embeddings.}
\textbf{A,} Hyperparameter-selection curves across UMAP neighborhood sizes for
MEDAL, scDEED, and EMBEDR. neMDBD is omitted because the available package does
not support UMAP. Dashed vertical lines indicate the selected neighborhood size.
\textbf{B,} Embeddings selected by each method. Top row: embeddings colored by
digit label. Bottom row: MEDAL reconstruction error, scDEED
trustworthy/intermediate/dubious labels, and EMBEDR p-values. MEDAL selects an
intermediate neighborhood size, whereas scDEED and EMBEDR select the largest
neighborhood size in the grid. neMDBD is omitted because the available implementation supports t-SNE but not UMAP.}
    \label{fig:mnist-compare-umap}
\end{figure}

\begin{figure}[!p]
    \centering
    \includegraphics[width=\linewidth]{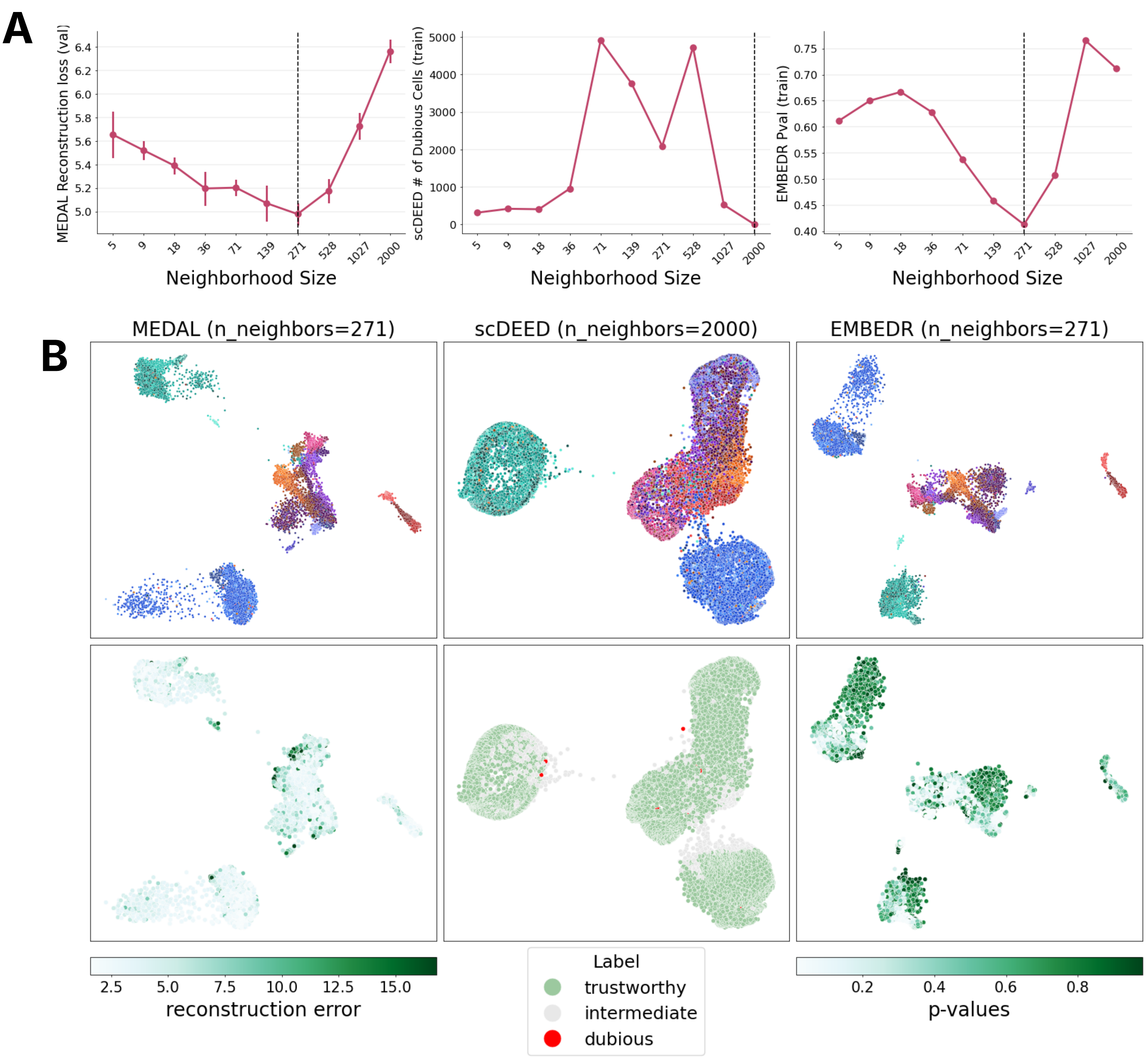}
    \caption{\textbf{Comparison of MEDAL with existing diagnostics on Hydra UMAP
embeddings.}
\textbf{A,} Hyperparameter-selection curves across UMAP neighborhood sizes for
MEDAL, scDEED, and EMBEDR. neMDBD is omitted because the available package does
not support UMAP. Dashed vertical lines indicate the selected neighborhood size.
\textbf{B,} Hydra embeddings selected by each method. Top row: embeddings colored
by reported cell type. Bottom row: MEDAL reconstruction error, scDEED
trustworthy/intermediate/dubious labels, and EMBEDR p-values. MEDAL and EMBEDR
select \(n=271\), whereas scDEED selects the largest neighborhood size in the
grid. Even when methods select the same nominal hyperparameter, embeddings and
diagnostic signals may differ because each method uses its own implementation
and optimization workflow. neMDBD is omitted because the available implementation supports t-SNE but not UMAP.}
    \label{fig:hydra-compare-umap}
\end{figure}

\newpage
\begin{figure}[!p]
    \centering
    \includegraphics[width=\linewidth]{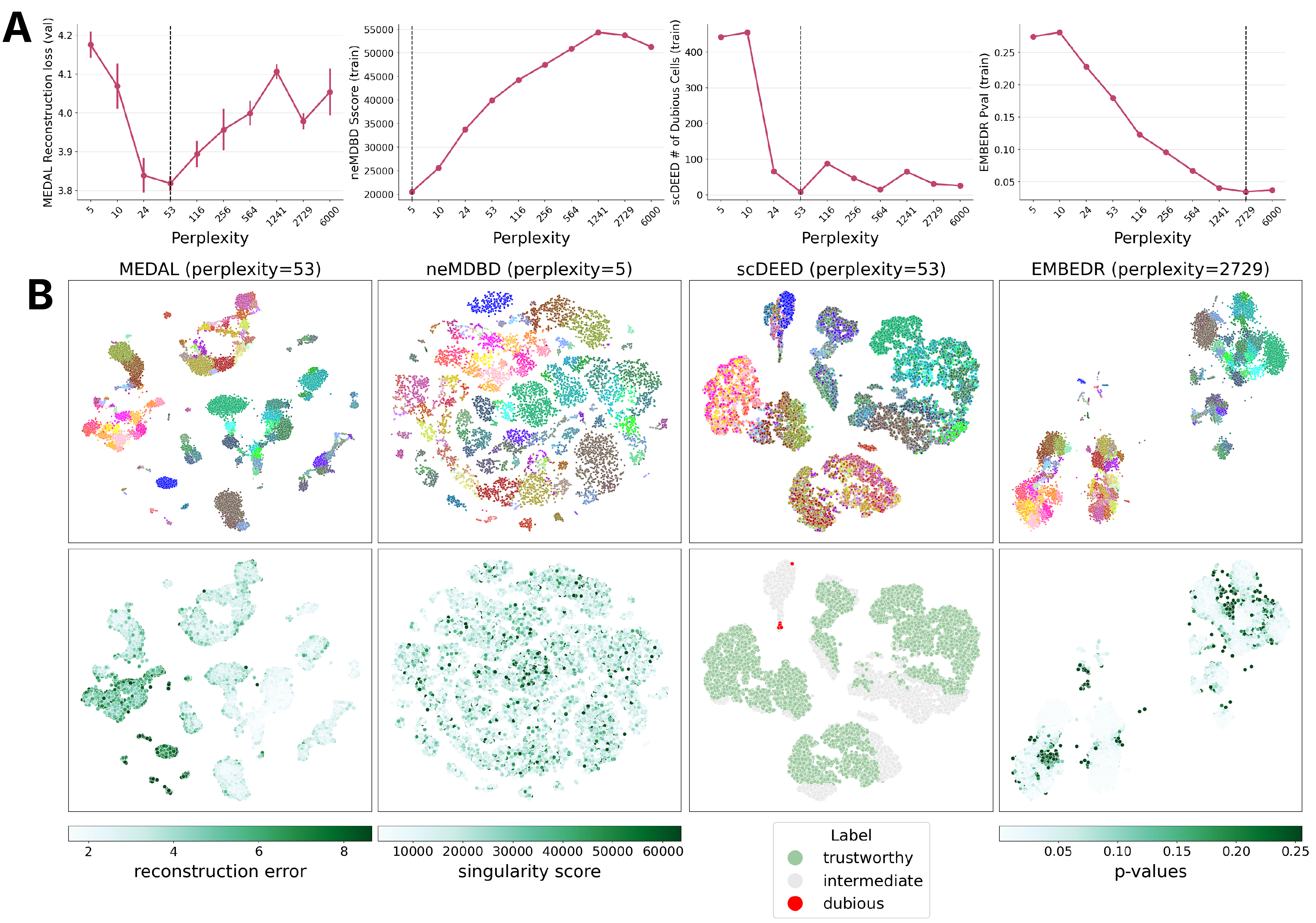}
    \caption{\textbf{Comparison of MEDAL with existing diagnostics on Neocortex t-SNE
embeddings.}
\textbf{A,} Hyperparameter-selection curves across t-SNE perplexities for MEDAL,
neMDBD, scDEED, and EMBEDR. Dashed vertical lines indicate the selected
perplexity under each method's criterion.
\textbf{B,} Neocortex embeddings selected by each method. Top row: embeddings colored
by annotated transcriptomic cell type. Bottom row: MEDAL reconstruction error,
neMDBD singularity score, scDEED trustworthy/intermediate/dubious labels, and
EMBEDR p-values. MEDAL and scDEED select an intermediate perplexity, whereas
neMDBD selects the smallest perplexity and EMBEDR selects a large perplexity.}
    \label{fig:tasic-compare-tsne}
\end{figure}

\newpage
\begin{figure}[!p]
    \centering
    \includegraphics[width=\linewidth]{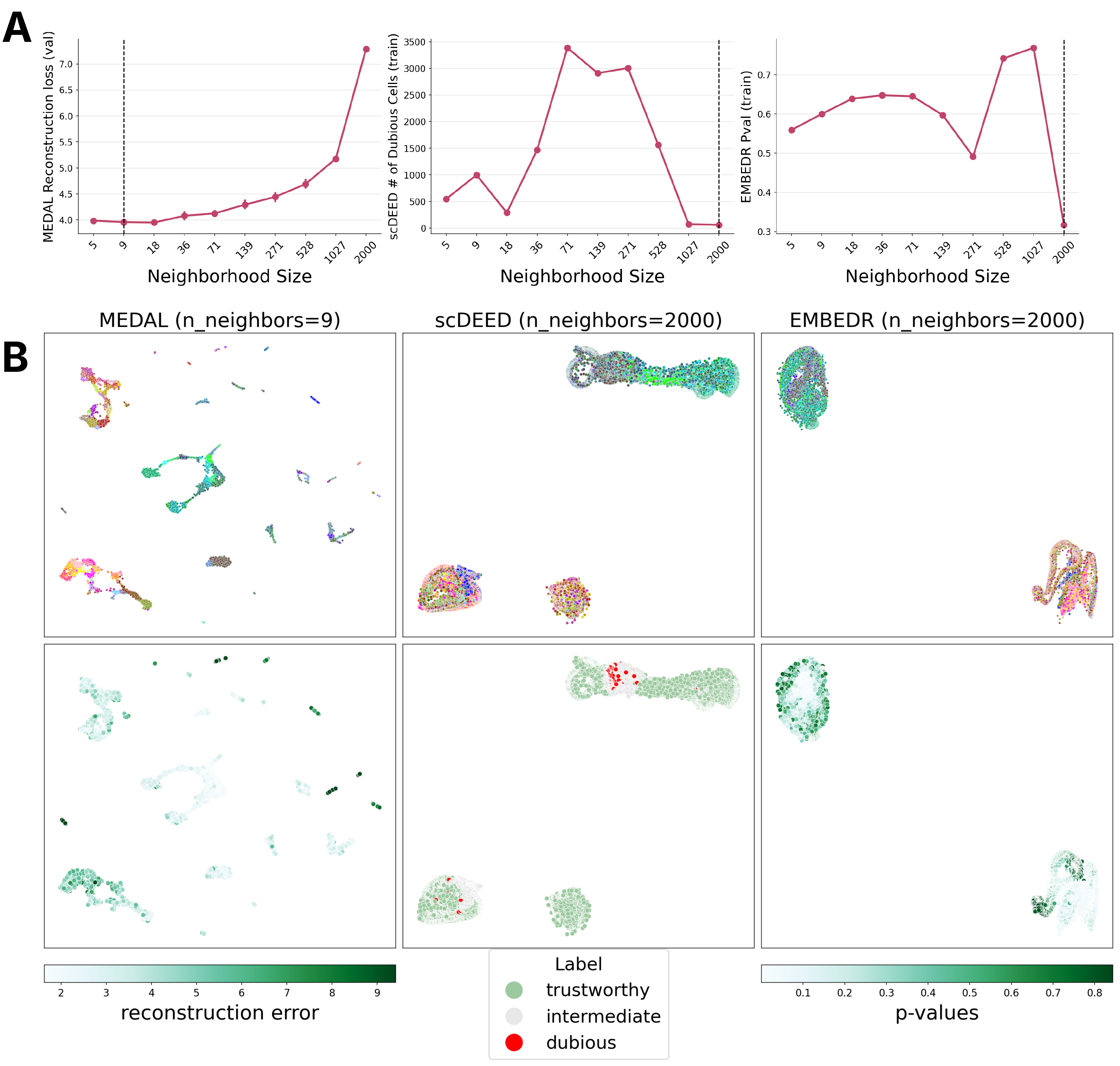}
    \caption{\textbf{Comparison of MEDAL with existing diagnostics on Neocortex UMAP
embeddings.}
\textbf{A,} Hyperparameter-selection curves across UMAP neighborhood sizes for
MEDAL, scDEED, and EMBEDR. neMDBD is omitted because the available package does
not support UMAP. Dashed vertical lines indicate the selected neighborhood size.
\textbf{B,} Neocortex embeddings selected by each method. Top row: embeddings colored
by annotated transcriptomic cell type. Bottom row: MEDAL reconstruction error,
scDEED trustworthy/intermediate/dubious labels, and EMBEDR p-values. MEDAL
selects a small neighborhood size, whereas scDEED and EMBEDR select the largest
neighborhood size in the grid. neMDBD is omitted because the available implementation supports t-SNE but not UMAP.}
    \label{fig:tasic-compare-umap}
\end{figure}
\FloatBarrier

\paragraph{Global and local metrics}
We also evaluated the embeddings selected by each method using two external
structure-preservation metrics. For global structure preservation, we used
random triplet accuracy, defined as the proportion of randomly sampled triplets
whose relative pairwise-distance ordering is preserved between the original and
embedded spaces \cite{an2025consensusdimensionreductionmultiview}. For local
structure preservation, we used the Local Continuity Meta-Criterion (LCMC), which
measures overlap between nearest-neighbor sets in the original and embedded
spaces \cite{an2025consensusdimensionreductionmultiview}. These metrics are
used only as complementary summaries since no single unsupervised metric fully
characterizes embedding quality, and local and global criteria often favor
different hyperparameter regimes. 

\autoref{tab:supp_full_results} reports triplet accuracy and LCMC across all
datasets, teacher methods, and diagnostics at each method's selected
hyperparameter. MEDAL is competitive across settings and often achieves the
strongest local-structure preservation, particularly for UMAP comparisons,
whereas other methods sometimes achieve higher triplet accuracy by selecting
larger or edge-of-grid hyperparameters. scDEED results are marked separately
because scDEED is evaluated relative to its post-PCA-preprocessed embedding rather
than the embedding reduced from the original feature space.
\begin{table}
\centering
\small
\caption{Triplet accuracy and LCMC at each method's selected hyperparameter,
         evaluated on embeddings refit on pooled train and validation data.
         For t-SNE the selected hyperparameter is perplexity; for UMAP it is
         the number of nearest neighbors $n$.
         MEDAL results are mean $\pm$ SE across 3 seeds.
         \textbf{Bold} indicates the best reported value within each dataset
         and teacher family. LCMC is averaged over neighborhood sizes sampled at equal
         intervals within the following ranges: Neocortex $[10, 300]$ (step 20),
         Hydra $[10, 200]$ (step 20), MNIST $[5, 50]$ (step 5),
         APOGEE $[5, 20]$ (step 1).
         Because this is a whole-system comparison, identical nominal hyperparameters do not imply identical embeddings; for example, MEDAL and EMBEDR both select \(n=271\) on Hydra UMAP but use different embedding implementations and therefore produce different metric values.
         $^\dagger$scDEED metrics are computed relative to a PCA-reduced input space rather than the original feature space and are therefore not directly comparable to the other methods; see \hyperref[par:stability-implementation-caveats]{Stability and implementation caveats}.}
\label{tab:supp_full_results}
\setlength{\tabcolsep}{4pt}
\renewcommand{\arraystretch}{1.2}

\begin{tabular}{ll ccc ccc}
\toprule
& & \multicolumn{3}{c}{\textbf{t-SNE}} & \multicolumn{3}{c}{\textbf{UMAP}} \\
\cmidrule(lr){3-5} \cmidrule(lr){6-8}
\textbf{Dataset} & \textbf{Method}
    & \texttt{perplexity} & \textbf{Triplet acc.} & \textbf{LCMC}
    & $n$ & \textbf{Triplet acc.} & \textbf{LCMC} \\
\midrule

\multirow{4}{*}{Neocortex}
    & neMDBD
        & 5    & 0.5872            & 0.454
        & ---  & ---               & ---   \\
    & EMBEDR
        & 2729 & \textbf{0.7964}          & {0.486}
        & 2000 & 0.7214            & 0.056 \\
    & scDEED$^\dagger$
        & 2729 & 0.7256            & 0.193
        & 2000 & \textbf{0.7596}            & 0.0802 \\
    & MEDAL
        & 53 & {0.683}   & \textbf{0.528}
        & 9   & {0.7106}   & \textbf{0.4677} \\

\midrule

\multirow{4}{*}{Hydra}
    & neMDBD
        & 5    & 0.6362            & 0.386
        & ---  & ---               & ---   \\
    & EMBEDR
        & 4999 & \textbf{0.7446}   & 0.314
        & 271  & \textbf{0.6236}   & \textbf{0.244} \\
    & scDEED$^\dagger$
        & 4999 & 0.6646            & 0.1646
        & 2000 & 0.621            & 0.0741 \\
    & MEDAL
        & 499 & 0.6628            & \textbf{0.4229}
        & 271    & {0.614}   & {0.241} \\

\midrule

\multirow{4}{*}{MNIST}
    & neMDBD
        & 5    & 0.6112            & 0.418
        & ---  & ---               & ---   \\
    & EMBEDR
        & 5    & 0.6544            & \textbf{0.425}
        & 499  & 0.6302            & 0.253 \\
    & scDEED$^\dagger$
        & 341  & 0.6574            & 0.249
        & 499  & 0.6362            & 0.2145 \\
    & MEDAL
        & 793  & \textbf{0.6804}   & 0.383
        & 35   & \textbf{0.6396}   & \textbf{0.342} \\

\midrule

\multirow{4}{*}{APOGEE}
    & neMDBD
        & 3    & 0.6480            & \textbf{0.391}
        & ---  & ---               & ---   \\
    & EMBEDR
        & 499  & \textbf{0.7418}   & 0.264
        & 499  & 0.5006            & 0.096 \\
    & scDEED$^\dagger$
        & 499  & 0.6688            & 0.2145
        & 134  & 0.5036            & 0.0961 \\
    & MEDAL
        & 6    & 0.6426            & 0.348
        & 9    & \textbf{0.5994}   & \textbf{0.3375} \\

\bottomrule
\end{tabular}
\end{table}

\paragraph{LCMC k-range choices}
LCMC was computed over a dataset-specific range of neighborhood sizes: Neocortex
\(k \in [10,300]\) with step size 20, Hydra \(k \in [10,200]\) with step size
20, MNIST \(k \in [5,50]\) with step size 5, and APOGEE \(k \in [5,20]\) with
step size 1. These ranges were chosen to reflect differences in dataset size
and annotation granularity. For the larger single-cell datasets, we evaluated
LCMC over broader neighborhoods so that the metric captured local cell-state or
cell-type structure without being dominated by only the nearest few neighbors.
For the smaller or more discretely clustered datasets, we used smaller
neighborhood ranges to focus the metric on local neighborhood preservation.
Because the choice of \(k\)-range affects the numerical value of LCMC, these
values should be interpreted primarily as within-dataset comparisons across
methods, rather than as absolute measures that are comparable across datasets.

\paragraph{Runtime and scalability}

\begin{figure}[!p]
    \centering
    \includegraphics[width=\linewidth]{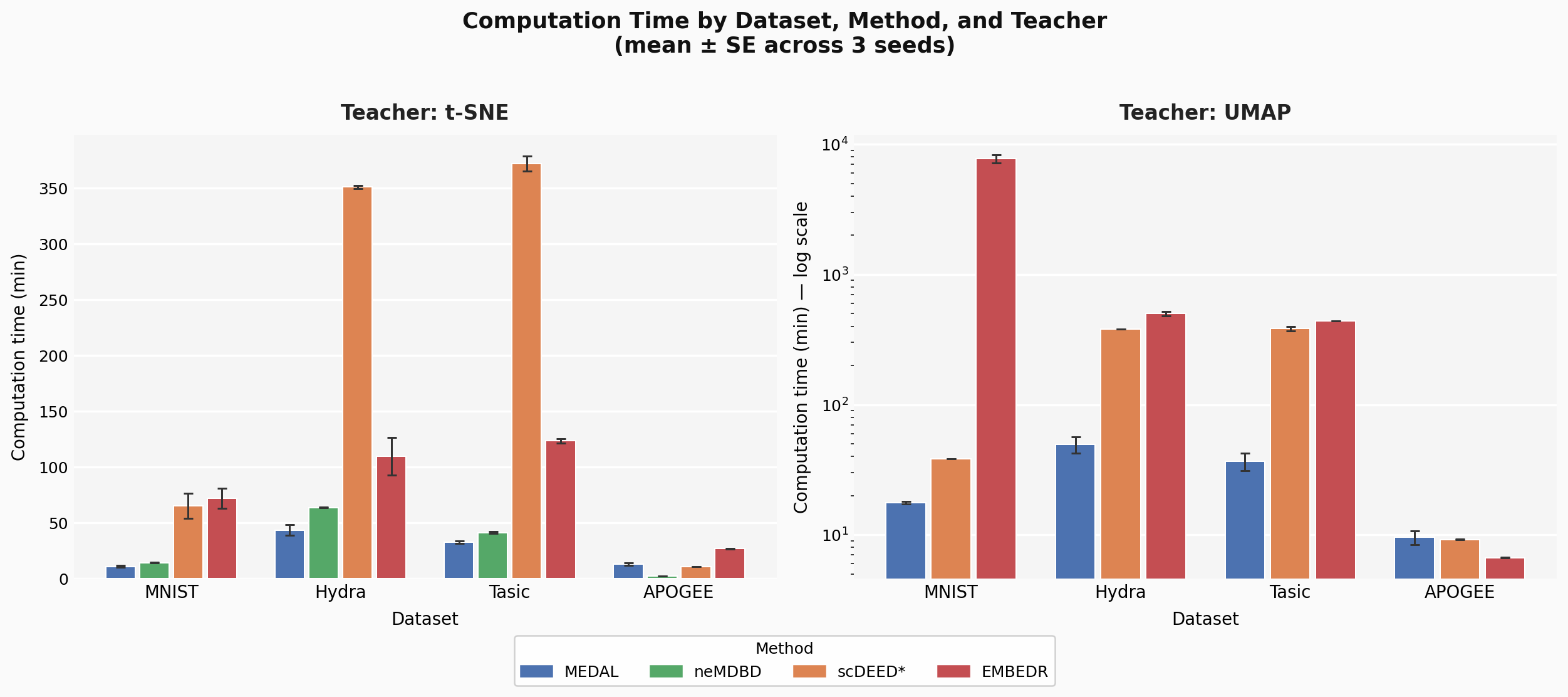}
    \caption{\textbf{Runtime comparison across embedding diagnostics.}
Wall-clock runtime for one full hyperparameter sweep across datasets and teacher
methods. Times reflect each method's complete deployed workflow for a single
seed, including its own embedding implementation and diagnostic computation.
For MEDAL, this corresponds to one sweep of student distillations across the
candidate teacher hyperparameters; repeated sweeps are used to average
validation curves. scDEED and EMBEDR runtimes include their null- or
permutation-based calculations. neMDBD runtimes report only the singularity
score, because the permutation score was computationally prohibitive at these
dataset scales.}
    \label{fig:runtime-all-methods}
\end{figure}

We measured wall-clock runtime for a full hyperparameter sweep for each method
on each dataset and teacher setting where the method was available
(\autoref{fig:runtime-all-methods}). These runtimes reflect one complete
workflow per seed, including embedding computation and diagnostic scoring.
For MEDAL, this corresponds to one sweep of student distillations across the
candidate teacher hyperparameters; repeated MEDAL runs can then be used to
estimate averaged validation curves. EMBEDR and scDEED include their
permutation- or null-based diagnostic calculations. neMDBD runtimes report the
singularity score only, because the permutation score was computationally
prohibitive at the dataset scales considered here.

\begin{figure}[!p]
    \centering
    \includegraphics[width=\linewidth]{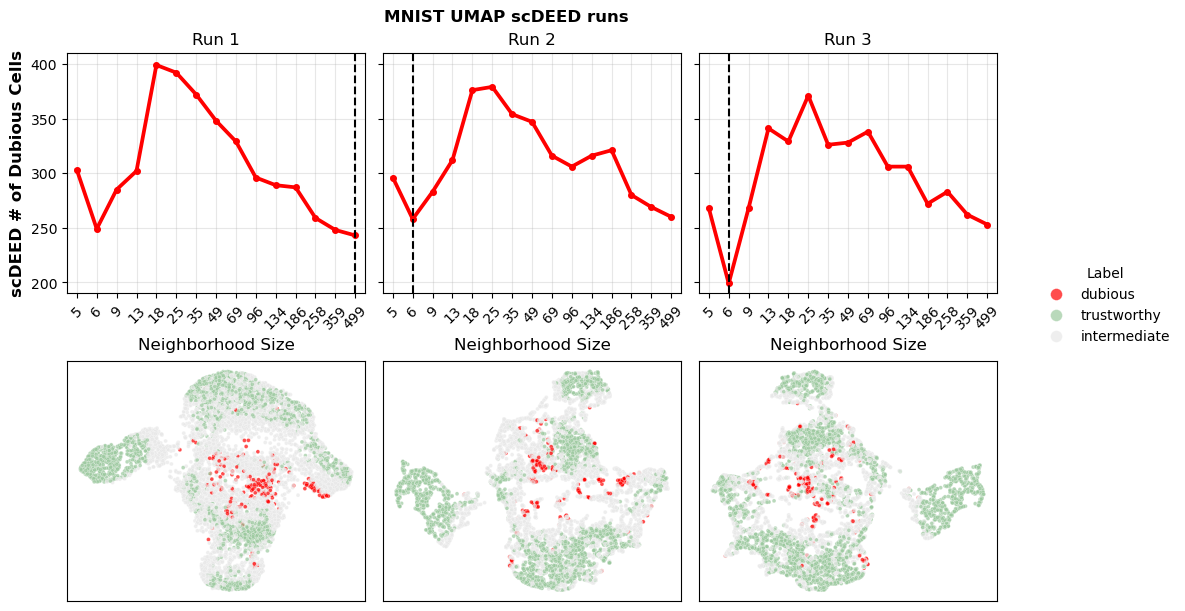}
    \caption{\textbf{Sensitivity of scDEED hyperparameter selection to random seed.}
scDEED was run across three random seeds while holding the reference embeddings for each hyperparameter value fixed. The selected hyperparameter and resulting embedding-quality metrics
vary across seeds, illustrating the stochasticity introduced by null-embedding
construction.}
    \label{fig:scdeed-unstable}
\end{figure}

\paragraph{Stability and implementation caveats} \label{par:stability-implementation-caveats}
Because scDEED relies on null embeddings, its selected hyperparameter can vary
with random seed even when the embedding being scored is held fixed. In
\autoref{fig:scdeed-unstable}, we show scDEED results across three random seeds,
illustrating variation in both selected hyperparameter and downstream embedding
quality metrics. This motivates our use of repeated MEDAL distillations and
averaged validation curves for stable selection.

A limitation of this comparison is that the three systems use different t-SNE implementations internally — EMBEDR uses a Python/Cython Barnes-Hut implementation, neMDBD uses Rtsne (C++), and MEDAL uses openTSNE — which produce numerically distinct embeddings even at identical perplexity values. This arises from differences in kNN graph construction, affinity matrix bandwidth search tolerances, and gradient computation backends, all of which are internal to each implementation and not externally controllable without altering the methods themselves. As a result, observed metric differences reflect the performance of each method as a whole system, encompassing both hyperparameter selection and embedding computation, rather than isolating the contribution of hyperparameter selection alone. A fully controlled comparison would require fixing a canonical embedding implementation across all methods; however, this would require each method to re-select its hyperparameter on embeddings it was not designed to score, introducing a different form of bias. We therefore adopt the whole-system evaluation as the most faithful representation of how each method would be used in practice. For UMAP, all three methods use their own internal UMAP implementations, and the same argument applies.

Finally, neMDBD proposes both a singularity score and a permutation score for evaluating embeddings. We report only the singularity score in our comparison due to the prohibitive computational cost of the permutation score at the dataset scales considered here.

\section{Related Work}
\subparagraph{Nonparametric and parametric dimension reduction.}
A broad range of methods has been developed for dimension reduction. Linear
methods such as PCA provide an explicit parametric map and an exact notion of
reconstruction error, making them interpretable and easy to evaluate.
Nonlinear methods such as Isomap \cite{tenenbaum2000global}, LLE
\cite{roweis2000nonlinear}, t-SNE \cite{van2008visualizing}, UMAP
\cite{mcinnes2020umapuniformmanifoldapproximation}, PaCMAP
\cite{wang2021understandingdimensionreductiontools}, and PHATE
\cite{moon2019visualizing} are designed to capture curved or
neighborhood-based structure that linear projections miss. These methods have
become central tools for exploratory analysis, especially in scientific
applications where low-dimensional visualizations are used to discover clusters,
trajectories, and other latent structure.

Despite their empirical success, many popular manifold-learning methods produce
embeddings only for the observed training data rather than a reusable map from
input space to embedding space. Existing out-of-sample and parametric extensions
\cite{bengio2003out, van2009learning, sainburg2021parametric} partially
address this limitation by learning an explicit forward map. However, these
approaches are typically method-specific and usually optimize the original
neighbor-preserving objective rather than faithfully reproducing a fixed,
precomputed embedding. Moreover, most parametric neighbor-embedding methods
provide an encoder but no decoder, making it difficult to evaluate how much
high-dimensional information is retained by the learned low-dimensional
representation.

\subparagraph{Autoencoders and geometry-regularized representation learning.}
Autoencoder based methods provide a complementary approach to representation
learning by optimizing a reconstruction objective through a low-dimensional
bottleneck \cite{tschannen2018recent}. Unlike most nonparametric manifold
learners, autoencoders naturally provide both a forward map into the latent space
and a decoder back to the original feature space. This makes reconstruction loss
a natural measure of compression fidelity for the autoencoder's own latent
representation. However, vanilla autoencoders do not generally enforce agreement
with a specific manifold-learning embedding, and their latent spaces may not
reproduce the neighborhood structure or visual geometry selected by methods such
as t-SNE or UMAP. As a result, their reconstruction error cannot be interpreted
as the information loss induced by a particular fitted manifold embedding.

Several methods therefore combine autoencoding with geometric regularization,
for example by encouraging the latent representation to preserve distances,
neighborhoods, graphs, or other geometric relationships computed from the input
data. Closest in spirit are Geometry Regularized Autoencoders (GRAE)
\cite{Duque_2020}, which augment the autoencoder objective with a geometric
regularization term so that the latent representation follows a
manifold-learning geometry while retaining out-of-sample extension and
reconstruction. MEDAL differs in both objective and use case. Rather than
proposing a new dimension-reduction method, MEDAL treats a fitted teacher
embedding as the object of evaluation, explicitly enforces near-exact recovery
of that teacher embedding, and then interprets reconstruction error conditional
on successful distillation.

\subparagraph{Automated hyperparameter selection.}
A related line of work aims to make dimension-reduction workflows less dependent
on manual visual tuning. For t-SNE, \citet{cao2017automaticselectiontsneperplexity}
propose model-selection objectives based on information criteria such as BIC or
MDL to choose perplexity values automatically. Other work uses
constraint-preserving scores, Bayesian optimization, or multi-objective
surrogate models to tune hyperparameters for methods such as t-SNE, UMAP, and
LargeVis \cite{9476903, liao2023efficientrobustbayesianselection}. HyperNP
further accelerates interactive exploration by training a neural network to
approximate projections across hyperparameter values, enabling approximate
t-SNE, UMAP, or Isomap embeddings to be generated in real time
\cite{appleby2021hypernpinteractivevisualexploration}.

Domain-specific studies have also proposed automated tuning strategies. For
example, \citet{Lin_2024} select t-SNE and UMAP hyperparameters by modeling data
as signal plus noise and maximizing neighborhood-preservation with respect to an
estimated signal. Other approaches optimize UMAP or manifold-learning
hyperparameters using downstream objectives such as clustering quality,
classification accuracy, or task-level performance
\cite{gikera2025khyperparam, 10811181, nasim2025automatedmanifoldlearningreduced}.
Together, these methods mark an important shift from ad hoc visual inspection
toward systematic hyperparameter search. However, their conclusions still depend
on the chosen quality metric, task objective, or algorithm-specific criterion.
They therefore do not by themselves define a method-agnostic measure of
embedding information loss, nor do they provide a common scale for comparing
different manifold learners.

\subparagraph{Embedding quality metrics and local reliability diagnostics.}
Another line of work evaluates the quality or reliability of a completed
embedding. Classical criteria such as trustworthiness, continuity, co-ranking
summaries, stress, and neighborhood-overlap scores quantify how well local or
global relationships in the high-dimensional data are preserved after
projection. These metrics are useful for diagnosing rank or neighborhood
distortions, but they typically operate directly on the completed embedding and
do not provide a learned inverse map, held-out reconstruction protocol, or
pointwise estimate of information loss in the original data space.

More recent diagnostic methods focus on local reliability and pointwise
distortion. EMBEDR evaluates embedding reliability by comparing local structure
in the observed embedding against a null or resampling-based reference
\cite{johnson2022embedr}. scDEED computes pointwise reliability scores using
permutations of nearest-neighbor distances and labels low-confidence points as
``dubious'' \cite{scdeeds}. Related continuity- or singularity-based
approaches diagnose where neighbor embeddings may behave unstably under local
perturbations \cite{liu2025assessingimprovingreliabilityneighbor}. These methods provide valuable warnings about
specific regions of an embedding, and some can be used to guide hyperparameter
selection. However, their outputs are often expressed as p-values, reliability
labels, singularity scores, or neighborhood-based diagnostics. Such quantities
are informative, but they are not always directly comparable across datasets,
methods, and hyperparameter settings, and they do not quantify reconstruction
loss in the original feature space.

\subparagraph{Positioning of MEDAL.}
MEDAL addresses a different problem from these prior lines of work. Rather than
proposing another manifold learner, another parametric approximation to a
neighbor-embedding objective, or another diagnostic computed directly on a
finished visualization, MEDAL converts any fitted teacher embedding into a
reusable encoder--decoder model. The encoder is trained to recover the fixed
teacher coordinates, while the decoder maps those coordinates back to the
original feature space. Conditional on near-zero teacher recovery, the remaining
reconstruction error measures how much input-space information is lost when data
are compressed through that teacher manifold.

This distinction gives MEDAL three practical advantages. First, because the
teacher embedding is fixed, MEDAL evaluates the information content of the
embedding actually used by the analyst rather than a newly optimized parametric
variant. Second, because MEDAL learns a decoder, it provides a direct
reconstruction-based measure of compression fidelity. Third, because the same
student class and reconstruction criterion can be applied to embeddings produced
by t-SNE, UMAP, Isomap, PCA, or other methods, MEDAL enables hyperparameter
settings and dimension-reduction algorithms to be compared on a common,
dataset-specific scale. In this way, MEDAL provides a bridge between classical
nonparametric manifold learning and modern parametric representation learning,
enabling quantitative validation, comparison, and selection of
dimension-reduction methods in scientific applications.
\end{document}